\documentclass[a4paper]{article}
\usepackage{amssymb,amsmath,amsthm}
\usepackage[colorlinks=false,allbordercolors={1 1 1}]{hyperref}
\usepackage{cleveref} 
\crefname{algocf}{alg.}{algs.}
\Crefname{algocf}{Algorithm}{Algorithms}

\title{Efficient Numerical Integration in Reproducing Kernel Hilbert Spaces via Leverage Scores Sampling}

\usepackage{authblk}
\author[1]{Antoine Chatalic}
\author[1]{Nicolas Schreuder}
\author[2]{Ernesto De Vito}
\author[1,3,4]{Lorenzo Rosasco}
\affil[1]{MaLGa Center - DIBRIS - Università di Genova, Genoa, Italy}
\affil[2]{MaLGa Center - DIMA - Università di Genova, Genoa, Italy}
\affil[3]{CBMM - Massachusets Institute of Technology, Cambridge, MA, USA}
\affil[4]{Istituto Italiano di Tecnologia, Genoa, Italy}

\newif\iffancyversion
\fancyversiontrue	

\usepackage{fancyvrb}
\usepackage[ruled,algosection,vlined,linesnumbered]{algorithm2e}
\crefalias{algocfline}{line}
\usepackage{mathtools}
\usepackage[overload]{empheq}
\usepackage{array}
\usepackage{booktabs,boldline}
\usepackage{cellspace}	
\usepackage{multirow}
\usepackage{commath} 
\usepackage{graphicx}
\usepackage{nameref}
\hypersetup{ colorlinks = true, citecolor = blue, linkcolor=blue }
\usepackage{cleveref}
\usepackage{nicefrac}
\usepackage{braket}
\usepackage{makecell}
\usepackage{xparse}

\usepackage[table,x11names,usenames,dvipsnames]{xcolor}
\usepackage{tikz}
\usetikzlibrary{backgrounds,intersections}
\usepackage{pgfplots}

\usepackage{fullpage}

\usepackage[%
style=authoryear-comp,
sorting=nyt,
hyperref=true,
backend=biber,
maxbibnames=8,
maxcitenames=1,
uniquelist=false]{biblatex}
\usepackage{import} 
\usepackage{etoolbox}
\usepackage[most]{tcolorbox}
\usepackage{xstring} 

\definecolor{customBlue}{RGB}{18,75,126}
\colorlet{titleCol}{customBlue}	
\colorlet{titleThmCol}{titleCol} 
\colorlet{backCol}{titleCol!08!white}
\colorlet{backThmCol}{titleThmCol!08!white}

\tcbset{
  compactdefaultstyle/.style={
	breakable,
	theorem style=plain,
	enhanced,
	sharp corners,
	frame hidden,
	colback=backCol,
	opacityfill=0, 
	fontupper=\itshape,
	fonttitle=\bfseries\upshape,
	coltitle=black, 
	boxrule=0pt,
	before skip=1mm,
	after skip=1mm,
	right = 0mm, 
	left = -1mm, 
	top = 0mm,
	bottom = 0mm,
  },
  defaultthmstyle/.style={
	breakable,
	theorem style=plain,
	enhanced,
	sharp corners,
	frame hidden,
	colback=backCol,
	opacityfill=0, 
	fontupper=\itshape,
	fonttitle=\bfseries\upshape,
	coltitle=black, 
	boxrule=0pt,
	before skip=2mm,
	after skip=2mm,
	right = 0mm, 
	left = -1mm, 
	top = 1mm,
	bottom = 1mm,
  },
  defaultproofstyle/.style={
	breakable,
	theorem style=plain,
	terminator sign={.},  
	enhanced,
	sharp corners,
	frame hidden,
	colback=backCol,
	opacityfill=0, 
	fontupper=\upshape,
	fonttitle=\itshape,
	separator sign={:\ }, 
	coltitle=black, 
	boxrule=0pt,
	before skip=2mm,
	after skip=2mm,
	right = 0mm, 
	left = -1mm, 
	top = 1mm,
	bottom = 1mm,
  },
  thmstyle/.style={
	breakable,
	sharp corners,
	colbacktitle=titleThmCol,
	colback=backThmCol,
	colframe=titleThmCol, 
	fonttitle=\upshape\bfseries\hypersetup{linkcolor=white,citecolor=white},
	left = 2mm, 
	right = 2mm, 
	before skip=2mm,
	after skip=2mm,
  },
  lemmastyle/.style={
	breakable,
	enhanced,
	theorem style=plain,
	frame hidden,
	sharp corners,
	colback=backCol,
	boxrule=0pt,
	before skip=2mm,
	after skip=2mm,
	right = 2mm, 
	borderline={0.4mm}{0pt}{titleCol},
	fonttitle=\bfseries,
	coltitle=titleCol,
  },
  defstyle/.style={
	breakable,
	theorem style=plain,
	enhanced,
	sharp corners,
	frame hidden,
	colback=backCol,
	coltitle=titleCol,
	boxrule=0pt,
	borderline west={0.8mm}{0pt}{backCol},
	fonttitle=\upshape\bfseries\hypersetup{citecolor=titleCol},
	fontupper=\upshape,
	before skip=2mm,
	after skip=2mm,
	left = 3mm, 
  },
  proofstyle/.style={
	breakable,
	enhanced,
	theorem style=plain,
	frame hidden,
	sharp corners,
	colback=backCol,
	boxrule=0pt,
	before skip=2mm,
	after skip=2mm,
	right = 2mm, 
	borderline west={0.4mm}{0pt}{titleCol},
	fonttitle=\bfseries,
	coltitle=titleCol,
  }, 
  bluewestlinestyle/.style={
	breakable,
	theorem style=plain,
	enhanced,
	sharp corners,
	frame hidden,
	colback=black!03!white,
	coltitle=customBlue,
	boxrule=0pt,
	borderline west={0.8mm}{0pt}{customBlue},
	fonttitle=\upshape\bfseries\hypersetup{citecolor=customBlue,linkcolor=customBlue},
	before skip=2mm,
	after skip=2mm,
	left = 3mm, 
	right = 2mm, 
  },
  redvioletwestlinestyle/.style={
	breakable,
	theorem style=plain,
	enhanced,
	sharp corners,
	frame hidden,
	colback=black!03!white,
	coltitle=RedViolet,
	boxrule=0pt,
	borderline west={0.8mm}{0pt}{RedViolet},
	fonttitle=\upshape\bfseries\hypersetup{citecolor=RedViolet,linkcolor=RedViolet},
	before skip=2mm,
	after skip=2mm,
	left = 3mm, 
	right = 2mm, 
  },
  greenwestlinestyle/.style={
	breakable,
	theorem style=plain,
	enhanced,
	sharp corners,
	frame hidden,
	colback=black!03!white,
	coltitle=ForestGreen,
	boxrule=0pt,
	borderline west={0.8mm}{0pt}{ForestGreen},
	fonttitle=\upshape\bfseries\hypersetup{citecolor=ForestGreen,linkcolor=ForestGreen},
	before skip=2mm,
	after skip=2mm,
	left = 3mm, 
	right = 2mm, 
  },
}

\tcbset{todobox/.style={
  title={Some work is needed here}, 
  colframe=red!70!black,
  colback=red!10,
  fonttitle=\bfseries, 
  coltext=red!70!black, 
  breakable,
  boxed title style={
    outer arc=0pt,
    arc=0pt,
    top=0pt,
    bottom=0pt,
    },
  }
}

\tcbset{warnbox/.style={
  colframe=black,
  colback=yellow!20,
  breakable,
  boxed title style={
    outer arc=0pt,
    arc=0pt,
    top=0pt,
    bottom=0pt,
    },
  }
}
\tcbset{infobox/.style={
  colframe=blue,
  colback=blue!05,
  breakable,
  boxed title style={
    outer arc=0pt,
    arc=0pt,
    top=0pt,
    bottom=0pt,
    },
  }
}

\makeatletter
\def\@LN@depthbox{%
  \ifdim\@tempdima = -1000pt
  \else
    \dp\@tempboxa=\@tempdima
    \nointerlineskip \kern-\@tempdima 
  \fi
  \box\@tempboxa
  } 
\makeatother

\iffancyversion

\newtcbtheorem[crefname={theorem}{theorems},Crefname={Theorem}{Theorems},number within=section]{ttheoremb}{Theorem}{bluewestlinestyle}{r}
\newtcbtheorem[crefname={lemma}{lemmas},Crefname={Lemma}{Lemmas},use counter from=ttheoremb]{tlemmab}{Lemma}{bluewestlinestyle}{r}
\newtcbtheorem[crefname={corollary}{corollaries},Crefname={Corollary}{Corollaries},use counter from=ttheoremb]{tcorollaryb}{Corollary}{bluewestlinestyle}{r}
\newtcbtheorem[crefname={proof}{proofs},Crefname={Proof}{Proofs},number within=section]{tproof}{Proof}{redvioletwestlinestyle}{p}
\newtcbtheorem[crefname={proposition}{propositions},Crefname={Proposition}{Propositions},use counter from=ttheoremb]{tprop}{Proposition}{bluewestlinestyle}{r}
\newtcbtheorem[crefname={assumption}{assumptions},Crefname={Assumption}{Assumptions},use counter from=ttheoremb]{tassumption}{Assumption}{redvioletwestlinestyle}{a}
\newtcbtheorem[crefname={remark}{remarks},Crefname={Remark}{Remarks},number within=section, ]{tremark}{Remark}{greenwestlinestyle}{rm}
\newtcbtheorem[crefname={example}{examples},Crefname={Example}{Examples},number within=section,]{texample}{Example}{greenwestlinestyle}{ex}
\newtcbtheorem[crefname={definition}{definitions},Crefname={Definition}{Definitions},number within=section]{tdefinition}{Definition}{bluewestlinestyle}{d}

\else

\newtcbtheorem[crefname={theorem}{theorems},Crefname={Theorem}{Theorems},number within=section]{ttheoremb}{Theorem}{defaultthmstyle}{r}
\newtcbtheorem[crefname={lemma}{lemmas},Crefname={Lemma}{Lemmas},use counter from=ttheoremb]{tlemmab}{Lemma}{defaultthmstyle}{r}
\newtcbtheorem[crefname={corollary}{corollaries},Crefname={Corollary}{Corollaries},use counter from=ttheoremb]{tcorollaryb}{Corollary}{defaultthmstyle}{r}
\newtcbtheorem[crefname={proposition}{propositions},Crefname={Proposition}{Propositions},use counter from=ttheoremb]{tprop}{Proposition}{defaultthmstyle}{r}
\newtcbtheorem[crefname={proof}{proofs},Crefname={Proof}{Proofs},number within=section]{tproof}{Proof}{defaultproofstyle}{p}
\newtcbtheorem[crefname={remark}{remarks},Crefname={Remark}{Remarks},number within=section, ]{tremark}{Remark}{defaultthmstyle}{rm}
\newtcbtheorem[crefname={assumption}{assumptions},Crefname={Assumption}{Assumptions},use counter from=ttheoremb, ]{tassumption}{Assumption}{compactdefaultstyle}{a}
\newtcbtheorem[crefname={example}{examples},Crefname={Example}{Examples},number within=section,]{texample}{Example}{compactdefaultstyle}{ex}
\newtcbtheorem[crefname={definition}{definitions},Crefname={Definition}{Definitions},number within=section]{tdefinition}{Definition}{compactdefaultstyle}{d}

\fi

\makeatletter
\newcommand\linktoproofifdef[1]{%
  \@ifundefined{r@#1}{}{%
\null\hfill\hyperref[#1]{\color{customBlue}{($→$ Proof)}}%
  }%
}
\makeatother

\newif\ifnolinkoption
\NewDocumentEnvironment{tlemma}{omm}{
\IfValueTF{#1}{\IfSubStr{#1}{n}{\nolinkoptiontrue}{\nolinkoptionfalse}
}{ \nolinkoptionfalse }
\begin{tlemmab}[%
	after upper={\ifnolinkoption{}\else{\linktoproofifdef{p:r:#3}}\fi},
	]{#2}{#3}	
}{
\end{tlemmab}
}
\NewDocumentEnvironment{ttheorem}{omm}{
\IfValueTF{#1}{\IfSubStr{#1}{n}{\nolinkoptiontrue}{\nolinkoptionfalse}
}{ \nolinkoptionfalse }
\begin{ttheoremb}[%
	after upper={\ifnolinkoption{}\else{\linktoproofifdef{p:r:#3}}\fi},
	]{#2}{#3}	
}{
\end{ttheoremb}
}
\NewDocumentEnvironment{tcorollary}{omm}{
\IfValueTF{#1}{\IfSubStr{#1}{n}{\nolinkoptiontrue}{\nolinkoptionfalse}
}{ \nolinkoptionfalse }
\begin{tcorollaryb}[%
	after upper={\ifnolinkoption{}\else{\linktoproofifdef{p:r:#3}}\fi},
	]{#2}{#3}	
}{
\end{tcorollaryb}
}

\newenvironment{tproofof*}[1]{
\begin{tproof}[title={Proof of \Cref{#1}:}]{}{#1}
}{
\end{tproof}
}

\usepackage{iftex}
\usepackage{xspace}

\newcommand\de{:=}
\newcommand\eg{e.g.\xspace }
\newcommand\ie{i.e.\xspace }
\newcommand\aka{a.k.a.\xspace }
\newcommand\wrt{w.r.t.\xspace }

\newcommand{\cred}[1]{{\color{red}#1}}

\newcommand{\tiid}{i.i.d.\ }

\newcommand\numberthis{\addtocounter{equation}{1}\tag{\theequation}}

\newcommand\restr[2]{{
  \left.\kern-\nulldelimiterspace 
  #1 
  \vphantom{\big|} 
  \right|_{#2} 
  }}

\DeclareMathOperator*{\argmin}{arg\,min}
\DeclareMathOperator*{\argmax}{arg\,max}

\DeclareMathOperator*{\polylog}{polylog}

\DeclareMathOperator*{\esssup}{ess\,sup}

\newcommand{\Var}{\text{Var}}

\DeclareMathOperator*{\Tr}{tr}
\DeclareMathOperator*{\spa}{span}

\ifLuaTeX
	\newcommand{\V}[1]{\symbf{#1}} 
\else
	\usepackage{bm}
	\newcommand{\V}[1]{\bm{#1}} 
\fi

\DeclareMathOperator*{\ran}{ran}

\newcommand{\kron}{\otimes}

\newcommand{\bN}{\mathbb{N}}

\newcommand{\bR}{\mathbb{R}}

\newcommand{\E}{\mathbf{E}}

\DeclarePairedDelimiter{\prt}{(}{)}
\DeclarePairedDelimiter{\brk}{[}{]}
\DeclarePairedDelimiter{\cb}{\{}{\}}
\let\norm\relax
\DeclarePairedDelimiter{\norm}{\lVert}{\rVert}
\DeclarePairedDelimiter{\n}{\lVert}{\rVert}
\DeclarePairedDelimiter{\ip}{\langle}{\rangle}
\DeclarePairedDelimiter{\absv}{|}{|}

\newcommand\thickmidrule{\midrule[0.15em]}
\newcommand\thicktoprule{\toprule[0.2em]}
\newcommand\thickbottomrule{\bottomrule[0.2em]}

\usepackage{pgffor}
\foreach \x in {a,...,z}{
  \expandafter\xdef\csname V\x \endcsname{\noexpand\ensuremath{\noexpand\V{\x}}}
}
\foreach \x in {A,...,Z}{
  \expandafter\xdef\csname V\x \endcsname{\noexpand\ensuremath{\noexpand\V{\x}}}
  \expandafter\xdef\csname c\x \endcsname{\noexpand\ensuremath{\noexpand\mathcal{\x}}}
  \expandafter\xdef\csname f\x \endcsname{\noexpand\ensuremath{\noexpand\mathfrak{\x}}}
}

\NewDocumentCommand\inputpgf{O{.}m}{
\let\pgfimageWithoutPath\pgfimage
\renewcommand{\pgfimage}[2][]{\pgfimageWithoutPath[##1]{#1/##2}}
\let\includegraphicsWithoutPath\includegraphics
\renewcommand{\includegraphics}[2][]{\includegraphicsWithoutPath[##1]{#1/##2}}
\begingroup\renewcommand\sffamily{}\input{#1/#2}\endgroup
}

\usepackage{pifont}
\newcommand{\xmark}{\ding{55}}%

\newcommand{\lcba}[1]{\left\{\begin{aligned}#1\end{aligned}\right.}

\DeclareUnicodeCharacter{03B1}{\alpha}  	 
\DeclareUnicodeCharacter{0391}{\Alpha}       
\DeclareUnicodeCharacter{03B2}{\beta}        
\DeclareUnicodeCharacter{0392}{\Beta}        
\DeclareUnicodeCharacter{03B3}{\gamma}       
\DeclareUnicodeCharacter{0393}{\Gamma}       
\DeclareUnicodeCharacter{03B4}{\delta}       
\DeclareUnicodeCharacter{0394}{\Delta}       
\DeclareUnicodeCharacter{03B5}{\epsilon}     
\DeclareUnicodeCharacter{0395}{\Epsilon}     
\DeclareUnicodeCharacter{03B6}{\zeta}        
\DeclareUnicodeCharacter{0396}{\Zeta}        
\DeclareUnicodeCharacter{03B7}{\eta}         
\DeclareUnicodeCharacter{0397}{\Eta}         
\DeclareUnicodeCharacter{03B8}{\theta}       
\DeclareUnicodeCharacter{0398}{\Theta}       
\DeclareUnicodeCharacter{03B9}{\iota}        
\DeclareUnicodeCharacter{0399}{\Iota}        
\DeclareUnicodeCharacter{03BA}{\kappa}       
\DeclareUnicodeCharacter{039A}{\Kappa}       
\DeclareUnicodeCharacter{03BB}{\lambda}      
\DeclareUnicodeCharacter{039B}{\Lambda}      
\DeclareUnicodeCharacter{03BC}{\mu}          
\DeclareUnicodeCharacter{039C}{\Mu}          
\DeclareUnicodeCharacter{03BD}{\nu}          
\DeclareUnicodeCharacter{039D}{\Nu}          
\DeclareUnicodeCharacter{03BE}{\xi}          
\DeclareUnicodeCharacter{039E}{\Xi}          
\DeclareUnicodeCharacter{03BF}{\omega}       
\DeclareUnicodeCharacter{039F}{\Omega}       
\DeclareUnicodeCharacter{03C0}{\pi}          
\DeclareUnicodeCharacter{03A0}{\Pi}          
\DeclareUnicodeCharacter{03C1}{\rho}         
\DeclareUnicodeCharacter{03A1}{\Rho}         
\DeclareUnicodeCharacter{03C3}{\sigma}       
\DeclareUnicodeCharacter{03A3}{\Sigma}       
\DeclareUnicodeCharacter{03C4}{\tau}         
\DeclareUnicodeCharacter{03A4}{\Tau}         
\DeclareUnicodeCharacter{03C5}{\upsilon}     
\DeclareUnicodeCharacter{03A5}{\Upsilon}     
\DeclareUnicodeCharacter{03D5}{\phi}         
\DeclareUnicodeCharacter{03C6}{\varphi}      
\DeclareUnicodeCharacter{03A6}{\Phi}         
\DeclareUnicodeCharacter{03C7}{\xi}          
\DeclareUnicodeCharacter{03A7}{\Xi}          
\DeclareUnicodeCharacter{03C8}{\psi}         
\DeclareUnicodeCharacter{03A8}{\Psi}         
\DeclareUnicodeCharacter{03C9}{\omega}       
\DeclareUnicodeCharacter{03A9}{\Omega}       
\DeclareUnicodeCharacter{1D62}{_i}       	 %

\DeclareUnicodeCharacter{2248}{\approx}      
\DeclareUnicodeCharacter{2264}{\leq}         
\DeclareUnicodeCharacter{2265}{\geq}         
\DeclareUnicodeCharacter{227C}{\preccurlyeq} 
\DeclareUnicodeCharacter{227D}{\succcurlyeq} 
\DeclareUnicodeCharacter{2286}{\subseteq}    
\DeclareUnicodeCharacter{2260}{\neq}   		 
\DeclareUnicodeCharacter{1D40}{^T}           
\DeclareUnicodeCharacter{2203}{\exists}      
\DeclareUnicodeCharacter{2200}{\forall}      
\DeclareUnicodeCharacter{220A}{\in}          
\DeclareUnicodeCharacter{00B7}{\cdot}        
\DeclareUnicodeCharacter{00D7}{\times}       
\DeclareUnicodeCharacter{00B7}{\cdot}        
\DeclareUnicodeCharacter{2016}{\Vert}        
\DeclareUnicodeCharacter{2026}{\dots}        
\DeclareUnicodeCharacter{2020}{\dagger}      
\DeclareUnicodeCharacter{27C2}{\perp}        %
\DeclareUnicodeCharacter{222A}{\cup}        
\DeclareUnicodeCharacter{00BD}{\nicefrac{1}{2}} 
\DeclareUnicodeCharacter{2190}{\leftarrow}   
\DeclareUnicodeCharacter{2192}{\rightarrow}  
\DeclareUnicodeCharacter{2194}{\leftrightarrow} %
\DeclareUnicodeCharacter{21A6}{\mapsto}       
\DeclareUnicodeCharacter{21D4}{\iff} 		 
\DeclareUnicodeCharacter{211D}{\mathbb{R}}   
\DeclareUnicodeCharacter{221E}{\infty}   
\DeclareUnicodeCharacter{00BD}{\nicefrac{1}{2}} 
\DeclareUnicodeCharacter{00B2}{^2}           
\DeclareUnicodeCharacter{00B3}{^3}           
\DeclareUnicodeCharacter{2074}{^4}           
\DeclareUnicodeCharacter{2075}{^5}           
\DeclareUnicodeCharacter{2076}{^6}           
\DeclareUnicodeCharacter{2077}{^7}           
\DeclareUnicodeCharacter{2078}{^8}           
\DeclareUnicodeCharacter{2079}{^9}           
\DeclareUnicodeCharacter{1D62}{_i}           %
\DeclareUnicodeCharacter{2C7C}{_j}           %
\DeclareUnicodeCharacter{2096}{_k}           %
\DeclareUnicodeCharacter{2097}{_l}           %
\DeclareUnicodeCharacter{2080}{_0}           
\DeclareUnicodeCharacter{2081}{_1}           
\DeclareUnicodeCharacter{2082}{_2}           
\DeclareUnicodeCharacter{2083}{_3}           
\DeclareUnicodeCharacter{2084}{_4}           
\DeclareUnicodeCharacter{2085}{_5}           
\DeclareUnicodeCharacter{2086}{_6}           
\DeclareUnicodeCharacter{2087}{_7}           
\DeclareUnicodeCharacter{2088}{_8}           
\DeclareUnicodeCharacter{2089}{_9}           

\DeclareUnicodeCharacter{2212}{-}            
\DeclareUnicodeCharacter{2013}{--}           
\DeclareUnicodeCharacter{2014}{---}          
\DeclareUnicodeCharacter{202F}{~}            
\DeclareUnicodeCharacter{00A0}{~}            

\newcommand\deff[1][λ]{d_{\textup{eff}}\prt*{#1}}
\newcommand\dsup[1][λ]{d_{∞}(#1)}
\NewDocumentCommand\pfun{O{t}}{p\IfNoValueF{#1}{_{#1}}}
\newcommand\sqpfunc{\textup{powfun}²}	
\newcommand\res{r}
\newcommand\resc{r}	
\DeclareMathOperator*{\zeros}{zeros}

\newcommand{\rkhs}{\cH}		
\newcommand{\fclass}{\cH}
\newcommand\dspace{\cX}
\newcommand\sob[1][\dspace]{\textup{H}^s(#1)}
\newcommand{\fs}[1][s]{\cH_{\td}^{#1}} 	

\newcommand\td{ρ}	

\DeclarePairedDelimiter{\iprkhs}{\langle}{\rangle}		

\DeclarePairedDelimiter{\nrkhs}{\lVert}{\rVert}		
\DeclarePairedDelimiter{\noprkhs}{\lVert}{\rVert_{\text{\cL(\rkhs)}}}		

\newcommand{\ltsp}[1][\td]{L^2(#1)}

\DeclarePairedDelimiter{\nltsp}{\lVert}{\rVert_{\ltsp}}		

\NewDocumentCommand\fmap{g}{ϕ\IfNoValueF{#1}{(#1)}}
\NewDocumentCommand\fmapF{g}{ϕ_{\fclass}\IfNoValueF{#1}{(#1)}}
\NewDocumentCommand\kemb{g}{ϕ\IfNoValueF{#1}{(#1)}}

\newcommand{\covop}{C}									
\newcommand{\covopl}{\covop_\lambda}					
\newcommand{\ecovop}{\hat{\covop}_n}					
	
\newcommand\Pm{P_m}
\newcommand{\Pmo}{P_m^\perp}	

\newcommand\pcite[2][]{\parencite[#1]{#2}}
\newcommand{\data}{X} 
\newcommand{\ldms}{\tilde{X}} 
\DeclareDocumentCommand\ldm{g}{\tilde{X}\IfNoValueF{#1}{_{#1}}}
\DeclareDocumentCommand\aerr{O{f}O{\nI}}{\textup{E}(#1,#2)}
\newcommand\WCE{\cE}
\DeclareDocumentCommand\wce{O{\cH}O{\nI}}{\WCE(#1,#2)}
\DeclareDocumentCommand\wceH{O{\nI}}{\wce[\rkhs][#1]}
\DeclareDocumentCommand\wceS{O{\cH}}{\WCE(#1)}	
\DeclareDocumentCommand\mmaerr{O{\cH}}{\textup{E}_m(#1)}

\DeclareDocumentCommand\I{}{\textup{I}}
\DeclareDocumentCommand\eI{}{\hat{\textup{I}}}
\DeclareDocumentCommand\nI{}{\textup{I}_{\ldms,w}}

\newcommand{\rkhsm}{\cH_m}	

\NewDocumentCommand\kmat{O{n}}{K_{#1}}
\newcommand{\kmatm}{K_m}
\newcommand{\kmatmn}{K_{mn}}

\newcommand{\ed}{\hat{\rho}_n}		
\DeclareDocumentCommand\ldmd{g}{p\IfNoValueF{#1}{_{#1}}} 
\newcommand{\pldm}[1]{\tilde{\ldmd{i}}}

\newcommand{\mftldms}{Φ_m}	 

\newcommand{\mftx}{Φ_n}	 	
\NewDocumentCommand\mft{O{}}{\Phi\IfNoValueF{#1}{_{#1}}}		
\NewDocumentCommand\amft{O{}}{\Phi\IfNoValueF{#1}{_{#1}}^*} 	

\newcommand{\kr}{\kappa} 
\NewDocumentCommand\me{O{}}{μ\IfNoValueF{#1}{^{#1}}} 			
\NewDocumentCommand\eme{O{}}{\hat{μ}_n\IfNoValueF{#1}{^{#1}}} 	
\newcommand{\emem}{\hat{\mu}_m}								
\NewDocumentCommand\nyseme{O{}}{\tilde{μ}_m\IfNoValueF{#1}{^{#1}}} 	
\newcommand{\supfmap}{K}
\newcommand{\supk}{\supfmap^2 }
\newcommand{\cker}{c_{\kr}}
\DeclareDocumentCommand\als{g}{\hat{\ell}_λ\IfNoValueF{#1}{(#1)}} 
\DeclareDocumentCommand\tls{g}{\ell_λ\IfNoValueF{#1}{(#1)}} 

\colorlet{tablerowsbg}{SeaGreen!20!}

\usepackage{enumitem} 
\setitemize{noitemsep,topsep=2pt,parsep=2pt,partopsep=0pt}
\setenumerate{noitemsep,topsep=2pt,parsep=2pt,partopsep=0pt}

\newcommand{\ACc}[1]{\textup{\color{magenta}[Antoine: #1]}}
\newcommand\AC\ACc	

\newcommand{\new}[1]{#1}

\begin{document}

\maketitle


\paragraph{Abstract}
In this work we consider the problem of numerical integration, \ie, approximating integrals with respect to a target probability measure using only pointwise evaluations of the integrand.
We focus on the setting in which the target distribution is only accessible through a set of $n$ \tiid observations, 
and the integrand belongs to a reproducing kernel Hilbert space. 
We propose an efficient procedure which exploits a small \tiid random subset of $m<n$ samples drawn either uniformly or using approximate leverage scores from the initial observations.
Our main result is an upper bound on the approximation error of this procedure for both sampling strategies. 
It yields sufficient conditions on the subsample size to recover the standard (optimal) $n^{−1/2}$ rate while reducing drastically the number of functions evaluations---and thus the overall computational cost.
Moreover, we obtain rates with respect to the number $m$ of evaluations of the integrand which adapt to its smoothness, and match known optimal rates for instance for Sobolev spaces.
We illustrate our theoretical findings with numerical experiments on real datasets, which highlight the attractive efficiency-accuracy tradeoff of our method compared to existing randomized and greedy quadrature methods.
We note that, the problem of numerical integration in RKHS amounts to designing a discrete approximation of the kernel mean embedding of the target distribution.
As a consequence, direct applications of our results also include the efficient computation of maximum mean discrepancies between distributions and the design of efficient kernel-based tests.  



\newpage
\tableofcontents
\newpage


\color{black}
\section{Introduction}
\label{s:quadratures}

Numerical integration is a key tool in applied mathematics and physics \pcite{davis2007methods}. It is particularly useful for approximating integrals that cannot be computed in closed form---for instance when the integrand depends on some data and does not have a simple analytical expression.
It is used extensively in Bayesian inference \pcite{gelman1995bayesian} 
as well as for the resolution of PDEs \pcite{quarteroni2008NumericalApproximationPartial}, \eg for the computation of the entries of the stiffness matrix used in finite elements methods, or in deep-learning-based approaches to estimate the loss function which is typically derived from a variational formulation of the problem~\pcite{rivera2022QuadratureRulesSolving}.
Quadrature techniques are also commonly used in statistical physics for the computation of free energies, where one typically needs to integrate over large state spaces \pcite{newman1999monte}.



Now we provide a formal definition of the problem. 
Let \new{$(\dspace, \cB, \td)$ be a measurable space,} 
and let $(\fclass,\n{·})$ be a normed vector space of functions defined over $\cX$. 
We consider the problem of designing quadrature rules for functions in $\fclass$ with respect to a probability measure $\td$. 
More precisely, we search for points $\ldms\de (\ldm{1},…,\ldm{m}) \in \dspace^m$ (called the nodes or landmark points) and weights $w=[w₁,…, w_m]ᵀ \in \bR^m$ such that, for any function $f$ in the unit ball of $\fclass$, the integral
\begin{align}
	\I(f) &\de \int f(x)\dif\td(x) 
	\label{e:def_int}
\end{align}
is well approximated by the quadrature rule defined by the weighted sum of pointwise evaluations
\begin{align}
	\nI(f) &\de \sum_{j=1}^{m} w_i f(\ldm{i}).
	\label{e:def_nI}
\end{align}
Importantly, the weights $(w_i)_{1≤i≤m}$ can depend on the nodes $\ldm$, but not on the integrand $f \in \mathcal{H}$. 
Moreover, we will consider the general setting in which the weights $w$ are not required to be positive nor to sum to one, albeit some methods in the literature have been developed in order to satisfy such additional constraints, see for instance the work by \textcite{hayakawa2022PositivelyWeightedKernel}. 
To quantify the performance of a given quadrature rule $\nI$, we define its approximation error as the worst-case error over the unit ball in $\fclass$, 
\begin{align}
	\wce &\de \sup_{f∊\fclass: \n{f}≤1} 
	\absv*{\I(f) - \nI(f)} \enspace.
	\label{e:wce}
\end{align}
We use the shorter notation $\wceS$ when the quadrature rule $\nI$ is clear from the context.

\paragraph{Quadratures from empirical data} We assume to have at our disposal a dataset of $n$ \tiid samples $\data=\cb*{X₁, …,X_n}$. 
%
A natural estimator of $\I(f)$ is the Monte-Carlo estimator
\begin{align}
	\eI(f) 
	&\de \frac{1}{n} \sum_{i=1}^n f(X_i), 
	\label{e:empirical_quad}
\end{align}
which estimates $\I(f)$ uniformly over the unit ball of $\ltsp$ at the rate $O(1/\sqrt{n})$ with high probability. 
The complexity for computing this estimator grows linearly with the number $n$ of samples in the dataset.  We will be interested by applications in which one can easily obtain \tiid samples from a target probability distribution, but pointwise evaluation of the integrand can be expensive.
\new{The need for approximating integrals of functions that are expensive to evaluate is a common problem which appears across many application fields, we refer for instance the reader to \textcite{oates2017ProbabilisticModelsIntegration} for an application to a computational cardiac model where each integrand evaluation takes about 100 CPU hours. 
It should be noted however that sampling from the target probability distribution may sometimes also be a major hurdle, in which case strategies such as Markov chain Monte Carlo (MCMC) or density estimation may be used~\pcite{oates2017ProbabilisticModelsIntegration,delyon2016IntegralApproximationKernel}.}

\paragraph{Objective}
Given a dataset $X$ of $n$ \tiid samples, our goal is to design a quadrature rule of the form \eqref{e:def_nI} that (i) is computed using the knowledge of the $n$ samples $\data$ and yet (ii) is supported on only $m<n$ nodes, while (iii) achieving the same finite-sample rate as the Monte-Carlo estimator~$\eI$. %
%
We will show that these requirements are not incompatible, and that computational efficiency can be improved without sacrificing statistical accuracy.

%
We consider in particular the setting in which we first sample the nodes $(\ldm{j})_{1≤j≤m}$ from the dataset $\data$ 
(so that the approximation bounds must hold with high probability on the draw of these points), 
and then set the weights $w=(w_j)_{1≤j≤m}$ deterministically.

\subsection{\new{Quadratures in Reproducing Kernel Hilbert Spaces}}
\label{s:kquadratures}

In this paper, we consider the setting in which $\fclass$ is a reproducing kernel Hilbert space (RKHS) of functions over $\dspace$ with reproducing kernel $\kr$ \pcite{aronszajn1950TheoryReproducing}. 
Such spaces encompass many typical smoothness spaces considered in the learning literature.
For instance, Sobolev spaces of high enough smoothness are RKHS, as reminded in the following example.
\begin{texample}{Sobolev}{sobolev}
	Let $s∊\bN$. 
	If $\dspace=\bR^d$, 
	denoting $\hat{f}$ the Fourier transform of $f$, 
	the Sobolev space \new{$\sob[\bR^d]$} is defined as 
	\begin{align*}
		\sob[\bR^d]
		\de \Set{f∊L^2(\bR^d)| \prt*{\int_{\cX} (1+|ξ|²)^s |\hat{f}\new{(ξ)}|² \dif ξ}^{1/2}=:\n{f}<∞}\enspace.
	\end{align*}
	When the smoothness parameter is high enough, namely $s>d/2$, \new{$\sob[\bR^d]$} is \new{an} RKHS. 
	\new{For any non-empty domain $Ω⊆ℝ^d$, we define $\sob[Ω]$ as the RKHS induced by the restriction of the reproducing kernel of $\sob[\bR^d]$ to $Ω×Ω$.}
\end{texample}
\new{When $Ω$ has Lipschitz boundary, the above definition is known to be norm-equivalent to the alternative definition of Sobolev spaces involving weak derivatives~\pcite[Corollary 10.48]{wendland2004ScatteredDataApproximation}. }
When $s>d/2$, it has been shown by \textcite[Section 1.3.12 Proposition 3]{novak1988DeterministicStochasticError} that the optimal rate for a deterministic quadrature rule of the form \eqref{e:def_nI} \new{on the unit hypercube} is $\inf_{\ldm,w}\wce[\new{\textup{H}^s([0,1]^d)}][\nI]=Θ(m^{-s/d})$, which suggests that the Monte-Carlo estimator (with $m=n$) might not be optimal in this setting as it gives the rate $m^{-1/2}$; \new{see also \textcite{novak2006FunctionSpacesLipschitz} for similar results on more general domains and the Lebesgue measure.}
The optimal rate can be reached in practice 
(see for instance the work by \textcite{briol2019ProbabilisticIntegrationRole} for quadrature rules based on Markov chain Monte-Carlo%
, by \textcite{santin2022SamplingBasedApproximation} for greedy methods), and our goal is indeed to design quadrature rules that have this adaptivity to the smoothness of the considered RKHS in order to 
reduce the cost of numerical integration.

While the parameter $s$ provides a direct control on the smoothness in the Sobolev setting, 
in this paper we develop a more generic analysis which depends on the decay of the spectrum of the integral operator associated to the reproducing kernel $\kr$ and the target distribution $\td$.

\paragraph{Existing approaches}
Although we postpone to \Cref{s:related_work} the presentation of related works, we provide here a preliminary overview of the different approaches which have been proposed to 
\new{design quadrature rules in reproducing kernel Hilbert spaces.}
Our method belongs to the family of random designs, obtained by sampling randomly and simultaneously the quadrature nodes; this includes \tiid uniform and importance sampling~\pcite{bach2017EquivalenceKernelQuadrature}, as well as non-\tiid sampling strategies~\pcite{belhadji2019KernelQuadratureDPPs}. 
Multiple greedy methods exist to iteratively select the nodes, typically by minimizing some notion of residual, or by filling the space as uniformly as possible~\pcite{briol2019ProbabilisticIntegrationRole}.
In the literature on core-sets, multiple algorithms have been proposed to compress the set of $n$ samples down to $m$ points using \eg recursive halving approaches~\pcite{dwivedi2022GeneralizedKernelThinning}.
Note that some of the methods can be declined in both deterministic and randomized variants, which makes it difficult to provide a clear classification of the literature.

Although formulated in the context of numerical integration in RKHS, our bounds can also be interpreted as approximation bounds for the computation of mean embeddings in reproducing kernel Hilbert spaces.
\begin{tremark}{Kernel Mean Embedding and Maximum Mean Discrepancy}{kme}
	Any \new{quadrature rule in a RKHS} can be interpreted as a way to approximate the so-called kernel mean embedding $μ\de \int \kr(x,·)\dif \td(x)∊\rkhs$ of the probability distribution~$\td$. Indeed, when $\rkhs$ is \new{an} RKHS it holds $f(x)=\ip{f,\kr(x,·)}$ for any $f∊\rkhs,x∊\dspace$, and thus $\I(f)=\ip{f,μ}$. This connection will be introduced and discussed in \Cref{s:theory}. It implies in particular that our work directly translates to algorithms and bounds for the efficient approximation of the maximum mean discrepancy, a standard metric between probability distributions in the context of kernel methods. The maximum mean discrepancy between two distributions indeed corresponds to the distance between their kernel mean embeddings.
\end{tremark}

\subsection{Summary of Contributions}
\label{ss:summary_contributions}

This paper builds on the results by \textcite{chatalic2022NystromKernelMean}, 
that study kernel mean embeddings (see \Cref{rm:kme}) 
obtained by uniformly sampling the nodes.
Our main contributions are the following:
\begin{itemize}
	\item We introduce a quadrature rule whose nodes are randomly subsampled from the dataset $\data$ either uniformly or using leverage scores, and whose weights are optimally chosen by solving a least-square problem. 
	This extends in particular the setting considered in \pcite{chatalic2022NystromKernelMean}, which covers only uniform sampling.
	\item We provide high-probability bounds on the worst-case error of this quadrature rule, and obtain quantization rates (\ie \wrt the number of nodes $m$) which are faster than the Monte-Carlo rate. For leverage score sampling, we obtain in particular asymptotic rates that match known optimal rates for Sobolev spaces \pcite{novak1988DeterministicStochasticError}.
	\item We show that our method adapts to the smoothness of the integrand by showing that faster rates can be derived for fractional subspaces of $\rkhs$, \ie assuming a source condition on the integrand \pcite{engl2000RegularizationInverseProblems}.
	\item We compare empirically our method to other randomized and greedy approaches from the literature on real datasets, and show that our approach has a particularly interesting efficiency-accuracy tradeoff. 
\end{itemize}

\paragraph{Layout}
The rest of the paper is organized as follows.
We introduce our algorithm in \Cref{s:nystrom}. In \Cref{s:main_results}, we summarize our main hypotheses and theoretical results, and put them in perspective by reviewing the state of the art.
Leveraging tools from kernel methods, 
in \Cref{s:theory} we detail how our bounds on the worst-case error are derived for both uniform and leverage scores sampling. 
We then compare experimentally our method with other quadrature approaches in \Cref{s:experiments}. 
A table of notations is provided in \Cref{s:table_notations}.

\section{Two Algorithms Based on Subsampling}
\label{s:nystrom}

In this section, we describe the method we will analyze in the rest of the paper. 
It corresponds to a quadrature rule of the type \eqref{e:def_nI} with randomly sampled nodes $(\ldm{j})_{1≤j≤m}$ (\Cref{ss:sampling_schemes}) and weights $w$ obtained by solving an unconstrained least-squares problem (\Cref{s:weights}).

\subsection{Choice of the Nodes}
\label{ss:sampling_schemes}

We consider in the following two strategies for sampling the nodes $\ldms$ from the empirical data $\data$: uniform sampling and (ridge) leverage score sampling. 


\paragraph{Uniform sampling} 
The nodes $\ldms=\cb{\ldm{1},…,\ldm{m}}$ are sampled uniformly from the set of all subsets of cardinality $m$ of 
$\cb{X₁,…,X_n}$. This is the most intuitive sampling strategy and arguably the easiest to implement. It will serve as a baseline against leverage scores sampling.

\paragraph{Approximate Ridge Leverage Score sampling (ARLS)} Ridge leverage scores have been introduced by \textcite{alaoui2015FastRandomizedKernel} in the setting of kernel ridge regression. They are related to the more general notion of statistical leverage score \pcite{mahoney2009CURMatrixDecompositions}. We now provide a formal definition.
\begin{tdefinition}{Ridge leverage scores}{}
Given $n \geq 1$ data points $X_1, \dots, X_n$, let $\kmat∊\bR^{n×n}$ denote the kernel matrix with entries $(\kmat)_{i,j} = \kr(X_i, X_j)$ for all $i,j \in [n]$.  Let $\lambda >0$. For any $i \in [n]$, the ridge leverage score of the datapoint $X_i$ is defined as
\begin{equation}
    \tls(i) \coloneqq \prt*{ \kmat(\kmat + λn I)^{-1} }_{ii}.
	\label{e:def_ls}
\end{equation}
\end{tdefinition}
\noindent
Such scores can be interpreted as a measure of the relative importance of each point in the dataset. They are directly related to Christoffel functions~\pcite{fanuel2022NystromLandmarkSampling,pauwels2018RelatingLeverageScores}.
The cost of exactly computing leverage scores quickly becomes prohibitive as the sample size grows due to the matrix inversion. Since the purpose of our approach is to reduce computational cost, we will rely on a approximate notion that has been studied in the literature. 
\begin{tdefinition}{ARLS}{leverage_scores}
Let $δ \in (0,1]$, $λ_0 > 0$ and $z \in [1 , \infty)$. 
A set $(\als(i))_{i\in[n]}$ is said to be $(z,λ_0,δ)$-approximate ridge leverage scores (ARLS) of $\data$ if it satisfies with probability at least $1-δ$,
\begin{equation}
    \frac{1}{z}~\new{\tls(i)} \leq \new{\als(i)}\leq z~\new{\tls(i)}, \qquad \forall λ\geq λ_0, \forall i \in [n].
	\label{e:def_als}
\end{equation}
\end{tdefinition}

\noindent
Different algorithms have been proposed in the literature to obtain approximate ridge leverage scores. In this work we use BLESS \pcite{rudi2018FastLeverageScore}. It is based on a coarse-to-fine strategy with a computational cost of order $O(\deff²/λ)$, where $\deff$ denotes the effective dimension defined in the next section. 
After computing the values $\new{\als(i)}$, the landmarks $\ldms$ are drawn with replacement from $\data$ 
proportionally to $\new{\als(i)}$. 
We refer in the following to this method as ARLS sampling.

\subsection{Choice of the Weights}
\label{s:weights}

%
Once the landmarks $\ldm$ are selected, the weights are chosen as
\begin{align}
	w^* &= \min_{w∊\bR^m} \sup_{f∊\rkhs:\n{f}≤1} |\eI(f) - \nI(f)|.
	\label{e:def_weights}
\end{align}
This problem is a least squares problem and  
 our estimator can be computed using the closed form $w =\frac{1}{n} \kmatm^+ \kmatmn 1_n$ where $A^+$ denotes the Moore-Penrose pseudo-inverse of $A$, $\kmatm\in \bR^{m\times m}$ and $\kmatmn\in \bR^{m\times n}$ denote the kernel matrices with entries $(\kmatm)_{ij}=\kr(\ldm{i},\ldm{j})$ for any $1\leq i,j\leq m$ and $(\kmatmn)_{ij}=\kr(\ldm{i},x_j)$ for any $1\leq i\leq m$ and $1\leq j\leq n$, and $1_n$ denotes a $n$-dimensional vector of ones.
We refer the reader to \Cref{s:computation_weights} for a precise derivation of this expression.
%
%

We will see in \Cref{s:theory} that the quadrature rule built using subsampling and optimal weights \eqref{e:def_weights} is closely related to the so-called Nyström approximation. 
The latter is a standard way to approximate kernel matrices by rows/columns subsampling in the machine learning literature \pcite{williams2001UsingNystromMethod}, 
but actually takes its name from the work by \textcite{nystroem1930UeberPraktischeAufloesung} to discretize linear integral equations, see also \pcite[Sec.~12.2]{kress2014LinearIntegralEquations}. 
In this work, we thus use this designation in a broad sense: the subsampling procedure for the selection of nodes follows the literature on low-rank approximations of kernel matrices, however what we care about is the approximation of a linear operator, and thus the bounds we derive differ from what is usually done in the machine learning literature (see also \Cref{rm:kmat} in this regard). 

\paragraph{Complexity} 
The space complexity of the method (excluding the sampling phase) is $Θ(m^2 +md)$ for storing $\kmatm$ and the nodes. Note that $\kmatmn$ does not need to be stored as $\kmatmn 1_n$ can be computed sequentially in $Θ(m)$ space.
The time complexity (still excluding sampling) is $Θ(nm\cker+m^3 )$ where $\cker$ corresponds to the cost of a kernel evaluation. The first term corresponds to the computation of $\kmatmn 1_n$ while the second correspond to computing the pseudo-inverse of $\kmatm$ (numerically stable algorithms can be used instead, but the complexity will be of this order regardless).
When $\dspace ⊆\bR^d$, many standard kernel functions come with an evaluation cost which is of the order of the dimension, \ie $\cker=d$.

\section{Main Results}
\label{s:main_results}

We detail our technical assumptions in \Cref{s:assumptions}, and give an overview of our main results in \Cref{s:main_rates}.
We then put our results in perspective by reviewing the state of the art in \Cref{s:related_work}.

\subsection{Assumptions}
\label{s:assumptions}

We \new{recall that $(\dspace, \cB, \td)$ is a probability space,} where 
$\td$ is the data probability distribution. 
\begin{tassumption}{Independent and identically distributed samples}{iid}
	We have access to $n$ data points $X_1, \dots, X_n$, drawn \tiid from the probability distribution $\td$.
\end{tassumption}
The first assumption we make concerns the boundedness of the kernel.
\begin{tassumption}{Bounded kernel}{bounded_fmap}
$\rkhs$ is a \new{separable} RKHS of functions on $\dspace$ with reproducing kernel $\kr$. The canonical feature map $\fmap: \dspace\rightarrow \rkhs$, defined as $\fmap(x)\de\kr(x,·)$, \new{is measurable for any $x∊\dspace$}.  %
There exists a positive constant $\supfmap<\infty$ such that $\sup_{x\in \dspace}\nrkhs{\fmap(x)}\leq \supfmap$. 
\end{tassumption}

Here and in the following, we denote $\iprkhs{·,·}$ and $\nrkhs{·}$ the RKHS inner-product and the associated norm. 
\Cref{a:bounded_fmap} is satisfied for feature maps derived from a large class of standard kernels such as, e.g., Gaussian and Laplacian kernels on the Euclidean space $\mathbb{R}^d$. It is also satisfied for polynomial kernels on a bounded domain $\dspace$.

We define the (uncentered) covariance operator of $\rkhs$ for the target distribution $\td$ as 
\[ \covop = \int \fmap(x) \kron \fmap(x) \dif\td(x) : \rkhs→\rkhs. \]
where $(\fmap{x} \kron \fmap{x})(f) \de \iprkhs{f, \fmap{x}} \fmap{x}$.
Under \Cref{a:bounded_fmap} it holds $\Tr(\fmap(x) \kron \fmap(x))=\nrkhs{\fmap{x}}²≤\supfmap²$, and thus the operator $\covop$ is a self-adjoint trace class operator on $\rkhs$, which allows us to leverage tools from spectral theory. \new{It is moreover a positive operator since $\fmap(x) \kron \fmap(x)$ is positive for any $x$ (cf. \Cref{a:notations}).}

We now define, for any $λ>0$ the function
\begin{align}
\deff 
	&\de\E_{x\sim\td} \n{\covopl^{-1/2}\fmap(x)}²
	=\Tr(\covop \covopl^{-1}), 
\end{align}
where $C_λ\de C+λI$. 
Under \Cref{a:bounded_fmap}, it always holds that 
$\deff≤\supk/λ<∞$ for any $λ>0$.
The quantity $\deff$ is known as the effective dimension, and is a measure of the interaction between the kernel (or feature map) and the data probability distribution. It is tightly linked to the notion of leverage scores and has been shown to constitute a proper measure of hardness for kernel ridge regression problems \pcite{caponnetto2007OptimalRatesRegularized}. 
It is a quantity of paramount importance in our analysis, and its decay \wrt  $λ$ essentially depends on the decay of the eigenvalues $(σ_i)_{i∊\bN}$ of the covariance operator $\covop$, which characterizes the smoothness of the functions in $\rkhs$. 
In this paper, we will assume that this decay is either polynomial or exponential, as formalized in the next two assumptions.
\begin{tassumption}{Polynomial Decay}{poly_decay}
	There exist 
	$γ∊]0,1]$ and $a_\gamma>0$ such that 
	$σ_i ≤ a_γ i^{-1/γ}$.
\end{tassumption}
Given that $\covop$ is trace class, \Cref{a:poly_decay} always holds at least for $γ=1$, however we are obviously interested in settings $γ<1$ where better rates can be derived.
We stress that assuming a polynomial decay of the spectrum of $\covop$ is equivalent to assuming a polynomial decay of the effective dimension $\deff$, see for instance \textcite[Lemma 11]{fischer2020SobolevNormLearning}.
\begin{tassumption}{Exponential Decay}{exp_decay}
	There exists $β>0$ and $a_β>0$ such that $σ_i ≤ a_β e^{-βi}$.
\end{tassumption}

This assumption implies a bound on the effective dimension which is logarithmic in $1/λ$, as recalled in \Cref{s:bounds_deff}.


The spectral decay of the covariance operator has been studied by \textcite{widom1963AsymptoticBehaviorEigenvalues,widom1964AsymptoticBehaviorEigenvalues}, 
and it is known that Sobolev spaces \new{on bounded domains} correspond to a polynomial decay.  
\begin{tremark}{Sobolev Decay}{sobolev_decay}
\new{For a bounded domain $Ω⊆ℝ^d$}, the Sobolev space \new{$\sob[Ω]$} from \Cref{ex:sobolev} satisfies the polynomial decay assumption with $γ=d/(2s)<1$ as shown by \textcite{widom1963AsymptoticBehaviorEigenvalues}. 
\end{tremark}
\noindent
\new{In dimension $d=1$,} the Gaussian kernel \new{associated to a gaussian density or a density supported on a compact domain} yields an exponential decay, \new{see \textcite[Section 4.3.1]{rasmussen2006GaussianProcessesMachine} and~\textcite[Section B.3]{hayakawa2022PositivelyWeightedKernel}.
These results can be generalized to higher dimensions for product measures \pcite[Appendix A]{bach2017EquivalenceKernelQuadrature}, 
yielding a decay of the form $σ_i≤a_βe^{-βi^{1/d}}$.
The same decay has been derived for the Gaussian kernel and a density with Gaussian tails in \textcite[Lemma 27]{harchaoui2008TestingHomogeneityKernel} by combining the result of \textcite{widom1964AsymptoticBehaviorEigenvalues} with a perturbation argument. 
}

\subsection{Main Rates}
\label{s:main_rates}

We now provide an informal version of \Cref{r:main_result_bis}, which provides rates for our quadrature rule based on leverage scores sampling. 
We provide in \Cref{s:theory} additional variants of this result for uniform sampling (yielding weaker rates) as well as for smoother fractional subspaces of $\rkhs$ (yielding faster rates).

\begin{ttheorem}{Main result, informal}{summary_main_results}
	Let \cref{a:iid,a:bounded_fmap} hold. 
	Let the nodes $\ldm{1}, …, \ldm{m}$ be drawn according to 
	approximate leverage scores \eqref{e:def_als} from the dataset $\{X_1 \dots, X_n\}$, 
	and $w$ be the optimal weights \eqref{e:def_weights}. 
	For $n$ large enough
	it holds:
	\begin{itemize}
		\item under \Cref{a:poly_decay} (polynomial decay),
		choosing $m = Ω(n^γ \log(n)^{1-γ})$, with high probability
		\begin{align*}
			\wce
				&= O\prt*{\frac{\log(m)^{1/(2γ)}}{m^{1/(2γ)}} }
				= O\prt*{\frac{\log(n)^{1/2}}{\new{n^{1/2}}} };
		\end{align*}
		\item under \Cref{a:exp_decay} (exponential decay),
		choosing $m = Ω(\log(n)²)$, with high probability
		\begin{align*}
			\wce
				&= O\prt*{\frac{m^{1/4}}{\exp(\sqrt{m}/c)} }
				= O\prt*{\frac{\log(n)^{1/2}}{\new{n^{1/2}}} }.
		\end{align*}
		for some constant $c$ which does not depend on the dimension.
	\end{itemize}
\end{ttheorem}
Our analysis uses some tools developed by \textcite{rudi2015LessMoreNystrom} in the context of kernel ridge regression, as well as ideas developed by \textcite{chatalic2022MeanNystromEmbeddings,chatalic2022NystromKernelMean} for the approximation of kernel mean embeddings.
Note that contrarily to methods which try to fill the domain as uniformly as possible, our analysis is not restricted to a bounded domain, and the constants in the $O(·)$ and $Ω(·)$ notations of \Cref{r:summary_main_results} do not depend on the dimension. 

According to \Cref{rm:sobolev_decay}, the polynomial decay hypothesis covers as a special case the Sobolev setting taking $γ=d/(2s)$.   
\begin{tcorollary}{Sobolev space}{summary_main_result_sobolev}
	Under the hypotheses of \Cref{r:summary_main_results}, for $\rkhs=\sob$ it holds
		\begin{align*}
			\wce[\sob] &= O\prt*{\frac{\log(m)^{s/d}}{m^{s/d}} }.
		\end{align*}
\end{tcorollary}

\paragraph{Optimality and smoothness adaptivity} 
We stress that in all our results, 
the number of nodes $m$ is directly chosen as a function of the number $n$ of samples, and thus all rates can be interpreted \wrt to both variables.
On one side, it is known from a statistical perspective that
the minimax estimation rate when building the quadrature from $n$ \tiid samples is $O(n^{-1/2})$
for continuous translation-invariant kernels on $\mathbb{R}^d$ and discrete measure, or measures with infinitely differentiable densities (e.g., Gaussian)~\pcite{tolstikhin2017minimax}; a similar rate has also been obtained in a non-iid setting~\pcite{cherief-abdellatif2022FiniteSampleProperties}. 
\new{Since we do not make extra assumptions on the probability distribution $ρ$ in this paper, the bound in \Cref{r:summary_main_results} is thus optimal (up to the log term) in the sense that no other estimator could reach a better rate with respect to $n$ under the same assumptions.} 
Quantization rates, on the other side, correspond to rates with respect to the number $m$ of nodes on which the quadrature is supported, and lower bounds are known to be faster than $O(m^{−1/2})$ in this case, such as shown in \Cref{r:summary_main_result_sobolev} where we obtain a rate which adapts to the smoothness of the underlying space. This result turns out to match (up to log terms) the optimal rate in this setting \pcite[1.3.12 Proposition 3]{novak1988DeterministicStochasticError}.


For instance, 
\Cref{r:summary_main_results} shows that in the case of polynomial decay 
our estimator achieves the quantization rate $O(m^{-1/(2γ)})$ (up to log terms) provided that one has access to $n=Ω(m^{1/γ})$ \tiid samples in the first place. 
Alternatively, we recover the rate $O(n^{-1/2})$ (up to logarithmic terms) at the reduced cost of manipulating an estimator built using only $m=Ω(n^γ)$ samples.
In the following, we will formulate the rates in this first manner (i.e. as a function of $m$) and always compare estimators built using the same number of samples: although the complexity of the algorithms used to pick these $m$ points and weights may differ, the complexity of afterwards evaluating the quadrature rule for a new function is directly driven by $m$.

\subsection{Related Work}
\label{s:related_work}

\begin{table*}
\setlength{\tabcolsep}{3.5pt}

\centerline{%
\resizebox{1.2\linewidth}{!}{%
\begin{tabular}{@{}lllll@{}}
	\thicktoprule
	\multicolumn{2}{l}{ \textbf{Method} }
		& \textbf{Weights} 
		& \textbf{Time complexity} 
		& \textbf{Guarantees} \\
	\thickmidrule
\multicolumn{2}{l}{\textbf{Random selection of the nodes}} &&& \\
	\cmidrule[1.2pt]{2-5}
	& Monte-Carlo (Uniform)
		& Uniform
		& $O(m)$  
		& $O(m^{-1/2})$ \pcite[2.1.3]{novak1988DeterministicStochasticError} 
	\\
\cmidrule[1.2pt]{2-5}
	& MCMC targeting $\td$ \pcite{briol2019ProbabilisticIntegrationRole}
		& Optimized
		& Not found 
		& $\wce[\sob[{[0,1]^d}]]=O(n^{-s/d+ε})$ for any $ε>0$
		\\
\cmidrule[1.2pt]{2-5}
	& \makecell[tl]{Projection DPP \pcite{belhadji2019KernelQuadratureDPPs}\\
		\cred{(Requires eigendecomposition of $\covop$)}}
		& Optimized
		& Rejection sampling + $O(m³)$
		& $(\E\, \WCE ²)^{1/2} \lesssim r_{m+1}^{1/2}$ \pcite[Theorem 4]{belhadji2021AnalysisErmakovZolotukhinQuadrature}\\
	\cmidrule{2-5}
	& Ermakov-Zolotukhin \pcite{belhadji2021AnalysisErmakovZolotukhinQuadrature}
		& Non-optimal
		& Rejection sampling + $O(m³)$
		& $(\E\, \WCE ²)^{1/2} \lesssim r_{m+1}^{1/2}$ \pcite[Theorem 3]{belhadji2021AnalysisErmakovZolotukhinQuadrature}
		\\
	\cmidrule{2-5}
	& \makecell[tl]{Continuous volume sampling\\ \pcite{belhadji2020KernelInterpolationContinuous}}
		& Optimized
		& $O(m⁵)$ for MCMC mixing guarantees
		& $(\E\,\WCE ²)^{1/2}\lesssim σ_{m+1}^{1/2}$
	\\
\cmidrule[1.2pt]{2-5}
	& (True) Leverage scores sampling \pcite{bach2017EquivalenceKernelQuadrature}
		& By regularized LS
		& \cred{\xmark{}}\, No algorithm
		& $\WCE ≤ 4λ$ provided $m\gtrsim \deff \log(λ^{-1})$ \\
	\cmidrule{2-5}
	& This work, \Cref{r:corollary_assumption_poly_decay}
		(Uniform)
		& Optimized
		& $Θ(m³+nmd)$
		& \makecell[tl]{ 
			Under \Cref{a:poly_decay}: 
			$\WCE\lesssim m^{-(1-γ/2)} \log\prt*{ m }$\\
			 } \\
		&  
		& 
		& 
		& \quad\text{in particular} $\wceS[\sob]=O\prt*{m^{-(1-\nicefrac{d}{4s})} \log\prt*{ m } }$
			\\
		& 
		& 
		& 
		& 
		Under \Cref{a:exp_decay}
			$\WCE =O(m^{-1}\log(m))$
	\\
	\cmidrule{2-5}
		& This work, \Cref{r:main_result_bis}
		((A)RLS)
		& Optimized
		& $Θ(m³ + nmd + n^{1+2γ})$
		& \makecell[tl]{%
			Under \Cref{a:poly_decay}:
			 $\WCE=O( m^{-1/(2γ)} \log(m))$ 
		} \\
		& 
		& 
		& 
		& \quad\quad\text{in particular }
			$\wceS[\sob]\lesssim m^{-s/d}\polylog(m)$ 
			\\
		& 
		&
		& $Θ(m³+nmd +\log(n)²n)$
		&
		Under \Cref{a:exp_decay}: 
			$\WCE =O( m^{1/4}\exp(-\sqrt{m}/\sqrt{cst}))$
			\\
\thickmidrule
\multicolumn{2}{l}{\textbf{Greedy methods focusing on the residual}} &&& \\
	\cmidrule[1.2pt]{2-5}
	\hspace{0.5cm}
	& \makecell[tl]{$f/P$-greedy on $\dspace$~\pcite{mueller2009KomplexitaetUndStabilitaet} / \\
		\quad SBQ~\pcite{huszar2012OptimallyweightedHerdingBayesian} }
		& Optimized
		& \makecell[tl]{$O(m³)$ + $m$ nonconvex subproblems\\ $O(dn+m²)$/objective evaluation}
		& \makecell[tl]{ $\WCE=O(m^{-1/2})$ ($\dspace$ bounded) \\
	 \pcite[Theorem 5.1]{santin2022SamplingBasedApproximation}
		}
	\\
	\cmidrule{2-5}
	& $f/P$-greedy on $\data$
		& Optimized
		& $O(\cred{n²}+nm(d+m))$ 
		& \cred{\xmark} Not found. 
	\\
	\cmidrule[1.2pt]{2-5}
	& Herding \pcite{chen2010SupersamplesKernelHerding} 
		& Uniform
		& \multirowcell{3}[-0.8em][c]{ $m$ non-convex subproblems with\\ $O(nd)$/objective evaluation }
		& \makecell[tl]{
			$\WCE=O(m^{-1})$ in finite dimension \pcite{chen2010SupersamplesKernelHerding} \\ $\WCE=O(m^{-1/2})$ otherwise
			} \\
	\cmidrule{2-3}
	\cmidrule{5-5}
	&Frank-Wolfe (FW) with line search
		& In the simplex
		& 
		& \makecell[tl]{Exponential in finite dimension~\pcite{bach2012EquivalenceHerdingConditional}\\
		$\WCE=O(m^{-1/2})$ otherwise }
		 \\
	\cmidrule{2-3}
	\cmidrule{5-5}
	& \makecell[tl]{Fully-corrective FW \pcite{jaggi2013RevisitingFrankWolfeProjectionfree} / \\
		\quad Continuous OMP /
		$f$-greedy}
		& Optimized
		& 
		& \makecell[tl]{
			Exponential in finite dimension \pcite{bach2012EquivalenceHerdingConditional} \\ $\WCE=O(m^{-1/2})$ otherwise }
	 \\
	\cmidrule{2-5}
	& OMP (\aka $f$-greedy) on $X$
		& Optimized
		& $O(\cred{n²}+nm(d+m))$ 
		& $\WCE=O(m^{-1/2})$ \pcite{devore1996RemarksGreedyAlgorithms}
		\\
	\cmidrule{2-5}
	& \makecell[tl]{
		Continuous OMP w/ global steps \\
		+ Nyström or RF approximation \\
		\pcite{chatalic2022MeanNystromEmbeddings,keriven2017CompressiveKmeans}
		}
		& Optimized
		& \makecell[tl]{For RF: $O(nmd\log(d))$ \\
		+ $m$ non-convex subproblems\\ $O(m²d\log(d))$/objective eval.} 
		& \cred{\xmark{}} Not found. 
		\\
\thickmidrule
\multicolumn{2}{l}{\textbf{Other approaches}} &&& \\
	\cmidrule[1.2pt]{2-5}
	& \makecell[tl]{Recombination Mercer \pcite{hayakawa2022PositivelyWeightedKernel} \\ \cred{(Requires eigendecomposition of $\covop$)}}
		& Opt. in simplex
		& $O(nm²+m³)$ in average
		& \makecell[tl]{ 
			$(\E_{\data}\, \WCE ²)^{1/2}\lesssim r_m^{1/2}+n^{-1/2}$\\
			\pcite[Cor. 2]{hayakawa2022PositivelyWeightedKernel}
		} \\
	\cmidrule{2-5}
	& Recombination Nyström
			\pcite{hayakawa2022PositivelyWeightedKernel}
		& Opt. in simplex 
		& $O(nm²+m³\log(n/m))$
		& \makecell[tl]{
			Under \Cref{a:exp_decay}: \\
			$\E[\WCE] = O(r_{m+1}^{1/2} + \polylog(m)/m + n^{-1/2})$ \\
			\pcite[Th. 6 + Rem. 1]{hayakawa2023SamplingbasedNystromApproximation}
		} \\
\cmidrule[1.2pt]{2-5}
	& Thinning \pcite{dwivedi2022GeneralizedKernelThinning,dwivedi2021KernelThinning}
& Uniform
& $O(\cred{n²}c_κ)$
& \multirowcell{2}{
	For $m=\sqrt{n}$, \new{subexponential tails/compac support:}\\
	\new{analytic kernel:} $\WCE\lesssim\polylog(m)m^{-1}$ \\
	\new{Matérn kernel: 
	$\WCE\lesssim\polylog(m)m^{-(1-d/\lfloor s\rfloor)}$ }\\
} \\[0.4cm]	
\cmidrule{2-4}
	& Thinning \pcite{shetty2022DistributionCompressionNearlinear}
		& Uniform
		& $O(n\log(n)³)$
		&
		\\
\thickmidrule
\multicolumn{2}{l}{\textbf{Space-filling methods}} &&& \\
	\cmidrule[1.2pt]{2-5}
	& $P$-greedy
		& Optimized
		& \makecell[tl]{$m$ non-convex subproblems,\\ $O(m²+md)$ / objective evaluation}
		& \makecell[tl]{ $\WCE(\sob)=O(m^{-s/d})$ \\
		($\dspace$ bounded w/ cone condition, TI kernel)\\
		\pcite[Th. 3.2/Rem. 4.1]{santin2022SamplingBasedApproximation}
		}
	\\
	\cmidrule{2-5}
	&$P$-greedy on $X$
		& Optimized
		& $O(nm(d+m))$ 
		& \cred{\xmark{}} Not found.
		\\
\thickbottomrule
\end{tabular}%
}%
}%
\caption{\label{t:related_works}Summary of main quadrature methods.
We denote $\WCE\de\wce$ for conciseness for a generic RKHS $\rkhs$, and $r_m=\sum_{j≥m} σ_j$. 
Complexities are given assuming that the kernel evaluation costs $O(d)$. For greedy algorithms, complexities are intended w.r.t. the empirical problem, so that the nonconvex subproblems have complexities depending on $n$.
SBQ = sequential Bayesian quadrature; 
OMP = orthogonal matching pursuit; 
RF = random features; 
DPP = determinantal point process; 
TI = translation-invariant. 
Note that under \Cref{a:poly_decay}, for $γ<1$ it holds $r_m≤\tfrac{γ\, a_γ}{1-γ} (m-1)^{1-1/γ}$, and under \Cref{a:exp_decay} it holds $r_m≤\tfrac{a_β}{1-e^{-β}} e^{-βm}$.
}
\end{table*}

\new{Numerical approximation of integrals is a very broad topic and has a long history. 
Our focus here is on worst-case quadrature methods in reproducing kernel Hilbert spaces when one 
targets a non-uniform probability distribution known via \tiid samples. 
We provide below} an overview of existing methods in the literature with a focus on available rates and associated computational complexities.
One can roughly categorize these methods in a few categories: random designs (where the $m$ nodes are sampled, either independently or jointly), coreset methods (which reduce, often recursively, the initial set of $n$ samples while maintaining some key properties), methods which try to fill the space, and greedy methods which pick the nodes iteratively.
\Cref{t:related_works} provides a summary of the different approaches.
\new{We refer the reader to \textcite{novak2010TractabilityMultivariateProblems} for a broader coverage of the topic, and in particular of existing lower bounds in terms of information complexity. }

\paragraph{Random designs} 
Our method belongs to the family of random designs, in the sense that the locations of the nodes are randomly drawn − in our case subsampled among the \tiid samples $\data$, but this could be relaxed.
The simplest way to produce a random design is the Monte-Carlo method, which achieves a $O(m^{-1/2})$ rate~\pcite[2.1.3]{novak1988DeterministicStochasticError}.
This rate is optimal in many settings when having access to $m$ \tiid samples, \eg for translation-invariant kernels and discrete measures or measures with infinitely differentiable densities~\pcite{tolstikhin2017minimax}, however we consider here quadrature rules built starting from $n>m$ \tiid samples and that can thus have better rates with respect to $m$.

Our method is closely related to the work of \textcite{bach2017EquivalenceKernelQuadrature}, who considers \tiid sampling of the nodes according to (continuous) leverage scores and slightly different weights.
For a particular choice of the random features, the bound in \textcite[Proposition 1]{bach2017EquivalenceKernelQuadrature} translates to a bound on the worst-case error. 
However, in general the method cannot be implemented as it involves multiple quantities that cannot be computed. 

\textcite{briol2017SamplingProblemKernel} have also introduced a heuristic distribution with heavy tails as well as a sequential Monte-Carlo procedure to sample from it, and reported empirically better stability.

Joint sampling of the nodes has been considered, for instance using determinantal point processes~\pcite{belhadji2021AnalysisErmakovZolotukhinQuadrature}, which is also related to the Ermakov-Zolotukhin quadrature rule \pcite{belhadji2021AnalysisErmakovZolotukhinQuadrature}.
Defining $r_{m}=\sum_{i≥m} σ_i$, theoretical convergence rates of order $\E[\wceS²]=O(r_{m+1})$ have been proven for both methods. 
\textcite{belhadji2020KernelInterpolationContinuous} also considered continuous volume sampling, which consists in jointly sampling the nodes following a probability density $\det(K_m)$ with respect to the base measure $\td^{\otimes m}$. 
This method yields a faster theoretical rate $\E[\wceS²]=O(σ_{m+1})$. 
Empirically, DPP sampling has also been reported to converge at this faster rate.

Random sampling from data streams, \ie in one pass over the data without knowing beforehand the size $n$ of the dataset, has been investigated by \textcite{paige2016SupersamplingReservoir}; no convergence rates have however been reported in this setting.
Note that our quadrature rule can be interpreted as a kind of Nyström approximation~\pcite{williams2001UsingNystromMethod},
and many other sampling rules have been studied in this context \pcite{fanuel2022NystromLandmarkSampling,kumar2012SamplingMethodsNystrom}.

\paragraph{Space-filling methods} 
In the setting where $\dspace$ is a compact set,  multiple methods have been proposed to fill the space with more regularity than what a Monte-Carlo sample would typically produce.
Such methods have been studied for decades in the literature on model-free design of experiments, see for instance \textcite{garud2017DesignComputerExperiments}.
Quasi Monte-Carlo (QMC) methods is a well-known way to generate low-discrepancy sequences, but is usually restricted to very particular domain and distributions − such as the uniform distribution on the hypercube, or the Gaussian distribution on the sphere.
\new{\textcite[Theorem 15.21]{dick2010DigitalNetsSequences}} 
for instance 
\new{derived rates that are arbitrary close to the optimal one} 
for QMC and Sobolev spaces of dominating mixed smoothness on $[0,1]^d$, 
\new{see also \textcite{briol2019ProbabilisticIntegrationRole}.}
\new{We refer the reader to \textcite{dick2022LatticeRulesNumerical} for a broader coverage of these methods. }

In the context of kernel interpolation, Fekete points are defined as the nodes $\ldm$ maximizing $\det(K_m)$, 
by analogy with polynomial interpolation in 1d where one is interested in the points maximizing the determinant of the Vandermonde matrix \pcite{bos2010ComputingMultivariateFekete}.
Maximizing directly $\det(K_m)$ is most often untractable or expensive, but kernel approximations can naturally be used~\pcite{karvonen2021KernelbasedInterpolationApproximate}.
Note that this objective is related to the density used in the continuous volume sampling method mentioned above \pcite{belhadji2020KernelInterpolationContinuous}, however there is here no dependence in the probability measure $\td$ (or $\td$ is assumed to be uniform).

Greedy maximization of $\det(K_m)$ as been introduced as the $P$-greedy method in the kernel interpolation literature \pcite[Section 4]{demarchi2005NearoptimalDataindependentPoint} (cf. \Cref{s:greedy} for more details),
and used in multiple contexts \pcite{chen2018FastGreedyMAP,carratino2021ParKSoundEfficient}. 

\paragraph{Other randomized methods} 
Recently, \textcite{hayakawa2022PositivelyWeightedKernel,hayakawa2023SamplingbasedNystromApproximation} used recombination algorithms to compute a discrete measure $ρ_m$ supported on $m$ points such that for a set of $m$ test functions $(φ_i)_{1≤i≤m}$ it holds exactly $\int φ_i \dif ρ_m=\int φ_i\dif \ed$. The test functions are built either using the Mercer decomposition or using a Nyström approximation with truncation, and both randomized and deterministic algorithms are known to compute the reduction from $\ed$ to $ρ_m$. 
Assuming an exponential decay of the covariance's spectrum, the authors obtain a bound on the expected worst case error in $\E[\wceS] = O(\sqrt{r_{m+1}}+\polylog(m)/m+n^{-1/2})$~\pcite[Theorem 6, Remark 1]{hayakawa2022PositivelyWeightedKernel}.

Quadrature rules which are supported on a subset of the initial $n$ samples (as we do) can also be interpreted as (weighted) coresets.
For instance, the simple greedy algorithm of \textcite[Section 3.1]{karnin2019DiscrepancyCoresetsSketches} covers the case of kernel density estimation as a special case, however it only induces a $O(n^{-1/2})$ rate.
Thinning methods have been proposed to build a coreset of size $m=\sqrt{n}$ by recursively reducing by half the initial dataset. 
The initial $O(n²)$ complexity of kernel thinning \pcite{dwivedi2021KernelThinning} has been reduced to $O(n\log(n)³)$ by \textcite{shetty2022DistributionCompressionNearlinear}, and the error of the coreset has been studied under various hypotheses but goes down to $O(\polylog(m)/m)$ for \eg a Gaussian kernel with a sub-exponential data distribution \pcite{dwivedi2022GeneralizedKernelThinning}.

\paragraph{Greedy methods} 
An alternative to random design (where the $m$ nodes are sampled, either \tiid or jointly) and coreset methods (which often recursively reduce the initial set of $n$ samples), is to iteratively select the nodes by minimizing some notion of residual.

Kernel herding \pcite{chen2010SupersamplesKernelHerding} falls in this category, and has originally been introduced with uniform quadrature weights.
It can be interpreted as a particular case of the Frank-Wolfe algorithm~\pcite{bach2012EquivalenceHerdingConditional} and has been extended in multiple directions \pcite{jaggi2013RevisitingFrankWolfeProjectionfree,lacoste-julien2015SequentialKernelHerding,briol2015FrankWolfeBayesianQuadrature}.
These algorithms are also known to be closely related to matching pursuit and its variants~\pcite{locatello2017UnifiedOptimizationView}.
Fast rates in $O(1/m)$ and even exponential rates have been obtained for such methods, but 
depend on geometric quantities that cannot always be controlled easily and thus 
essentially cover finite-dimensional spaces. 
\textcite{khanna2021GeometricRatesConvergence} derived rates that hold in infinite dimension, but rely on the hypothesis that the target distribution is sparse. 
\textcite{tsuji2022PairwiseConditionalGradients} introduced blended pairwise conditional gradients as a variant of Frank-Wolfe more amenable to analysis in the infinite-dimensional setting, however theoretical rates remain of order $O(m^{-1/2})$.

In order to limit the impact of local minimas, global optimization steps can be added after each selection of a new node. This leads to the compressive clustering algorithm, which additionally relies on random features or Nyström approximations of the kernel~\pcite{keriven2017CompressiveKmeans,chatalic2022MeanNystromEmbeddings} and is closely related to the sliding Frank-Wolfe algorithm \pcite{denoyelle2019SlidingFrankWolfe}. 
Although theoretical guarantees in this context rather focus on the recovery of sparse measures, the considered objective function corresponds to a tractable approximation of the quadrature worst-case error and the algorithms proposed in this context are thus highly relevant for our goal.

Interestingly, greedy minimization of the quadrature worst-case error $\inf_{w}\wce$ actually does not lead to orthogonal matching pursuit, but to the $f/P$-greedy method from the kernel interpolation literature \pcite{mueller2009KomplexitaetUndStabilitaet}, which is also known as sequential Bayesian quadrature \pcite{huszar2012OptimallyweightedHerdingBayesian}. 
Rates of order $O(m^{-1/2})$ have been obtained both for $f$-greedy and $f/P$-greedy methods \pcite[Corollary 20]{santin2022SamplingBasedApproximation}, however faster rates are typically observed in practice.


\paragraph{Bayesian Quadratures} 
In the Bayesian literature, 
one is typically interested in computing not only the integral $\I(f)$, but also a probability distribution encoding the belief in this estimation. 
To achieve this goal, a prior distribution over the integrand $f$ is assumed. 
When this prior is chosen to be a Gaussian process whose covariance function is a kernel $\kr$, the maximum a posteriori estimator corresponds to the optimally-weighted quadrature rule in the RKHS associated to $\kr$. Moreover, the variance of $\I(g)$ when $g$ follows the posterior distribution corresponds exactly to the worst-case error of the optimally-weighted quadrature supported on the $m$ nodes, $\Var[\I(g)]=\inf_{w∊\bR^m}\wce$, see e.g. \pcite[Section 3.2]{huszar2012OptimallyweightedHerdingBayesian}. 
This gives another interpretation to our target objective, and justifies for instance that the sequential Bayesian quadrature is equivalent to the greedy minimization of the worst-case error (cf. \cref{s:greedy}). 

In this Bayesian context, \textcite{briol2019ProbabilisticIntegrationRole} derived optimal convergence rates for MCMC sampling \new{in Sobolev spaces on $[0,1]^d$ using bounds based of the fill distance,} and Quasi Monte-Carlo sampling in Sobolev spaces of dominating mixed smoothness \new{using a result from~\textcite{dick2010DigitalNetsSequences}.}

\new{Rates for adaptive bayesian quadrature methods, for which the choice of the nodes is allowed to depend on the integrand, have also been studied assuming that the target function can be modeled as a transformation of a Gaussian process~\pcite{kanagawa2019ConvergenceGuaranteesAdaptive}.}


\paragraph{Other contributions} 
A few other methods exist beyond the main families of algorithms presented above, such as particle methods which start directly from a pool of $m$ nodes whose locations are jointly updated by gradient descent \pcite{arbel2019MaximumMeanDiscrepancy}, but no rates have been reported in this setting. 
\textcite{muandet2014KernelMeanEstimationa} introduced shrinkage estimators 
and showed that they perform better than Monte-Carlo approaches under mild assumptions on the probability distribution of interest.
Such shrinkage strategies are complementary to our approach, in the sense that they can be combined with any existing estimator.
%
In another context, \textcite{kanagawa2020ConvergenceAnalysisDeterministic}
proposed a theoretical analysis in the misspecified setting (\ie when the integrand in \eqref{e:def_int} does not belong to the RKHS used to design the quadrature rule), and showed that adaptivity to the smoothness of the integrand can still be achieved.

\paragraph{Summary} 
Overall, our approach has the merit of achieving optimal rates while being efficiently implementable,  
which complements nicely the state of the art. 
For instance, greedy methods obtain very good empirical results, but the observed rates are not matched by existing theoretical guarantees. 
Other existing random designs do not always yield optimal rates, and are often costly to implement, when not intractable. 
Methods trying to fill uniformly the domain are restricted by definition to bounded domains, 
and perform poorly in practice (see \Cref{s:experiments}) despite optimal rates being known in some settings \pcite{santin2022SamplingBasedApproximation}; this can likely be explained by high multiplicative constants, and the fact that such methods do not adapt to the target distribution. 

\section{Theoretical Analysis}
\label{s:theory}

We show in \Cref{ss:kme} that the problem of designing quadrature rules can be recast as the approximation of the so-called kernel mean embedding, and then provide bounds on the worst-case error for uniform sampling (\Cref{s:uniform_sampling}), ARLS sampling (\Cref{s:ls_sampling}), as well as improved rates for ARLS sampling under an additional smoothness condition (\Cref{s:source_condition}).

\subsection{\new{RKHS Quadratures} and Kernel Mean Embeddings}
\label{ss:kme}

When considering $\rkhs$ to be a reproducing kernel Hilbert space associated to a kernel $\kr$ satisfying \Cref{a:bounded_fmap}, the quadrature error is connected to the approximation of the so-called kernel mean embedding of the considered probability measure $\rho$,
\begin{align}
	\me 
	&\de \me(\td) 
	\de \int\kemb{x}\dif\td(x). 
	\label{e:mean_embedding}
\end{align}
Indeed $\fmap$ is integrable with respect to any probability distribution over $\dspace$ under \Cref{a:bounded_fmap}, and thus the kernel mean embedding \eqref{e:mean_embedding} is well defined, interpreting the integral as a Bochner integral~\pcite[Chapter 2]{diestel1977VectorMeasures}. 
\new{%
It should be noted that for any $y$ the linear functional $h↦\ip{\fmap{y},h}$ is bounded under \Cref{a:bounded_fmap}, and thus closed~\pcite[4.13.5 (a)]{kreyszig1989IntroductoryFunctionalAnalysis}. Hence by Hille's theorem \pcite[Theorem 6]{diestel1977VectorMeasures}
it holds 
$\ip{\fmap{y}, \int \fmap{x}\dif ρ(x)}
	=\int \ip{\fmap{y},\fmap{x}}\dif ρ(x)
	=\int \kr(y,x)\dif ρ(x)$. 
Moreover, the operator $I_ρ:f ↦ \int f(x) \dif ρ(x)$ is a continuous linear functional under \Cref{a:bounded_fmap} given that $|I_ρf|≤\int|\iprkhs{f,\fmap{x}}|\dif ρ(x)≤\supfmap‖f‖$, 
and thus admits a Riesz representation $m_ρ$, \ie $I_ρ(f)=\iprkhs{f,m_ρ}$ holds for any $f∊\rkhs$ \pcite{reed1981FunctionalAnalysisVolume}. 
Considering $f=\fmap{y}$ we get $m_ρ(y)=\int \kr(y,x)\dif ρ(x)$ for any $y$, \ie, the kernel mean embedding is also the Riesz representant of $I_ρ$.
}

Initially introduced by \textcite{DBLP:conf/alt/SmolaGSS07}, kernel mean embeddings (KME) conveniently allow to represent a probability distribution via a mean vector in a Hilbert spaces \pcite{DBLP:journals/ftml/MuandetFSS17}. 
They have found applications in various areas such as anomaly detection \pcite{DBLP:journals/corr/ZouLPS14}, approximate Bayesian computation \pcite{DBLP:conf/aistats/ParkJS16}, domain adaptation \pcite{DBLP:conf/icml/ZhangSMW13}, imitation learning \pcite{DBLP:conf/aaai/KimP18}, nonparametric inference in graphical models \pcite{DBLP:journals/spm/SongFG13}, functional data analysis \pcite{hayati2020kernel}, discriminative learning for probability measures \pcite{DBLP:conf/nips/MuandetFDS12} and differential privacy \pcite{pmlr-v80-balog18a,chatalic2021CompressiveLearningPrivacy}. 

In the following lemma, we show how the error of a quadrature rule can be related to the error of a kernel mean embedding estimation problem. This result is common knowledge, but included for completeness.
\begin{tlemma}[noprooflink]{}{quad_to_kme}
For any 
set of points $\ldm=(\ldm{i})_{1≤j≤m}$ and any weights $(w_i)_{1≤i≤m}$, it holds
\begin{align*}
	\wce
	&= \nrkhs*{\,\me - \sum_{j=1}^{m} w_j \fmap{\ldm{j}} \,}\enspace.
\end{align*}
\end{tlemma}

\begin{tproofof*}{r:quad_to_kme}{}
For any $h∊\rkhs$ such that $\nrkhs{h}≤1$, it holds that
\begin{align*}
	\absv*{\int h(x)\dif\td(x) - \sum_{j=1}^{m} w_j h(\ldm{j})}
	&\stackrel{(i)}{=} \absv*{\int \iprkhs{h,\kemb{x}}\dif\td(x) - \sum_{j=1}^{m} w_i \iprkhs{h,\kemb{\ldm{j}}}} \\
	&\stackrel{(ii)}{=} \absv*{\iprkhs*{h, \int \kemb{x}\dif\td(x) - \sum_{j=1}^{m} w_j \kemb{\ldm{j}} }} \\
	 &\stackrel{(iii)}{≤} \nrkhs*{\,\me - \sum_{j=1}^{m} w_j \kemb{\ldm{j}}\,}\enspace,
\end{align*}
where we used the reproducing property of the RKHS $\rkhs$ for (i) and the Cauchy-Schwarz inequality for (iii).
\new{Equality $(ii)$ follows from Hille's theorem~\pcite[Theorem 6]{diestel1977VectorMeasures} applied to the linear functional $f↦\ip{h,f}$, which is bounded given that $‖h‖≤1$ and thus closed~\pcite[4.13.5 (a)]{kreyszig1989IntroductoryFunctionalAnalysis}. }

The proof is concluded by observing that
\begin{align*}
	h= \nrkhs*{\,\me - \sum_{j=1}^{m} w_j \kemb{\ldm{j}}\,}^{-1} \prt*{\me - \sum_{j=1}^{m} w_j \kemb{\ldm{j}}} \enspace,
\end{align*}
is on the unit sphere in $\mathcal{H}$ and gives the equality.
\end{tproofof*}

\paragraph{Discrete estimators} 
Denoting $\ed=\frac{1}{n}\sum_{1≤i≤n} δ(X_i)$ the empirical distribution of $\data$, where $δ(·)$ denotes the Dirac delta function, one can define
\begin{align}
	\eme &\de μ(\ed)
	= \frac{1}{n} \sum_{i=1}^n \fmap{X_i}.
	\label{e:empirical_kme}
\end{align}
By \Cref{r:quad_to_kme},
the error of the empirical estimator \eqref{e:empirical_quad} is $\wceH[\eI]=\nrkhs{\eme-\me}$. 
\new{This quantity decreases at the rate $O(1/\sqrt{n})$ for any $ρ$ as a consequence of Bernstein inequality in Hilbert spaces \pcite[Th. 3.3.4]{yurinsky1995SumsGaussianVectors}.}
More generally, any quadrature rule $\nI$ can be associated to a sparse estimator of the kernel mean embedding
\begin{align}
	\nyseme &\de \sum_{j=1}^m w_j \fmap(\ldm{j})
	\label{e:sparse_kme}
\end{align}
and the discrete approximation \eqref{e:def_nI} can be computed as $\nI(f)=\ip{\nyseme,f}$ for any $f∊\rkhs$.

\paragraph{A randomized Nyström estimator} 
Our quadrature rule, obtained by sampling the landmarks $\ldms$ from the data $\data$ and choosing the weights according to \eqref{e:def_weights}, has a simple expression in terms of kernel mean embeddings.
Let
\begin{align*}
	\rkhsm &\de \spa\cb*{\fmap(\ldm{1}),\ldots ,\fmap(\ldm{m})}
	⊆ \rkhs
\end{align*}
be the finite dimensional subspace spanned by the features of the landmarks, 
and $\Pm$ the orthogonal projection on this subspace,
one can easily check (see \Cref{s:computation_weights}) that
\begin{align}
	\nyseme &\de \Pm \eme. \label{e:expr_est_proj}
\end{align}
One can in particular think of $\nyseme$ as an interpolator of $\eme$ at the location of the nodes, given that for any $j∊\cb{1,…,m}$, as $\fmap(\ldm{j})∊\ran(\Pm)$ it holds $\nyseme(\ldm{j})=\iprkhs{\Pm\eme, \fmap(\ldm{j})}=\iprkhs{\eme,\fmap(\ldm{j})}=\eme(\ldm{j})$.

As a consequence of \eqref{e:expr_est_proj} and \Cref{r:quad_to_kme}, our main goal from a theoretical perspective is to bound the quantity 
\begin{align*}
	\wceH=\n{\me-\Pm\eme}
\end{align*}
both for uniform and ARLS sampling.




\begin{tremark}{Kernel matrix}{kmat}
It can easily be checked that
\begin{align*}
	\norm{\eme-\nyseme}²
	&= \norm{\Pm^\perp \eme}² 
	≤ \tfrac{1}{n} \norm{\kmat-\tilde{K}_n}_{\text{op}}
\end{align*}
where $\kmat$ and $\tilde{K}_n$ respectively denote the $n×n$ kernel matrices of the data $\data$ with and without Nyström approximation.
Hence, existing results on the Nyström approximation of the kernel matrix in operator norm induce bounds on the worst-case quadrature error, using the error decomposition $\wceH≤\n{\me-\eme}+\n{\eme-\Pm\eme}$.
Such bounds would however be sub-optimal, and we thus rely for our analysis on a different decomposition. 
\end{tremark}

\begin{tremark}{Power function}{}
In another context, \textcite{hayakawa2023SamplingbasedNystromApproximation} obtained quadrature guarantees by studying the integral \wrt the probability distribution $\td$ of the quantity $\nrkhs{\Pm^\perp \fmap{x}}$, which is known in the kernel interpolation literature as the power function and has been well studied \pcite{wendland2004ScatteredDataApproximation}. 
This still differs from our analysis, which rather relies on bounds on $\n{\Pm^\perp (\covop+λI)^{1/2}}$.


\end{tremark}

\begin{tremark}{Maximum Mean Discrepancy}{}
Mean embeddings naturally induce a semi-metric on the space of probability distributions $\mathcal{P}(\mathcal{X})$ known as the maximum mean discrepancy \pcite{DBLP:conf/alt/SmolaGSS07}. It is defined, for any two probability distributions $\rho_1$ and $\rho_2$, as 
\begin{align*}
    \textup{MMD}(ρ₁, ρ₂) \de \nrkhs{μ(ρ_1)-μ(ρ_2)}\enspace.
\end{align*}
It satisfies all the properties of a metric except, in general, the definiteness, depending on whether the mean embedding $\rho \mapsto \mu(\rho)$ is injective or not (we refer the interested reader to \textcite{sriperumbudur2010} for more details). Such metrics have found applications in many contexts such as, to cite a few, two-sample testing \pcite{DBLP:journals/jmlr/GrettonBRSS12,DBLP:conf/ismb/BorgwardtGRKSS06}, neural networks optimization \pcite{DBLP:conf/ismb/BorgwardtGRKSS06}, generative models \pcite{DBLP:conf/nips/LiCCYP17,DBLP:conf/iclr/SutherlandTSDRS17}. Given their wide applicability, maximum mean discrepancies are also an important motivation for better approximating mean embeddings.
An interesting property of the MMD is that it is an integral probability metric \pcite{muller1997integral}, a class of metrics which uses test functions to compare distributions. More precisely, we have
\begin{align*}
    \textup{MMD}(ρ₁, ρ₂){=}\sup_{f∊\rkhs: \lVert f \rVert \leq 1} \lvert \mathbb{E}_{X_1\sim\rho_1}f(X_1){-} \mathbb{E}_{X_2\sim\rho_2}f(X_2) \rvert
\end{align*}
where $\rkhs$ denotes the reproducing kernel Hilbert space associated to the chosen kernel.
These two representations of the MMD allow to leverage the wide set of tools from both kernel methods and integral probability metric theories (see \textcite{sriperumbudur2012EmpiricalEstimationIntegral,sriperumbudur2009integral} for examples of the latter).
Although we focus on the problem of designing quadratures, it should be noted that the algorithms and bounds discussed in this paper directly translate to results on the MMD, see for instance the discussion in \textcite[Section 5]{chatalic2022NystromKernelMean}.
\end{tremark}

\subsection{Rates for Uniform Sampling}
\label{s:uniform_sampling}

We now state our general result for uniform sampling.  We then specialize it using additional knowledge on the spectral properties of the covariance operator. This result was initially presented in \textcite{chatalic2022NystromKernelMean}. We restate it for completeness and for comparison with ARLS sampling.
\new{In the following, we denote $\cL(\rkhs)$ the set of bounded linear operators from $\rkhs$ to itself, and $\noprkhs{·}$ the operator norm on $\cL(\rkhs)$.}

\begin{ttheorem}{}{bound_fix_lambda_uniform}
	Let \Cref{a:bounded_fmap,a:iid} hold. Let $12\leq m \leq n$ and let $\delta \in (0,1)$. When the $m$ sub-samples $\ldm{1}, …, \ldm{m}$ are drawn uniformly without replacement from the dataset $\{X_1 \dots, X_n\}$ and $w$ is chosen as in \eqref{e:def_weights}, 
	it holds with probability at least $1-δ$ that
	\begin{align}
		\wceH
		&≤ \frac{c_1}{\sqrt{n}} + \frac{c_2}{m} +  \frac{c_3\sqrt{\log(\nicefrac{m}{δ})}}{m} \sqrt{\deff[\frac{12\supk\log(\nicefrac{m}{δ})}{m}]},
		\label{e:bound_fix_lambda_uniform}
	\end{align}
	provided that 
	\begin{align*}
	    m ≥ \max(67, 12K^2\noprkhs{\covop}^{-1})\log\prt*{\frac{m}{δ}},
	\end{align*}
	where $c_1,c_2,c_3$ are constants of order $\supfmap \log(1/δ)$.
\end{ttheorem}

The constants $c₁,c₂,c₃$ are made explicit in the proof.
A few remarks regarding \Cref{r:bound_fix_lambda_uniform} are in order.
First, denoting by $W$ the smallest branch of the Lambert's $W$ function on $]-e^{-1},0[$ \pcite{weisstein2002lambert}, the condition on the sub-sample size $m$ can also be expressed as $m≥-W(-δ/c)c$ with $c=\max(67, 12K^2\noprkhs{\covop}^{-1})$ and can thus easily be checked numerically.

Then, the bound on the error is split in three parts: the first part corresponds to the usual rate one gets estimating the kernel mean embedding by its standard empirical counterpart, while the second part and the third part result from the approximation.  
Note that the first two terms already illustrate the trade-off between computational cost and statistical performance of our estimator: a small value of $m$ (i.e $m< \sqrt{n}$) will reduce the computational burden, but yield a rate worse than $O(1/\sqrt{n})$; alternatively, taking $m > \sqrt{n}$ would not improve the overall error rate, but would require more computational and storage resources.
%
The precise trade-off can be settled by the third term, which depends simultaneously on the subsample size $m$ and on the effective dimension $\deff$.
Extra assumptions about the effective dimension – which depends both on the kernel and the probability distribution – are needed to obtain a more explicit bound. 
We thus specialize our result under \Cref{a:poly_decay} and \Cref{a:exp_decay}, and present in both cases sufficient conditions on $m$ and $n$ to guarantee a $O(n^{-1/2})$ rate, and quantization rates \wrt $m$ that are faster than the Monte-Carlo $O(m^{-1/2})$ rate.

\begin{tcorollary}{Polynomial decay}{corollary_assumption_poly_decay}
Under the assumptions of \Cref{r:bound_fix_lambda_uniform}, if the RKHS $\rkhs$ and $\rho$ satisfy \Cref{a:poly_decay},
taking $m\de n^{1/(2-γ)}\log(n/δ)$ it holds
%
\[
	\wceH
	= O\prt*{ \frac{\log\prt*{ m }^{1-γ/2}}{m^{1-γ/2}} }.
\]
\end{tcorollary}

According to \Cref{rm:sobolev_decay}, we get the following result for Sobolev spaces.

\begin{tcorollary}{Sobolev space}{uniform_sobolev}
When $s>d/2$, under the assumptions of \Cref{r:bound_fix_lambda_uniform}, taking $m\de n^{1/(2-γ)}\log(n/δ)$ it holds
\[
	\wce[\sob] 
	= O\prt*{ \frac{\log\prt*{ m }^{1-d/(4s)}}{m^{1-d/(4s)}} }.
\]
\end{tcorollary}

The polynomial decay assumption always holds with $γ=1$, but no compression is achieved in this setting. However as soon as $γ<1$, we obtain rates that, despite not being optimal (the rate from \Cref{r:uniform_sobolev} should be compared to the optimal rate $O(m^{-s/d})$ for Sobolevs that will be achieved with ARLS sampling below), are already faster-than-\tiid and obtained at a really contained computational cost.  
The rate goes up to order $O(\log(m)/m)$ when $γ$ goes to zero, which corresponds to what we get when the spectrum of the covariance $\covop$ decays exponentially, as formalized in the next corollary.


\begin{tcorollary}{Exponential decay}{corollary_assumption_exp_decay}
Under the assumptions of \Cref{r:bound_fix_lambda_uniform}
and \Cref{a:exp_decay}, 
taking $m \de \sqrt{n}\log(\sqrt{n}c_4)$ where $c_4$ is a constant, it holds
%
\[ 
	\wceH
	= O\prt*{\frac{\log(m)}{m}}.
\]
\end{tcorollary}

The expresion of $c₄$ is provided in the proof, and this corollary holds for instance for the Gaussian kernel with a subgaussian probability distribution.  
Although not being optimal, these rates are nonetheless interesting because they still adapt to the spectral decay of the covariance operator, and thus outperform the standard $O(m^{-1/2})$ Monte-Carlo rate.
We also stress that uniform sampling is, obviously, computationally extremely efficient − the overall complexity becoming then dominated by the cost of computing the quadrature weights.
We will now show that improved rates can be obtained with leverage scores sampling.

\subsection{Rates for Ridge Leverage Scores Sampling}
\label{s:ls_sampling}

In this section, we present quantization rates for ARLS sampling (as defined in \Cref{ss:sampling_schemes}).
This result relies on a slightly different error decomposition \wrt to uniform sampling as detailed in \Cref{s:deterministic_dec}.

\begin{ttheorem}{}{main_result_bis}
	Let \Cref{a:bounded_fmap,a:iid} hold. Let the sub-samples $\tilde{X}_1, \dots, \tilde{X}_m$ be drawn with replacement proportionally to $(z,\lambda_0,\delta/6)$-approximate leverage scores from the dataset $\{X_1 \dots, X_n\}$, for some $z≥1,\lambda_0>0$, 
	and $w$ chosen as in \eqref{e:def_weights}. 
	Assume $n\geq (1655 + 233 \log(12\supk/δ))\supk$ and $λ₀ ≤ \tfrac{19\supk \log(\frac{8n}{δ})}{n}$. Then, we have the two following results, depending on the assumption on the eigenvalue decay.
	\begin{itemize}
		\item Under \Cref{a:poly_decay} (polynomial decay),
		choosing $m = n^γ (\log\frac{32n}{δ})^{1-γ} \frac{78 c z² }{(19\supk)^γ}$ guarantees that, with probability at least $1-δ$,
		\begin{align*}
			\wce
				&= O\prt*{\frac{\log(m)^{1/(2γ)}}{m^{1/(2γ)}} }\enspace,
		\end{align*}
		provided that $n$ is large enough, \ie,
			$\frac{19\supk \prt*{ \log\frac{32n}{δ}}}{n} ≤ \min\left(\noprkhs{\covop}, \prt*{\tfrac{\new{c_γ} z²}{5}}^{1/γ}\right)$ 
		\new{where $c_γ\de a_γ/(1-γ)$ when $γ<1$ and $c_γ\de \supk$ when $γ=1$.}
		\item Under \Cref{a:exp_decay} (exponential decay),
		choosing 
		\[  m = \max(334, 78 z² β^{-1})\log\prt*{\max(\tfrac{2a_β}{19\supk}, \tfrac{48}{δ}) n}² \]
		guarantees that, with probability at least $1-δ$,
		\begin{align*}
			\wce
				&= O\prt*{\frac{m^{1/4}}{\exp(\sqrt{m}/c)} }\enspace,
		\end{align*}
		where $c$ is a constant, 
		provided that $n$ is large enough: 
			 $\tfrac{19\supk\log(\frac{8n}{δ})}{n} ≤ \min(a_β,\noprkhs{\covop})$.
	\end{itemize}
\end{ttheorem}

We stress that the constant $c$ appearing in the rate for the exponential decay setting is independent on the dimension. 
As one can see from the rates, ARLS sampling allows us to reach better rates both for polynomial and exponential decay.
Again, the Sobolev case corresponds to a polynomial decay of the eigenvalues with $γ=d/(2s)<1$, and we thus obtain the rate $\wce[\sob] = O(\log(m)^{s/d} m^{-s/d})$ in this setting, which up to the logarithmic term matches the known optimal rates mentioned in \Cref{s:related_work}. 

Note that the condition on $λ₀$ can be satisfied by directly feeding the desired value to the algorithm used to estimate the approximate empirical leverage scores, and should therefore not be seen as a limitation.

\subsection{Faster Rates Under a Source Condition}
\label{s:source_condition}

While previous rates were uniform over the RKHS $\rkhs$, it is possible to obtain improved quadrature rates when considering fractional subspaces, \ie nested \new{subspaces} of $\rkhs$ of increasing smoothness. 
To our knowledge, this setting has never been studied in the literature so far. 

\begin{tdefinition}{Fractional Subspaces}{frac_subspaces}
	If $\rkhs$ is \new{an} RKHS with covariance operator $\covop$, the fractional subspace of smoothness $s$ of $\rkhs$ for the data distribution $\td$ is defined as 
	$\fs=\covop^{s}\rkhs$, and is endowed with the norm 
	$‖f‖_s=\nrkhs{g}$
	where $g$ is the unique function satisfying $g∊(\ker \covop)^\perp$ and $\covop^s g=f$.
\end{tdefinition}

Note that this definition depends on both $\rkhs$ and $\td$, \ie not only on the properties of the base RKHS but also on its interaction with the data distribution.
It is connected to the source condition hypothesis made in the inverse problem literature \pcite{engl2000RegularizationInverseProblems}; the difference in our setting is that we are not interested in one single function, but rather in bounding the quadrature error uniformly over such fractional subspaces.
\color{black}

The fractional subspaces are themselves reproducing kernel Hilbert spaces and one could apply the previous result directly to them and define their associated kernels. However, in practice the smoothness is often unknown, and we obtain in this section improved rates without the need to estimate this smoothness: in particular the leverage scores are computed with respect to the base kernel $\kr$.

\new{Such improved rates are also reminiscent of the so-called superconvergence results in kernel interpolation, see \eg~\textcite{schaback2018SuperconvergenceKernelbasedInterpolation} and \textcite[Sec. 11.5]{wendland2004ScatteredDataApproximation}.}

\begin{ttheorem}{}{source_condition_final_rate}
	Let $s∊[0,1/2]$. 
	Let \Cref{a:bounded_fmap} hold. 
	Furthermore, assume that the data points $X_1, \dots, X_n$ are drawn \tiid from the distribution $\td$ and that $m ≤ n$ sub-samples $\tilde{X}_1, \dots, \tilde{X}_m$ are drawn using $(z,λ₀,δ/4)$-approximate leverage scores sampling with replacement (for some $z≥1,λ₀>0$) from the dataset $\{X_1 \dots, X_n\}$. 
	Let $w$ chosen as in \eqref{e:def_weights}. 
	Assume that: 
	\begin{align*}
		n &≥ (1655 + 233 \log(8\supk/δ))\supk  \\
		 λ₀^{2s+1} &≤ \frac{19 \supk \log(\nicefrac{32 n}{δ})}{n} ≤ \min(1, \noprkhs{\covop}^{2s+1}). 
	\end{align*}
	\color{black}
	\begin{itemize}
		\item 
		Under \Cref{a:poly_decay} (polynomial decay), 
		taking $m=Θ\prt*{n^{γ/(2s+1)}\log(32n/δ)^{1-γ/(2s+1)}}$, 
		we get with probability $1-δ$ the rate 
		\begin{align*}
			\wce[\fs] 
			&= O(m^{-(2s+1)/(2γ)})
		\end{align*}
		provided that $n$ is large enough to additionally ensure
			$n ≥ 19 \supk \prt*{\tfrac{334}{78 z² \new{c_γ} }}^{(2s+1)/γ} \log\prt*{\nicefrac{32n}{δ}}$. 
%
		\item 
		Under \Cref{a:exp_decay} (exponential decay), 
		taking 
		\begin{align*}
			m &\de \max\prt*{\frac{c_m}{2s+1}\log\prt*{ c_m' n }, 334} \log\prt*{ c_m' n }=O(\log(n)²)\\
			\quad\text{where}\quad
			c_m &\de 78 z²β^{-1},
			c_m' \de \max\prt*{\tfrac{(2a_β)^{2s+1}}{19\supk}, \tfrac{32}{δ}}
		\end{align*}
		it holds with probability $1-δ$
		\begin{align*}
			\wce[\fs]
			&= O\prt*{ m^{1/4}\exp\prt*{-\tfrac{2s+1}{2\sqrt{c_m}}\sqrt{m}} },
		\end{align*}
		provided that $n$ is large enough to additionally ensure
			$n ≥ {19\supk a_β^{-(2s+1)} \log(\nicefrac{32 n}{δ})}$.
	\end{itemize}

\end{ttheorem}

Note that depending on the constants, the conditions on $n$ might always be satisfied, or reduce to lower bounds on $n$, but can always be satisfied for $n$ large enough.

We observe under the polynomial decay assumption an improved rate of $O(m^{-(2s+1)/(2γ)})$, which should be compared to the rate $O(m^{-1/(2γ)})$ that we obtained (up to log terms) in \Cref{s:ls_sampling}.
In the exponential decay setting, we still obtain an exponential dependence in $\sqrt{m}$, however the constant appearing inside the exponential is reduced due to the factor $2s+1$ and faster convergence can hence be obtained.

\section{Numerical Experiments}

\label{s:experiments}

In this section, we evaluate empirically the performance of our proposed method in two different setting.
In \Cref{s:psobolev}, we consider periodic Sobolev spaces on $[0,1]$ and a uniform target distribution, a setting which has been extensively used to benchmark quadrature methods, 
and in \Cref{s:real_data} we use real datasets on $\bR^d$ and consider spaces generated by Gaussian and Laplacian kernels.

\paragraph{Error computation} 
Note that the (squared) error of a quadrature rule $\nI$ for the reproducing kernel Hilbert space $\rkhs$ can be computed using \Cref{r:quad_to_kme} as follows:
\begin{align*}
	\wce[\rkhs]²
	&= 
	\nrkhs*{\int \fmap(x) \dif ρ(x) - \sum_{j=1}^m w_j \fmap{\ldm{j}} \, }² \\
	&= \iint \kr(x,y) \dif \td(x) \dif \td(y)
	   - 2 \sum_{1≤j≤m} w_j \int \kr(x,\ldm{j}) \dif \td(x) 
	   + wᵀ\kmatm w
   \numberthis\label{e:expr_error}
\end{align*}
where we recall that $\kmatm$ denotes the kernel matrix at the landmarks $\ldms$. 
Hence,
to compute the kernel mean embedding one only needs a closed form of the kernel $κ$ and the Nyström landmarks,
but to compute the error via \eqref{e:expr_error} one needs a closed form for $\int κ(x,\ldm{i})\dif \td(x)$ and $\iint κ(x,y) \dif \td(x)\dif\td(y)$.
If $\td$ has a discrete support of size $n$, then evaluating the error requires only kernel evaluations and scales in $Θ(n²)$.
For this reason, we restrict ourselves in \Cref{s:real_data} to datasets of moderate size, although the quadrature methods themselves do not suffer from this quadratic dependency in the dimension and could scale to larger datasets.

\new{Optimal weights are always used in this section, for all landmark selection strategies.}

\subsection{Periodic Sobolev Spaces}
\label{s:psobolev}

We consider $\dspace=[0,1]$ and the translation-invariant kernel
\begin{align*}
	κ_s(x,y)
	&\de 1 + 2 \sum_{n∊\bN^*} \tfrac{1}{n^{2s}} \cos(2πn(x-y))
	= 1 + \frac{(-1)^{s-1}(2π)^{2s}}{(2s)!} B_{2s}(\cb{x-y})
\end{align*}
where $B_{2s}$ denotes the Bernoulli polynomial of order $2s$ and $\cb{·}$ the fractorial part.
The expression involving Bernoulli polynomials is for instance mentioned in \pcite[p.22]{wahba1990SplineModelsObservational}.
The associated reproducing kernel Hilbert space corresponds to the Sobolev space of periodic functions of order $s$ satisfying \new{the boundary conditions $f^{(i)}(0)=f^{(i)}(1)$ for $i=0,…,s-1$}, and we choose for $\td$ the uniform distribution on $\dspace$.

It holds $\int_0^1 κ(x,\tilde{x})\dif x=\iint_0^1 κ(x,y)\dif x\dif y=1$ so the error can easily be computed using \Cref{e:expr_error}.
This RKHS has been used by multiple authors to benchmark quadrature methods
because the eigendecomposition of the covariance operator is computable exactly, 
and we thus include this setting for completeness.
However, we stress that 
the kernel mean embedding is the constant function $\me(x)=\int_0^1 κ(x,y)\dif y=1$ (using the definition of the kernel as sum of cosines), and the continuous ridge leverage scores (of which the the leverage scores defined in \eqref{e:def_ls} can be seen as a tractable approximation based on the empirical data) are uniform in this setting as observed by \textcite[Sec. 4.4]{bach2017EquivalenceKernelQuadrature}. As a consequence, no improvement over uniform sampling should be expected in this setting when using ARLS. 

\new{For $\dspace=[0,1]^d$ with $d>1$, we consider the product kernel $κ_s^d(x,y)=Π_{i=1}^d κ_s(x_i,y_i)$. This kernel does not induce a Sobolev space but rather consists in functions having square integrable mixed partial derivatives of order up to $s$ in each variable. The eigenvalues of the associated integral operator for the uniform distribution are known to decay in $(\log i)^{2s(d-1)}i^{-2s}$, see \eg \textcite{bach2017EquivalenceKernelQuadrature}.}

We compare our approach to the method of \pcite{belhadji2019KernelQuadratureDPPs} based on determinantal point processes sampling, as well as the method of \pcite{hayakawa2022PositivelyWeightedKernel} which relies like us on a Nyström approximation but uses a recombination algorithm.
We also include for comparison three greedy deterministic methods: greedy minimization of the norm of the residual $\n{P_m^\perp \eme}$, orthogonal matching pursuit, and greedy maximization of $\det(K_m)$.
Note that these three methods correspond in the kernel interpolation literature respectively to the so-called $f/P$-greedy, $f$-greedy and $P$-greedy methods applied on the function $\eme$.
For these methods, the non-convex optimization steps to select the new atoms are approximated by an exhaustive search over the empirical data.
We provide additional details regarding these methods in \Cref{s:greedy}.

We implemented our approach as well as the three greedy methods in Julia\footnote{See \url{https://gitlab.com/achatali/efficient-numerical-integration-in-rkhs-via-ls-sampling}, code released under the AGPL3 license.}, and rely on the Python authors' implementations of the other two methods.
All implementations however use OpenBLAS as BLAS implementation with the same number of threads, see \Cref{s:impl_details} for technical details.

\begin{figure*}
	\includegraphics[width=\linewidth]{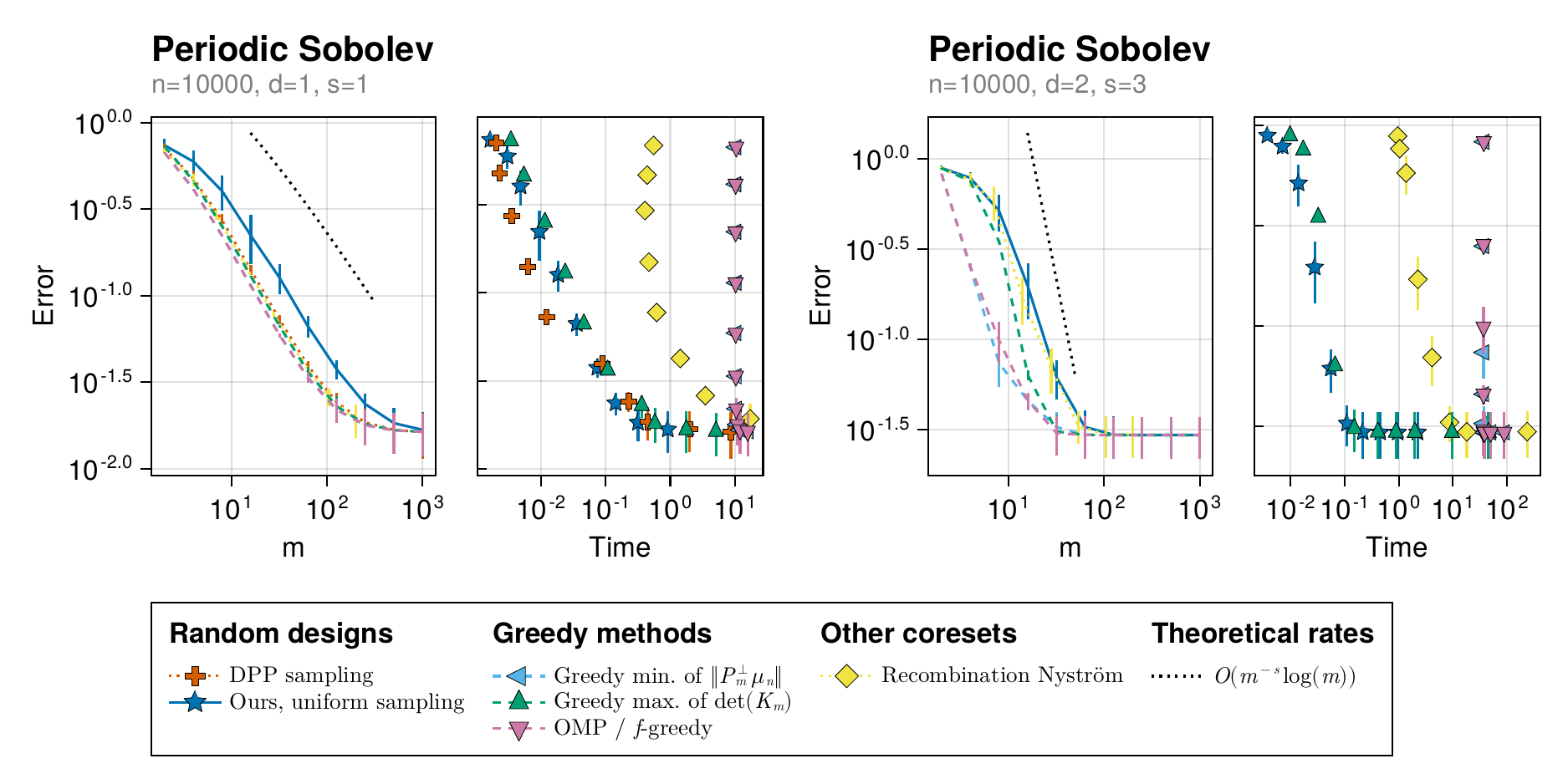}
	\caption{\label{f:psobolev}
	Periodic Sobolev setting. Medians and standard deviations over 50 trials.
	}
\end{figure*}

Results are reported in \Cref{f:psobolev} for $d=1,s=1$ and $d=2,s=3$.
We observe that all methods seem to roughly follow the optimal $O(m^{-s})$ rate \new{in dimension $d=1$.}
This is expected for our method by \Cref{r:summary_main_results} even though we are sampling uniformly, given that leverage scores are uniform in this setting. 
\new{For $d=2$, all methods appear to be slightly sub-optimal compared to the optimal theoretical rate, which is still $O(m^{-s})$ in this setting as discussed above.}
Although our method seems to 
suffer from 
a slightly larger error 
with respect to other methods for a fixed support size $m$
, it outperforms all of them when looking at the tradeoff between approximation error and runtime.
In particular, the three greedy methods suffer a lot from the linear search which is done at each iteration.
The method from \pcite{belhadji2019KernelQuadratureDPPs} is competitive with our approach in terms of accuracy-runtime tradeoff for $d=1,s=1$, but requires the knowledge of the covariance's eigendecomposition which is highly limiting for applications beyond this setting.
Greedy maximization of $\det(K_m)$ seems to yield a better convergence rate than our method at a moderate computational cost, however this method is not adaptive to the target distribution and we will show in the following experiment that it performs poorly for a non-uniform distribution.

\color{black}

\subsection{OpenML Datasets with Gaussian and Laplacian Kernels}
\label{s:real_data}

\begin{figure*}
	\includegraphics[width=\linewidth]{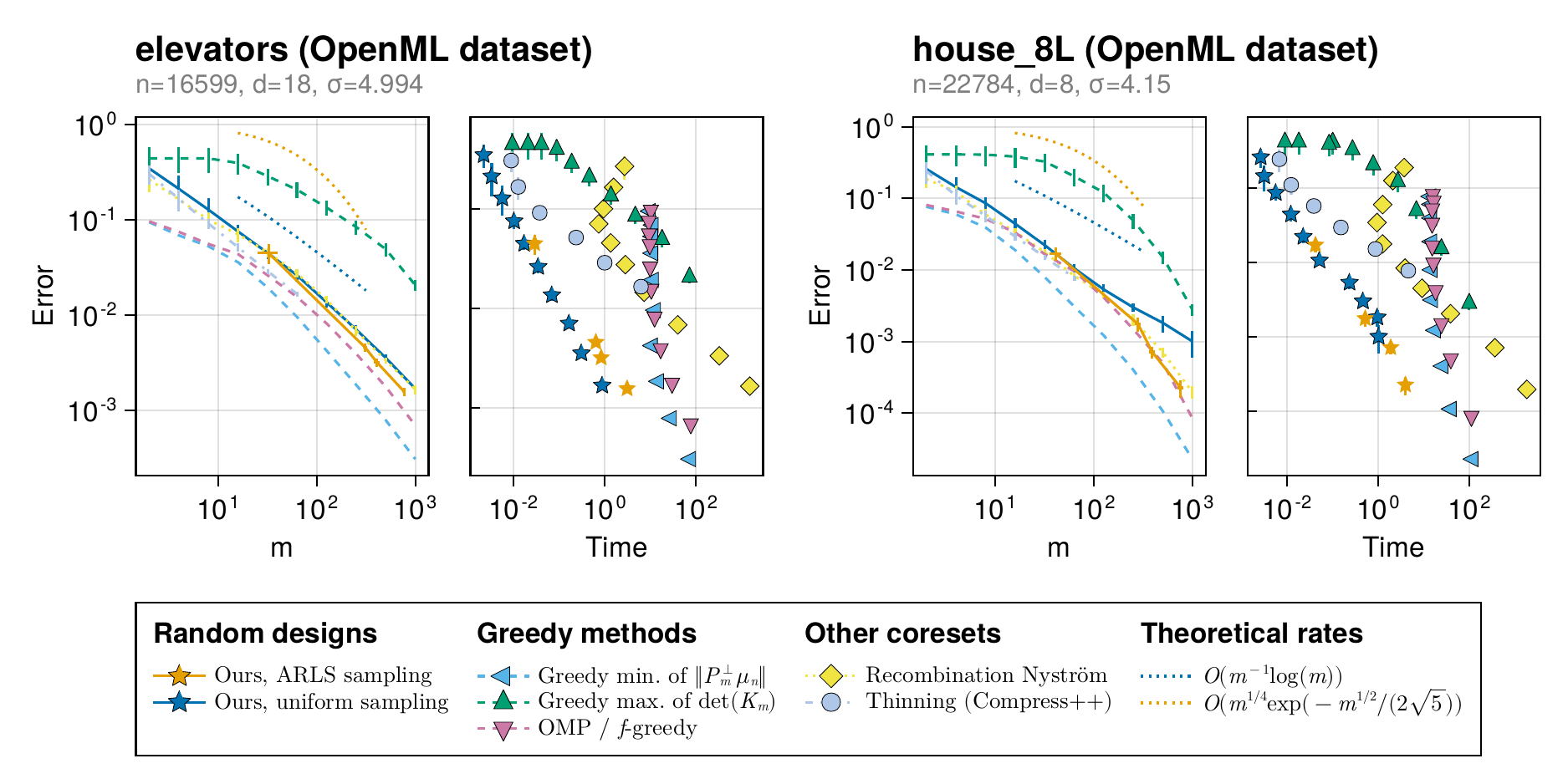}
	\caption{\label{f:gkernel}
	Gaussian kernel for two OpenML datasets. 
	Medians and standard deviations over 40 trials.
	}
\end{figure*}

We consider in this section multiple machine learning datasets from the OpenML database\footnote{\url{https://www.openml.org/}}.
To better see the rates of the different methods, we do not use data splitting and report the error computed using \eqref{e:expr_error} taking $\td$ to be the discrete measure corresponding to the full dataset $\td\de \tfrac{1}{n} \sum_{i=1}^n \fmap{x_i}$. 

\begin{figure*}
	\includegraphics[width=\linewidth]{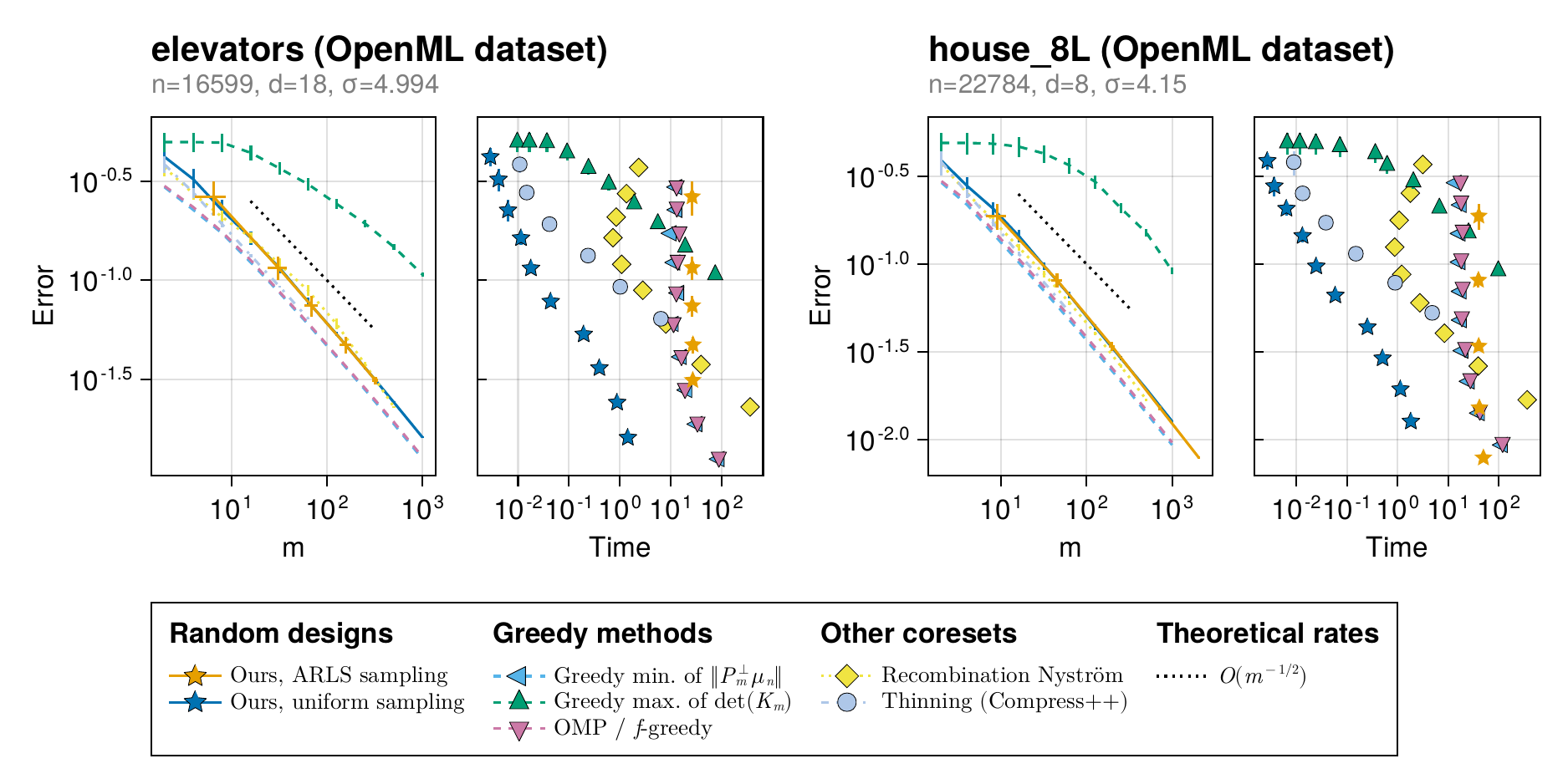}
	\caption{\label{f:laplacian}
	Laplacian kernel for two OpenML datasets. 
	Medians and standard deviations over 30 trials.
	}
\end{figure*}

We report here the error as a function of both the number of nodes $m$ and running times, for the Gaussian (\Cref{f:gkernel}) and Laplacian (\Cref{f:laplacian}) kernels and for two datasets, but additional results on a wider selection of datasets are provided in \Cref{s:additional_exp_results}.
The kernel scale is fixed by computing the median inter-point euclidean distance on a random subset of the data, and its value is reported on the figures for each dataset.
We compare our methods to the algorithms mentioned in \Cref{s:psobolev}, at the exception of the methods which rely on the Mercer decomposition, as the latter is unknown in this setting. 
We also include the thinning method of \pcite{shetty2022DistributionCompressionNearlinear}, for which we take as oversampling parameter $g=4$, which corresponds to the author's choice in their experimentations, and start building the coreset from $m²≤n$ samples drawn iid and uniformly from the dataset. 
Additional technical details are provided in \Cref{s:impl_details}.

\new{In \Cref{f:gkernel}, }
we plot in dotted line the theoretical rates predicted by \Cref{r:summary_main_results} \new{under \Cref{a:exp_decay}}, picking for the exponential rate a constant matching the observations.
We observe that on these two datasets uniform sampling indeed yields a fast $O(m^{-1})$ rate.
Leverage scores sampling improves the converge rate as predicted by theory, however this is observed in practice only when $m≥100$; it should be noted however in this setting that (i) the tails of the target distribution might be too heavy to satisfy the hypotheses and (ii) the exponential decay is conditioned in \Cref{r:main_result_bis} to having $n=\exp(m^{1/2})$, which is not satisfied for the larger values of $m$ used in the plot as computing the error exactly would become prohibitive on very large datasets. 

Here again, when looking at the error as a function of runtime, we see that our approach outperforms all the others algorithms.
It is clear that the method which greedily fills the space in a uniform manner, which seemed to be competitive in the Sobolev setting, yields here a really poor accuracy; this should be expected as this method is not adaptive to the target distribution.

With a Laplacian kernel $\kr(x,y)=\exp(-λ‖x-y‖)$, we do not observe any different between uniform and ARLS sampling, which matches our theoretical guarantees. Indeed due to the lack of smoothness of the Laplacian kernel, we expected to observe the rates for exponential decay with the weakest hypotheses (\Cref{a:poly_decay} with $γ→1$), which yields a rate of order $O(m^{-1/2})$ (\ie no better than Monte-Carlo) for both uniform and ARLS sampling.
All methods achieve the same rate, with slightly smaller constants for greedy methods − still at the price of a much larger computational cost.

\color{black}

\color{black}

\section{Conclusion}
\label{s:conclu}

In this article, we introduced an efficient 
quadrature method based on random subsampling,
which is related to the Nyström approximation used for the discretization of linear integral equations and to build low-rank approximations of kernel matrices. 
We derived worst-case error bounds for RKHS for both uniform and approximate ridge leverage scores sampling, and showed that optimal rates can be obtained for Sobolev spaces in the latter case.
Empirically, we showed that our method outperforms the state of the art in terms of accuracy-runtime tradeoff. 
Studying the performance of our approach in the misspecified setting, \ie when the integrand do not belong to the considered RKHS, would be of interest for future works.


\section*{Acknowledgments} 
\new{We would like to thank the anonymous referees as well as Lester Mackey and Mathias Sonnleitner for their helpful and constructive comments.} 
Lorenzo Rosasco acknowledges the financial support of the European Research Council (grant SLING 819789), the AFOSR projects FA9550-18-1-7009, FA9550-17-1-0390 and BAA-AFRL-AFOSR-2016-0007 (European Office of Aerospace Research and Development), the EU H2020-MSCA-RISE project NoMADS - DLV-777826, and the Center for Brains, Minds and Machines (CBMM), funded by NSF STC award CCF-1231216.
    \new{The research by Ernesto De Vito and Lorenzo Rosasco has been supported by the MIUR grant PRIN 202244A7YL. The research by Ernesto De Vito has been supported by the MIUR Excellence Department Project awarded to Dipartimento di Matematica, Università di Genova, CUP D33C23001110001. Ernesto De Vito is a member of the Gruppo Nazionale per l’Analisi Matematica, la Probabilità e le loro Applicazioni (GNAMPA) of the Istituto Nazionale di Alta Matematica (INdAM).} %

\newpage
\appendix

\section*{Structure of the Appendix}

We begin by introducing additional notations in \Cref{a:notations}. Then, we prove in \Cref{s:computation_weights} the expression of optimal weights.
A deterministic error decomposition is derived in \Cref{s:deterministic_dec}, and then used in \Cref{a:proof_main_result} to prove our main results. 
\Cref{a:concentration} contains the concentration results that our proof of \Cref{r:bound_fix_lambda_uniform} rely on, and we also recall in \Cref{s:aux_results} some key results on the effective dimensions and the Nyström approximation. 
Eventually we provide additional details regarding numerical experiments in \Cref{s:experiments_appendix}.

\section{Table of Notations}
\label{s:table_notations}

\begin{tabular}{ll}
	\toprule
	$\dspace$
		& Input space \\
	$\rkhs$
		& Generic RKHS \\
	$\sob$
		& Sobolev space (see \Cref{ex:sobolev}) \\
	$\fs$
		& Subspace of $\rkhs$ corresponding to a source condition \\
	\midrule
	$\td$ 
		& Target/data distribution \\
	\midrule
	$\ldms$
		& Quadrature nodes (= Nyström landmarks in our case) \\
	$w$
		& Quadrature weights \\ 
	$\nI$
		& Quadrature rule \\
	\midrule
	$\covop:\rkhs → \rkhs$
		& (Uncentered) covariance operator \\
	$(σ_i)_{i∊\bN}$
		& Eigenvalues of $\covop$ \\
	$γ,a_γ$
		& Parameter and constant for polynomial decay (\Cref{a:poly_decay}) \\
	$β,a_β$
		& Parameter and constant for exponential decay (\Cref{a:exp_decay}) \\
	\midrule
	$\wce$
		& Worst-case quadrature error on the unit ball of $\cF$ (cf. \eqref{e:wce}) \\
	\bottomrule
\end{tabular}

\section{Additional Notations}
\label{a:notations}

We define the operator $\mft : \ltsp→\rkhs$ for any $f∊\ltsp$ as 
\begin{align*}
	\mft f = \int_{\dspace} f(x)\fmap(x) \dif\td(x). 
\end{align*}
Its adjoint $\amft$ is defined by $\amft h = \iprkhs{h, \fmap{·}}$ for any $h∊\rkhs$ and corresponds to the inclusion operator from $\rkhs$ into $\ltsp$.

We define the (uncentered) covariance operator $\covop: \rkhs\rightarrow \rkhs$ as
\begin{align*}
\covop &\coloneqq \int \fmap{x} \kron \fmap{x} d\td(x)
\end{align*}
where $(\fmap{\Vx} \kron \fmap{\Vx})(f) \de \iprkhs{f, \fmap{\Vx}} \fmap{\Vx}$.
One can easily check that $\covop=\mft\amft$.
Moreover, \Cref{a:bounded_fmap} implies that the operator $\covop$ is a positive trace class operator on $\rkhs$ and allows to leverage tools from spectral theory. 
\new{Positivy derives from the fact that $\fmap(x) \kron \fmap(x)$ is positive for any $x$.
Indeed, for any $f∊\rkhs$, 
by applying twice Hille's theorem \pcite[Theorem 6]{diestel1977VectorMeasures}
on the linear bounded (and thus closed~\pcite[4.13.5 (a)]{kreyszig1989IntroductoryFunctionalAnalysis}) operators $M↦Mf$ and $v↦\ip{v,f}$
\begin{align}
	\ip{Cf,f}
		&= \ip*{\prt*{\int \fmap(x) \kron \fmap(x) \dif\td(x)}f,f} \\
		&= \ip*{\int \fmap(x) f(x) \dif\td(x),f} \\
		&= \int f(x)² \dif\td(x) 
		 ≥ 0.
\end{align}
}

The empirical covariance operator is defined as
\begin{align*}
\ecovop &= \sum_{i=1}^n \fmap{X_i } \kron \fmap{X_i}.
\end{align*}

For any operator $Q : \mathcal{H} : \rightarrow \mathcal{H}$ and any real number $λ > 0$, we denote by $Q_λ : \mathcal{H} \rightarrow \mathcal{H}$ the regularized operator $Q_λ = Q + λ I$.
We denote the (Moore-Penrose) pseudo-inverse of an operator $A$ by $A^+$. 

Given a random variable $X$,  we write $\esssup X$ to denote its essential supremum.

We write $1_n \in \mathbb{R}^n$ for the $n$-dimensional vector of ones.

We recall the definition of the effective dimension, and also introduce the notation $\dsup$:
\begin{subequations}
\begin{align}
\deff 
	&\de\E_{x\sim\td} \n{\covopl^{-1/2}\fmap(x)}²
	=\Tr(\covop \covopl^{-1}), \\
\dsup 
	&\de\esssup_{x\sim\td} \n{\covopl^{-1/2}\fmap(x)}².
\end{align}
\end{subequations}
It holds for any $λ>0$ that $\deff ≤ \dsup ≤\supk/λ<∞$.

\section{Derivation of the Weights}
\label{s:computation_weights}

This section provides a proof for the expression of the optimal weights claimed in \Cref{e:def_weights}. For ease of exposition, let us introduce the operators 
\begin{align*}
	\mftldms: \bR^m &\rightarrow  \rkhsm,\, w \mapsto  \sum_{j=1}^m w_j \fmap(\ldm{j}),\\
	\mftx: \bR^n &\rightarrow  \rkhs,\, w \mapsto  \sum_{i=1}^n w_i \fmap(X_i ).
\end{align*}
Since, by definition, $\nyseme$ is the orthogonal projection of $\eme$ onto the space $\rkhsm$, it can be expressed as $\nyseme= \mftldms w^*$ where the weights $w^* \in \bR^m$ minimize the mapping $w \mapsto \nrkhs{\eme-\mftldms w}^2$. Setting the gradient of this mapping to zero, we obtain that $w$ must satisfy
\begin{align*}
	\mftldms^*\mftldms w &= \mftldms^* \eme.
\end{align*}
The minimum norm solution of the above equation is given by $w =(\mftldms^*\mftldms)^+ \mftldms^* \eme$~\pcite{laub2004MatrixAnalysisScientists}.
Noting that the empirical kernel mean embedding $\eme$ can be expressed as $\eme=\frac{1}{n}\mftx 1_n$ and using the fact that $\mftldms^*\mftldms=\kmatm$, $\mftldms^* \mftx = \kmatmn$, we obtain the claimed equality
\begin{align*}
	w &= \kmatm^+ \mftldms^* (n^{-1}\mftx 1_n)
	= \frac{1}{n} \kmatm^+ \kmatmn 1_n.
\end{align*}

\section{Deterministic Error Bound}
\label{s:deterministic_dec}

In order to break down the approximation error, we introduce the quantity 
\[ \emem = \frac{1}{m}\sum_{j=1}^m \fmap{\ldm{j}}∊\rkhsm, \]
which is an unbiased estimate of the empirical kernel mean embedding $\eme$ when sampling uniformly the landmarks.

Our main results rely on the following deterministic error decompositions.
\begin{tlemma}{Error decomposition}{error_decomposition}
	For any $λ>0$, it holds (almost surely)
	\begin{align}
	\nrkhs{\me  - \nyseme} 
		&\leq  \nrkhs{\me  - \eme} + \noprkhs{\Pmo\covopl^{1/2}}\nrkhs{\covopl^{-1/2}(\eme-\emem)}
		\label{e:error_decomposition} \\
	\nrkhs{\me  - \nyseme} 
		&≤ \nrkhs{\me - \eme} + \noprkhs{\Pmo\covopl^{1/2}}.
		\label{e:error_decomposition_bis}
	\end{align}
\end{tlemma}


While the decomposition \eqref{e:error_decomposition} is convenient, it is not well suited for the analysis when sampling proportionally to leverage scores as described in \Cref{ss:sampling_schemes}, and we will see that the decomposition \eqref{e:error_decomposition_bis} is easier to work with in this setting.
\begin{tproofof*}{r:error_decomposition}{}
	We rely for both inequalities on the decomposition
	\begin{align*}
		\nrkhs{\me  - \nyseme} 
		&\leq  \nrkhs{\me  - \eme} + \nrkhs{\eme - \nyseme}
	\end{align*}

	\paragraph{First bound \eqref{e:error_decomposition}}
	Note that
	\[
	\nrkhs{\eme - \nyseme}
	= \nrkhs{\Pmo\eme}
	= \nrkhs{\Pmo(\eme-\emem)}
	\]
	where the last inequality follows from $\Pmo\emem=0$.
	Hence we get
	\begin{align*}
		\nrkhs{\me  - \nyseme} 
		&\leq  \nrkhs{\me  - \eme} + \nrkhs{\Pmo(\eme-\emem)} \\
		&\leq  \nrkhs{\me  - \eme} + \noprkhs{\Pmo\covopl^{1/2}}\nrkhs{\covopl^{-1/2}(\eme-\emem)}.
	\end{align*}


	\paragraph{Second bound \eqref{e:error_decomposition_bis}}
	We use the alternative decomposition
	\begin{align*}
		\nrkhs{\me - \nyseme}
		&= \nrkhs{\me - \Pm \eme} \\
		&≤ \nrkhs{\me - \Pm \me} + \nrkhs{\Pm(\me-\eme)} \\
		&≤ \noprkhs{\Pmo\covopl^{1/2}} \nrkhs{\covopl^{-1/2}\me} + \nrkhs{\me-\eme} 
	\end{align*}
	Note that because $\me$ is a mean embedding, it can be written $\me = \mft 1$ where $1∊\ltsp$ denotes the constant function, and $\mft$ admits a polar decomposition of the form $\mft=\covop^{1/2}U$ where $U$ is a partial isometry from $\ltsp$ to $\rkhs$. Hence we have
	\begin{align*}
		\nrkhs{\covopl^{-1/2} μ} 
		&= \nrkhs{\covopl^{-1/2} \covop^{1/2}U1} 
		≤  \noprkhs{\covopl^{-1/2} \covop^{1/2}}\nltsp{1} ≤ 1.
	\end{align*}
\end{tproofof*}

\section{Proofs of the Main Results}
\label{a:proof_main_result}

\subsection{Proofs for Uniform Sampling (\Cref{s:uniform_sampling})}

Theorem~\ref{r:bound_fix_lambda_uniform} is a consequence of a more general result which we state now.

\begin{ttheorem}{}{probabilistic_bound_lambda}
		Let Assumption~\ref{a:bounded_fmap} hold. Furthermore, assume that the data points $X_1, \dots, X_n$ are drawn \tiid from the distribution $\td$ and that $m \leq n$ sub-samples $\tilde{X}_1, \dots, \tilde{X}_m$ are drawn uniformly with replacement
		from the dataset $\{X_1 \dots, X_n\}$. Then, for any $λ∊]0,\noprkhs{\covop}]$ and $δ∊]0,1[$, with probability at least $1-δ$
	\begin{align*}
		\nrkhs{\mu  - \nyseme} 
		&≤
		\frac{2\supfmap \sqrt{2\log(6/δ)}}{\sqrt{n}} + 
		\sqrt{λ} \prt*{\frac{4\sqrt{3\dsup}\log(12/δ)}{m} + 6\sqrt{\frac{\deff \log(12/δ)}{m}}},
	\end{align*}
	provided that 
	\begin{itemize}
		\item $m\geq \max(67, 5\dsup)\log\prt*{\frac{12\supk}{λδ }}$,
		\item $λn≥12\supfmap²\log(\nicefrac{12}{δ})$. 
	\end{itemize}
\end{ttheorem}
\begin{tproofof*}{r:probabilistic_bound_lambda}{}
	Let $δ \in (0, 1)$ be the desired confidence level. Let $λ>0$, $m \in \mathbb{N}$ and $n \in \mathbb{N}$ satisfy the conditions of the theorem.
	Using the error decomposition of \Cref{r:error_decomposition}, we get
	\begin{align*}
		\nrkhs{\mu  - \nyseme} &\leq  \nrkhs{\mu  - \eme} + \noprkhs{\Pmo\covopl^{1/2}}\nrkhs{\covopl^{-1/2}(\eme-\emem)}.
	\end{align*}
	Controlling the first term amounts to measuring the concentration of an empirical mean around its true mean in a Hilbert space. Multiple variants of such results can be found in the literature (see, \eg, \pcite{pinelis1994optimum}).
	We apply here Lemma~\ref{r:concentration_iid_hilbert} on the random variables $\eta_i \coloneqq \fmap(X_i)-\mu, 1 \leq i \leq n$. Note that they are indeed bounded since, for any index $1\leq i \leq n$, $\nrkhs{\eta_i}\leq 2\sup_{x\in\dspace}\nrkhs{\fmap(x)} = 2\supfmap$. Thus, it holds with probability at least $1-\nicefrac{δ}{3}$ on the draw of the the dataset $X_1, \dots, X_n$ that
	\begin{equation*}
	    \nrkhs{\mu  - \eme} \leq  \frac{2\supfmap \sqrt{2\log(6/δ)}}{\sqrt{n}}.
	\end{equation*}

	Next,  we rely on \Cref{r:plain_nys_approx} to bound the term $\noprkhs{\Pmo \covopl^{1/2}}$ with high probability. Since the Nyström landmarks are uniformly drawn and $m\geq \max(67, 5\dsup)\log\frac{12\supk}{λδ}$, we have, for any $λ>0$, with probability at least $1-\nicefrac{δ}{3}$ on the draw of the landmarks $\tilde{X}_1, \dots, \tilde{X}_m$,
	\[ \noprkhs{\Pmo \covopl^{1/2}}  \leq  \sqrt{3λ}. \]
	
	Finally, the last term can be bounded using \Cref{r:bound_mean_embeddings_with_rgcovmhalf} which implies that, since $λ$ satisfies $0<λ≤\noprkhs{\covop}$ and $λn≥12\supfmap²\log(4/δ)$, it holds with probability at least $1-\nicefrac{δ}{3}$
	\[ \norm*{C_λ^{-1/2}(\eme - \emem)} ≤ \frac{4\sqrt{\dsup}\log(12/δ)}{m} + \sqrt{\frac{12\deff \log(12/δ)}{m}}. \]
	Taking the union bound over the three events yields the desired result: with probability at least $1-δ$ (over all sources of randomness), it holds that
	\begin{align*}
		\nrkhs{\mu  - \nyseme} 
		&≤ \frac{2\supfmap \sqrt{2\log(6/δ)}}{\sqrt{n}} + 
		\sqrt{3λ} \prt*{\frac{4\sqrt{\dsup}\log(12/δ)}{m} + \sqrt{\frac{12\deff \log(12/δ)}{m}}}.
	\end{align*}
\end{tproofof*}

\begin{tproofof*}{r:bound_fix_lambda_uniform}{}
	Assuming that our choice of $m$ and $λ$ satisfies the constraints
	\begin{align}
	\label{hyps_for_lemma}
		\left\{ \begin{array}{l}
			m\geq \max(67, 5\dsup)\log\frac{12\supk}{λδ }\\
			λn≥12\supfmap²\log(12/δ)\\
			0<λ \leq \noprkhs{\covop} 
		\end{array} \right.,
	\end{align}
	we can apply \Cref{r:probabilistic_bound_lambda} and use the fact that $\dsup≤\supk/λ$ to get
	\begin{align*}
		\nrkhs{\mu  - \nyseme} 
		&≤ \frac{2\supfmap \sqrt{2\log(6/δ)}}{\sqrt{n}} + \frac{4\sqrt{3}\supfmap\log(12/δ)}{m} + 6\sqrt{\log(12/δ)}\sqrt{\frac{λ\deff}{m}}.
	\end{align*}
	Setting $λ=\frac{12\supk\log(m/δ)}{m}$ we obtain by \Cref{r:quad_to_kme}
	the claimed result with constants
	$c_1=2\supfmap \sqrt{2\log(6/δ)}$, 
	$c_2=4\sqrt{3} \supfmap \log(12/δ)$, and
	$c_3=12\sqrt{3\log(12/δ)}\supfmap$.

    Let us now check that our choices are consistent with the constraints. We will also obtain a more user-friendly expression for the constraints and express the sub-sample size $m$ as a function of the sample size $n$. Using the fact that $\dsup≤\supk/λ$, one can easily check that a sufficient set of conditions to satisfy \eqref{hyps_for_lemma} is given by
	\begin{align*}
	\left\{ \begin{array}{l}
		m\geq 67\log\prt*{\frac{1}{δ}\frac{m}{\log(m/δ)}} \\
		m≥ \frac{5m}{12 \log\prt*{ \frac{m}{δ} }} \log\prt*{\frac{1}{δ} \frac{m}{\log(m/δ)} } \\
		\frac{\log(12/δ)}{n} ≤ \frac{\log(m/δ)}{m} \\
		12K^2 \frac{\log(m/δ)}{m} ≤ \noprkhs{\covop}
	\end{array} \right. .
	\end{align*}
	As $m≤n$, the third condition is satisfied as soon as $m≥12$.
	Moreover, with this choice of $m$, we have $\log(m/δ)>1$, hence the second constraint always holds and it is sufficient to show that
	\begin{equation*}
		m ≥ \max(67, 12K^2\noprkhs{\covop}^{-1})\log\prt*{\frac{m}{δ}}.
	\end{equation*}


\end{tproofof*}

\begin{tproofof*}{r:corollary_assumption_poly_decay}{}
Under \Cref{a:poly_decay}, by \Cref{r:deff_poly} it holds $\deff≤c_γλ^{-γ}$. 
Under the assumptions of \Cref{r:bound_fix_lambda_uniform}, setting $m\de n^{1/(2-γ)}\log(n/δ)$, we get
\begin{align}
	\label{mainthmineq}
	\wceH
	= \nrkhs{\mu  - \nyseme} 
	&≤ \frac{c_1}{\sqrt{n}} + \frac{c_2}{m} +  \frac{c_3\sqrt{\log(\nicefrac{m}{δ})}}{m} \sqrt{\deff[\frac{12\supk\log(\nicefrac{m}{δ})}{m}]} \\
	&≤ \frac{c_1}{\sqrt{n}} + \frac{c_2}{m} + c_3 \sqrt{c_γ} (12\supk)^{-γ/2}  \frac{\log(m/δ)^{\frac{1-γ}{2}}}{m^{\frac{2-γ}{2}}} \\
	&= O\prt*{\frac{\log(m/δ)^{1-\frac{γ}{2}}}{m^{1-\frac{γ}{2}}}} 
\end{align}


\end{tproofof*}

\begin{tproofof*}{r:corollary_assumption_exp_decay}{}
Under \Cref{a:exp_decay}, it holds by \Cref{r:deff_exp} $\deff≤\log(1+a_β/λ)/β$.
We apply \Cref{r:bound_fix_lambda_uniform}. 
Taking $m \geq \frac{12K^2\log(m/δ)}{a_β}$, and using the fact that $\log(1+x)\leq\log(2x)$ for $x>1$, the last term of \eqref{e:bound_fix_lambda_uniform} can be bounded by
\begin{align*}
\frac{\sqrt{\log(m/δ)}}{\sqrt{\beta}m}\sqrt{\log\left(1+\frac{a_β m}{12K^2\log(m/δ)}\right)} &\leq \frac{1}{\sqrt{\beta}m}\sqrt{\log(m/δ)\log\left(\frac{a_β m}{6K^2\log(m/δ)}\right)}\\
&\leq \frac{1}{\sqrt{\beta}m}\log(m\max(1/δ, a_β/(6K^2)))
\end{align*}
which is bounded by $\frac{1}{\sqrt{\beta n}}$ by taking $m\de\sqrt{n}\log(\sqrt{n}c_4)$ with $c_4\de\max(1/δ, a_β/(6K^2))$.
Plugging the latter bound in \eqref{e:bound_fix_lambda_uniform}, we obtain
\begin{align*}
	\wce
	&≤ \frac{c_1}{\sqrt{n}} + \frac{c_2}{\sqrt{n}\log(\sqrt{n}\max(1/δ, a_β/(6K^2))} + \frac{c_3}{\sqrt{\beta n}} = O\left(\frac{1}{\sqrt{n}}\right).\\
\end{align*}
The claimed quantization rate follows 
\begin{align*}
	\frac{1}{\sqrt{n}}
	&≤ \frac{\log(c_4\sqrt{n}\log(c_4\sqrt{n})))}{\sqrt{n}\log(c_4\sqrt{n})}
	= O\prt*{\frac{\log(m)}{m}}.
\end{align*}
\end{tproofof*}	

\subsection{Proofs for Leverage Scores Sampling (\Cref{s:ls_sampling})}

\begin{ttheorem}{}{probabilistic_bound_lambda_bis}
	Let \Cref{a:bounded_fmap,a:iid} hold. Let $\delta \in (0,1)$. Let the $m$ sub-samples $\tilde{X}_1, \dots, \tilde{X}_m$ be drawn according to $(z, \lambda_0, \delta/4)$-approximate leverage scores from the dataset $\{X_1 \dots, X_n\}$ for some $z \geq 1$ and $\lambda_0>0$. Then, for any $λ \in (λ₀ \vee \frac{19\supk}{n}\log(\frac{8n}{δ}),\noprkhs{\covop}]$, it holds, with probability at least $1-δ$,
	\begin{align*}
		\nrkhs{\mu  - \nyseme} 
		&≤ \frac{2\supfmap \sqrt{2\log(4/δ)}}{\sqrt{n}} + \sqrt{3\lambda}\enspace,
	\end{align*}
	provided that
	\begin{itemize}
		\item $n\geq (1655 + 233 \log(8\supk/δ))\supk$;
		\item $m ≥ \max\prt*{334, 78 z² \deff}\log\frac{32n}{δ}$.
	\end{itemize}
\end{ttheorem}
\begin{tproofof*}{r:probabilistic_bound_lambda_bis}{}
	Let the assumptions of the theorem hold.
	Let $δ \in (0, 1)$ be the desired confidence level.  Let the integers $m \in \mathbb{N}$ and $n \in \mathbb{N}$ satisfy the conditions of the theorem and let $λ \in (λ₀ \vee \frac{19\supk}{n}\log(\frac{8n}{δ}),\noprkhs{\covop}]$.
	Recall the error decomposition from \Cref{e:error_decomposition_bis},
	\begin{align*}
		\nrkhs{\me  - \nyseme} 
		&≤ \nrkhs{\me - \eme} + \noprkhs{\Pmo\covopl^{1/2}}.
	\end{align*}
	
	We apply here Lemma~\ref{r:concentration_iid_hilbert} on the random variables $\eta_i \coloneqq \fmap(X_i)-\mu, 1 \leq i \leq n$. Note that they are indeed bounded since, for any index $1\leq i \leq n$, $\nrkhs{\eta_i}\leq 2\sup_{x\in\dspace}\nrkhs{\fmap(x)} = 2\supfmap$. Thus, it holds with probability at least $1-\delta/2$ on the draw of the the dataset $X_1, \dots, X_n$ that
	\begin{equation*}
		\nrkhs{\me - \eme} ≤ \frac{2\supfmap \sqrt{2\log(4/\delta)}}{\sqrt{n}}\enspace.
	\end{equation*}
	
	Next,  we rely on \Cref{r:als_nys_approx} to bound the term $\noprkhs{\Pmo \covopl^{1/2}}$ with high probability. 
	Since the sub-samples are drawn according to $(z,λ₀,\delta/4)$-approximate leverage scores from the full dataset $\{X_1, \dots, X_n\}$, we have, with probability at least $1-\delta/2$ on the draw of the sub-samples $\tilde{X}_1, …, \tilde{X}_m$,
	\begin{align*}
		\noprkhs{\Pmo \covopl^{1/2}} ≤ \sqrt{3λ} \enspace.
	\end{align*}
	
	
	Taking the union bound over the two events yields the claimed result.
\end{tproofof*}

We now justify how the parameters $λ,m$ are chosen to yield the result claimed in \Cref{r:main_result_bis}.

\begin{tproofof*}{r:main_result_bis}{}
	We apply \Cref{r:probabilistic_bound_lambda_bis}, use the fact that $\deff≤\dsup≤\supk/λ$ 
	to get (without hypotheses on the eigenvalues decay)
	\begin{align}
		\nrkhs{\mu  - \nyseme} 
		&≤ \frac{2\supfmap \sqrt{2\log(4/δ)}}{\sqrt{n}} + 
		\sqrt{3λ}
		\label{e:main_result_bis_error_bound}
	\end{align}

	We now need to pick $m$ and 
	$λ$
	that ensure
	\begin{subequations}\label{e:hyps_for_probabilisic_bound_lambda_bis_als}
    \begin{align}[left = \empheqlbrace\,]
			λ₀ &< λ ≤ \noprkhs{\covop}  \\
			λ &≥\textstyle \frac{19\supk}{n}\log(\frac{8n}{δ})
				\label{e:hyps_for_probabilisic_bound_lambda_bis_als_1} \\
			m &≥ \textstyle 334\log\prt*{\frac{32n}{δ}} 
				\label{e:hyps_for_probabilisic_bound_lambda_bis_als_2} \\
			m &≥ \textstyle 78 z² \deff \log\prt*{\frac{32n}{δ}}
				\label{e:hyps_for_probabilisic_bound_lambda_bis_als_3}
    \end{align}
	\end{subequations}

	\paragraph{Polynomial decay} Under \Cref{a:poly_decay}, by \Cref{r:deff_poly} it holds $\deff ≤ c_γ λ^{-γ}$ for some constant $c_γ>0$. 
	In this setting, a sufficient condition to satisfy \eqref{e:hyps_for_probabilisic_bound_lambda_bis_als_3} is to take
	\begin{align}
			λ &\de \prt*{\frac{78 c_γ z² \log\frac{32n}{δ}}{m}}^{1/γ}.
			\label{e:poly_choice_lambda}
	\end{align}
	One can easily check that choosing additionally
	\begin{align}
		m \de n^γ \frac{78 c_γ z² (\log\frac{32n}{δ})^{1-γ} }{(19\supk)^γ}, 
		\label{e:main_result_bis_choice_m}
	\end{align}
	we get
	\begin{align*}
		λ &= \frac{19\supk \log\frac{32n}{δ}}{ n }
	\end{align*}
	and the sufficient conditions \eqref{e:hyps_for_probabilisic_bound_lambda_bis_als} are satisfied as long as $n$ is large enough 
	and $λ_0$ is small enough, \ie,
	\begin{subequations}
	\begin{align}[left = \empheqlbrace\,]
		λ₀ ≤ \frac{19\supk \prt*{ \log\frac{32n}{δ}}}{n} &≤ \noprkhs{\covop}  \\
		n^γ(\log\frac{32n}{δ})^{-γ} &≥ \frac{334 (19\supk)^γ}{78 c_γ z²  } 
	\end{align}
	\end{subequations}

	In this regime, the error \eqref{e:main_result_bis_error_bound} follows the rate 
		$O\left(\frac{\log(n)^{1/2}}{ n^{1/2} }\right)$.
	From \eqref{e:main_result_bis_choice_m}, one can observe that $\log(n)≤\log(n\log(32 n/δ)^{(1-γ)/γ})=\text{cst}+\log(m^{1/γ})$
	so that the error \eqref{e:main_result_bis_error_bound} also follows a quantization rate of order 
		$O(\sqrt{λ})=O\left(\frac{\log(m)^{1/(2γ)}}{m^{1/(2γ)}}\right)$ with respect to $m$.

	\paragraph{Exponential decay}
	Under \Cref{a:exp_decay}, by \Cref{r:deff_exp} it holds
	$\deff ≤ β^{-1}\log(1+\tfrac{a_β}{λ})$.  
	Given that $\log(1+x)≤\log(2x)$ whenever $x≥1$, the following conditions are sufficient to enforce \eqref{e:hyps_for_probabilisic_bound_lambda_bis_als}:
	\begin{align}\label{e:main_result_bis_hyps_expdecay}
		\left\{ \begin{array}{l}
			λ₀ ≤ λ ≤ \min(a_β, \noprkhs{\covop})  \\
			λ ≥ \frac{19\supk}{n}\log\prt*{\frac{8n}{δ}} \\
			m ≥ 334\log\prt*{\frac{32 n}{δ}} \\
			m ≥ 78 z² β^{-1}\log(2\tfrac{\new{a_β}}{λ}) \log\frac{32 n}{δ} 
		\end{array} \right.
	\end{align}
	One can easily check that the choice 
	\begin{align*}
		λ&\de\frac{19\supk}{n}\log\prt*{\frac{8n}{δ}}, 
		\quad 
		m \de \max(334, 78 z² β^{-1}\log\prt*{\tfrac{2a_β}{19\supk}n}) \log\prt*{\frac{32 n}{δ}} 
	\end{align*}
	satisfies \eqref{e:main_result_bis_hyps_expdecay} as long as 
	\begin{align}
		 & \max(a_β^{-1},\noprkhs{\covop}^{-1}) ≤ \tfrac{ n}{19\supk\log(\frac{8n}{δ})} ≤ λ₀^{-1}.
	\end{align}
	With these choices of parameters, we get a rate of order 
	$O(\sqrt{λ}) = O(\log(n)^{1/2} n^{-1/2})$.
	Moreover, if we assume for simplicity $m\de c_m \log(n)²$, this yields the quantization rate $O(\sqrt{λ})=O(m^{1/4}\exp(-\sqrt{m}/c)$ with $c=2\sqrt{c_m}$.
\end{tproofof*}

\subsection{Proofs With Source Condition (\Cref{s:source_condition})}

\begin{tlemma}{Faster rate with source condition}{rate_source_condition}
	Let \Cref{a:bounded_fmap,a:iid} hold. 
	Let the sub-samples $\tilde{X}_1, \dots, \tilde{X}_m$ be drawn according to $(z,\lambda_0,\delta/4)$-approximate leverage scores from the dataset $\{X_1 \dots, X_n\}$, for some $z≥1$.
	The for any 
	$λ∊[\max(λ₀, \tfrac{19\supk \log(\frac{8n}{δ})}{n}) ; \noprkhs{\covop}]$, $δ \in (0,1)$, 
	$s∊[0,1/2]$, it holds with probability at least $1-δ$,
	\begin{align*}
		\wce[\fs]
		&≤ \frac{2\supfmap^{1+2s} \sqrt{2\log(6/δ)}}{\sqrt{n}} + (3λ)^{s+1/2} 
	\end{align*}
	provided that
	\begin{align*}
	    n &≥ (1655 + 233 \log(8\supk/δ))\supk\\
		m &≥ \max\prt*{334, 78 z² \deff}\log\prt*{\nicefrac{32n}{δ}}.
	\end{align*}
\end{tlemma}
\begin{tproofof*}{r:rate_source_condition}{}
	Let $g \in \rkhs$ such that $\lVert g \rVert \leq 1$ and let $f=C^s g$. Using the reproducing property of the RKHS $\rkhs$, the fact that the operator $C^s$ is self-adjoint and Cauchy-Schwarz inequality, we have
\begin{align*}
	\absv*{\int  f \dif \td - \sum_{j=1}^m f(\tilde{X}_j)}
	&= \absv*{\iprkhs*{C^s g, \int \phi \dif \td - \sum_{j=1}^m w_j \phi(\tilde{X}_j)}} ≤ \nrkhs*{C^s (\me - \nyseme)}\enspace.
\end{align*}

Using the same decomposition as in the proof of \Cref{r:error_decomposition} for \eqref{e:error_decomposition_bis}, we have 
\begin{align}
	\nrkhs*{\covop^s (\me - \nyseme)}
	&≤ \noprkhs*{\covop^s \Pm^\perp}\noprkhs*{\Pm^\perp \covopl^{1/2} }
	+ \supfmap^{2s}\, \nrkhs*{\me - \eme} 
\end{align}
\new{Since $\Pm^\perp$ and $\covop^{1/2}$ are positive bounded operators (respectively as a projection, and as the square root of a positive operator, cf. \Cref{a:notations}),} 
it holds by Cordes inequality (\Cref{r:cordes_inequality})
\begin{align*}
	\noprkhs*{\covop^s \Pm^\perp}
	&= \noprkhs*{(\Pm^\perp)^{2s} (\covop^{1/2})^{2s} } 
	 ≤ \noprkhs*{\Pm^\perp \covop^{1/2} }^{2s}.
\end{align*}
so that
\begin{align}
	\label{eq:source_decomp}
	\nrkhs*{C^s (\me - \nyseme)}
	&≤ \noprkhs*{\Pm^\perp \covopl^{1/2} }^{2s+1}
	+ \supfmap^{2s}\, \nrkhs*{\me - \eme}\enspace.
\end{align}

To control the second term, we apply Lemma~\ref{r:concentration_iid_hilbert} on the random variables $\eta_i \coloneqq \fmap(X_i)-\mu, 1 \leq i \leq n$. For any index $1\leq i \leq n$, it holds $\nrkhs{\eta_i}\leq 2\sup_{x\in\dspace}\nrkhs{\fmap(x)} = 2\supfmap$. Thus, with probability at least $1-\delta/2$ on the draw of the the dataset $X_1, \dots, X_n$,
	\begin{equation*}
	    \nrkhs{\mu  - \eme} \leq  \frac{2\supfmap \sqrt{2\log(4/\delta)}}{\sqrt{n}}\enspace.
	\end{equation*}



	Next,  we rely on \Cref{r:als_nys_approx} to bound the term $\noprkhs{\Pmo \covopl^{1/2}}$ with high probability. Under the hypothesis of the Lemma, we have, for any $λ∊]0,\noprkhs{\covop}]$, with probability at least $1-\delta/2$ on the draw of the landmarks $\tilde{X}_1, …, \tilde{X}_m$,
	\[ \noprkhs{\Pmo \covopl^{1/2}} ≤ \sqrt{3λ}. \]
	
	The proof is concluded by taking the union bound over the two high-probability events on which we control the first and the second term of \eqref{eq:source_decomp}.
\end{tproofof*}

We now prove the quantization rates claimed with a source condition. 

\begin{tproofof*}{r:source_condition_final_rate}{}
By \Cref{r:rate_source_condition}, we have
\begin{align*}
	\nrkhs{\me  - \nyseme} 
	&≤ \frac{2\supfmap^{1+2s} \sqrt{2\log(4/δ)}}{\sqrt{n}} + 
	(3λ)^{s+1/2}.
\end{align*}
We now need to pick $λ,m$ ensuring
	\begin{subequations}
	\begin{align}[left = \empheqlbrace\,]
		m &≥ \max\prt*{334, 78 z² \deff}\log\prt*{\frac{32n}{δ}} \label{e:sc_double_condition_m}\\
		λ &≥\tfrac{19\supk \log(\frac{8n}{δ})}{n} \label{e:sc_rate_condition_lambda}\\
		 λ₀ &≤ λ ≤ \noprkhs{\covop}. \label{e:sc_poly_bounds_lambda}
	\end{align}
	\end{subequations}

We pick $\boxed{λ= \prt*{\frac{19 \supk \log(\nicefrac{32 n}{δ})}{n}}^{1/(2s+1)}}$,
which is the largest choice for $λ$ allowing to get (up to log term) a global rate of order $Θ(n^{-1/2})$ while satisfying \eqref{e:sc_rate_condition_lambda} (by assumption it holds $\frac{19 \supk \log(\nicefrac{32 n}{δ})}{n}<1$).
Note that as soon as $s>0$, the logarithmic term in $n$ can be avoided provided $n$ is large enough and one recovers exactly the optimal rate $O(n^{-1/2})$. 
We opt here for a unified analysis with simplified constraints at the cost of achieving only the rate $O(\log(n)n^{-1/2})$. 

Condition \eqref{e:sc_poly_bounds_lambda} holds as soon as
\begin{align*}
	& λ₀^{2s+1} ≤ \frac{19 \supk \log(\nicefrac{32 n}{δ})}{n} ≤ \noprkhs{\covop}^{2s+1}. 
\end{align*}

We now consider the settings of polynomial and exponential decay of the spectrum, 
and detail how to choose $m$ in order to satisfy the remaining constraints \eqref{e:sc_double_condition_m}, which we rewrite as: 

\begin{subequations}\label{e:sc_remaining_constraints}
\begin{empheq}[left=\empheqlbrace]{align}
	& m ≥ 334\log\frac{32n}{δ} \label{e:sc_poly_cond_lambda_cst} \\
	& m ≥ 78 z² \deff \log\frac{32n}{δ} \label{e:sc_poly_cond_m}
\end{empheq}
\end{subequations}

\paragraph{Polynomial decay}
	Under \Cref{a:poly_decay}, by \Cref{r:deff_poly}
	it holds $\deff ≤ c_γ λ^{-γ}$. We choose
	\begin{align*}
		\Aboxed{
		m &\de
			c_m \log\prt*{\nicefrac{32n}{δ}}^{1-γ/(2s+1)} n^{γ/(2s+1)} 
			\quad\text{where}\quad
			c_m \de 78 z² c_γ (19\supk)^{-γ/(2s+1)}
		}
	\end{align*}
	which satisfies \eqref{e:sc_poly_cond_m}.
	Condition \eqref{e:sc_poly_cond_lambda_cst} is satisfied whenever
	\begin{align*}
		n &≥ \prt*{\tfrac{334}{c_m}}^{(2s+1)/γ}\log\prt*{\nicefrac{32n}{δ}}.
	\end{align*}

	The quantization rate can be derived by noting that 
	\begin{align*}
		m^{-(2s+1)/(2γ)}
		&= Θ\prt*{\prt*{\log\prt*{\nicefrac{32n}{δ}}^{1-γ/(2s+1)} n^{γ/(2s+1)}}^{-(2s+1)/(2γ)}}\\
		&= Θ\prt*{n^{-1/2} \log\prt*{\nicefrac{32n}{δ}}^{1/2-\frac{2s+1}{2γ}}}
	\end{align*}
	with $\frac{2s+1}{2γ}≥1/2$.
	Thus get the rate
	\begin{align*}
		\wce[\fs] 
		&= Θ(λ^{(2s+1)/2}) 
		= Θ\prt*{\frac{\log(n)^{1/2}}{ n^{1/2}}}
		= O(m^{-(2s+1)/(2γ)}).
	\end{align*}

\paragraph{Exponential decay}
	Under \Cref{a:exp_decay} holds, by \Cref{r:deff_exp} it holds $\deff ≤ β^{-1}\log(1+\tfrac{a_β}{λ})$.
	Using that $\log(2x)≥\log(1+x)$ whenever $x≥1$, 
	the constraint
	\eqref{e:sc_remaining_constraints}
	is satisfied as soon as
	\begin{subequations}
	\begin{align}[left = \empheqlbrace\,]
		m &≥ \frac{78 z²}{β(2s+1)} \log\prt*{ (2a_β)^{2s+1} \frac{n}{19\supk\log(\nicefrac{32 n}{δ})} } \log\prt*{\nicefrac{32n}{δ}} \label{e:sc_exp_first_cond_m}\\
		n &≥ {19\supk a_β^{-(2s+1)} \log(\nicefrac{32 n}{δ})}\label{e:sc_exp_cond_n}\\
		m &≥ 334\log\prt*{\nicefrac{32n}{δ}}\label{e:sc_exp_second_cond_m}
	\end{align}
	\end{subequations}

	We choose 
	\begin{align*}
		\Aboxed{m &\de \max\prt*{\frac{c_m}{2s+1}\log\prt*{ c_m' n }, 334} \log\prt*{ c_m' n }}
		\\\quad\text{where}\quad
		c_m &\de 78 z²β^{-1},
		c_m' \de \max\prt*{\tfrac{(2a_β)^{2s+1}}{19\supk}, \tfrac{32}{δ}}
	\end{align*}
	in order to enforce both \eqref{e:sc_exp_first_cond_m} and \eqref{e:sc_exp_second_cond_m}, while \eqref{e:sc_exp_cond_n} is satisfied by assumption.
	Note that with this definition, there exists $N∊\bN$ such that for any $n≥N$, it holds
	\begin{align*}
		m &= \frac{ c_m }{ 2s+1 } \log\prt*{ c_m' n }²
	\end{align*}
	so that asymptotically $n=\exp(\sqrt{(2s+1)m/c_m})/c_m'$, 
	and the quantization rate can be expressed as
	\begin{align*}
		\wce[\fs] 
		&= Θ(λ^{(2s+1)/2}) 
		= Θ\prt*{\frac{\log(n)^{1/2}}{ n^{1/2}}}
		= O\prt*{  m^{1/4}\exp\prt*{ -\tfrac{\sqrt{2s+1}}{2\sqrt{c_m}} \sqrt{m} } }.
	\end{align*}
\end{tproofof*}

\section{Auxiliary Results}
\label{s:aux_results}

\subsection{Bounds on the Effective Dimension}
\label{s:bounds_deff}

We now recall how the effective dimension can be bounded under any of \Cref{a:poly_decay} or \Cref{a:exp_decay}. 

\begin{tlemma}{Effective dimension, polynomial decay}{deff_poly}
	Under \Cref{a:bounded_fmap,a:poly_decay} it holds
	\begin{align}
		\deff &\leq  c_γ \lambda^{-γ} 
		\text{ where }
		c_γ \de \lcba{\frac{a_γ}{1-γ }, \text{ if }  γ <1 \\ \supfmap^2 , \text{ if } γ =1 }\enspace.
		\label{e:decay_deff}
	\end{align}
\end{tlemma}
It is well known, see \eg \textcite[Lemma 11]{fischer2020SobolevNormLearning}, that the existence of a constant $c_γ$ such that the first part of \eqref{e:decay_deff} holds implies in return a polynomial decay of the spectrum, \ie $σ_i\lesssim i^{-1/γ}$.
\begin{tproofof*}{r:deff_poly}{}
	The case $γ<1$ is covered in \pcite[Proposition 3 with $b\rightarrow 1/γ$ and $γ \rightarrow c$]{caponnetto2007OptimalRatesRegularized}.
	The case $γ=1$ follows from the observation $\deff ≤ \dsup = \esssup_{x\sim \td} \n{\covopl^{-1/2}\fmap{x}}²≤\n{\covopl^{-1/2}}²\esssup_{x\sim \td} \nrkhs{\fmap(x)}²≤\supfmap²/λ$.
\end{tproofof*}

For the exponential decay setting (\Cref{a:exp_decay}), we use the following result of \textcite[Proposition 5]{dellavecchia2021RegularizedERMRandom}.
\begin{tlemma}{Effective dimension, exponential decay}{deff_exp}
	Under \Cref{a:exp_decay} it holds
	\begin{align}
		\deff &\leq \log(1+a_β/λ)/β
		\label{e:deff_exp_decay}
	\end{align}
\end{tlemma}

\subsection{Nyström Approximation Result}
\label{a:nystrom}

To control the term involving $\Pmo$, we rely on the following lemma from Rudi et. al~\pcite[Lemma 6]{rudi2015LessMoreNystrom}.
\begin{tlemma}{Uniform Nyström approximation}{plain_nys_approx}
	When the set of $m$ landmarks is drawn uniformly from all partitions of cardinality $m$, for any $λ∊]0,\noprkhs{\covop}]$ we have 
	\[ \noprkhs{\Pmo (\covop+λI)^{1/2}}^2 \leq 3λ\]
	with probability at least $1-δ$ provided 
	\[ m\geq\max(67, 5\dsup)\log\frac{4\supk}{λδ}. \] 
\end{tlemma}
The next lemma is a restatement of \pcite[Lemma 7]{rudi2015LessMoreNystrom}.
\begin{tlemma}{ALS Nyström approximation}{als_nys_approx}
Let $z≥1$, $λ₀>0$ and $δ∊]0,1[$.
Let $(\hat{\ell}_t(i))_{1\leq i\leq n}$ be a collection of $(z,λ_0,δ/2)$-approximate leverage scores. 
Let $λ<\noprkhs{\covop}$, and $p_λ$ be a probability distribution on the set of indexes $\cb*{1,\ldots,n}$ defined as $p_λ(i)\de \new{\hat{\ell}_λ(i)}/(\sum_{i=1}^n \new{\hat{\ell}_λ(i)})$. Let $\cb*{i_1,\ldots,i_m}$ be a collection of indices sampled independently with replacement from $p_λ$, and $\Pm$ the orthogonal projection on $\rkhsm=\spa\cb*{\fmap(\Vx_{i_1}),\ldots,\fmap(\Vx_{i_m})}$. We have with probability at least $1-δ$ 
\begin{align*}
	\noprkhs{\Pmo (\covop+λI)^{1/2}} \leq  \sqrt{3λ}\enspace,
\end{align*}
	provided that
	\begin{itemize}
		\item $m ≥ \max\prt*{334, 78z^2\deff}\log\frac{16n}{δ}$;
		\item $n ≥ (1655 + 233 \log(4\supk/δ))\supk$;
		\item $19\supk\log(\frac{4n}{δ}) ≤ λn$;
		\item $λ₀ ≤ λ$.
	\end{itemize}
\end{tlemma}
\subsection{Misc. Results}

\begin{ttheorem}{{Cordes Inequality %
\new{\pcite[Lemma 5.1]{cordes1987SpectralTheoryLinear}}%
}}{cordes_inequality}
	Let $A,B$ be two positive bounded linear operators on a Hilbert space $\cH$. Then for any $s∊[0,1]$, it holds
	\[ \norm{A^sB^s} ≤ \norm{AB}^s \]
\end{ttheorem}

\section{Concentration Inequalities}
\label{a:concentration} 

This section contains concentration results that we rely on to prove our main result. 
These results are standard, and we include proofs for completeness.

The first lemma provides a high-probability control on the norm of the average of bounded random variables taking values in a separable Hilbert space.

\begin{tlemma}{}{concentration_iid_hilbert}
	Let $X_1, \dots, X_n$ be \tiid random variables on a separable Hilbert space $(\cX,\norm{\cdot})$ such that $\sup_{i=1, \dots, n}\norm{X_i}\leq A$ almost surely, for some real number $A > 0$. Then, for any $δ \in (0, 1)$, it holds with probability at least $1-δ$ that
	\[
	\norm*{\frac{1}{n}\sum_{i=1}^n X_i } 
	\leq A\frac{\sqrt{2\log(2/δ)}}{\sqrt{n}}.
	\]
\end{tlemma}
The proof of Lemma~\ref{r:concentration_iid_hilbert} relies on \pcite[Theorem 3.5]{pinelis1994optimum} which we recall now for clarity of exposition. 
\begin{tlemma}{}{pinelis_inequality}
	Let $M=(M_i)_{i\in \bN}$ be a martingale on a $(2,D)$-smooth separable Banach space $(\cX, \norm{\cdot})$. Define $\sum_{j=1}^\infty \esssup \norm{M_j - M_{j-1}}^2\leq b_*^2$, for some real number $b_*>0$. Then, for all $r\geq 0$,
	\[ \Pr\brk*{\sup_{j\in\bN} \norm{M_j} \geq r} \leq 2\exp\prt*{-\frac{r^2}{2D^2b_*^2}}. \]
\end{tlemma}
We now prove Lemma~\ref{r:concentration_iid_hilbert}.
\begin{tproofof*}{r:concentration_iid_hilbert}{}
	Since $\cX$ is a Hilbert space, it is $2$-smooth with $2$-smoothness constant $D=1$.
	We define the martingale $(M_n)_{n \in \mathbb{N}}$ as $M_0=0$, $M_k=\sum_{1\leq i\leq k} X_k$ for $1\leq k \leq n$ and $M_k = M_n$ for $k\geq n$, so that
	\begin{equation*}
    d_k \coloneqq M_k - M_{k-1} =
    \begin{cases}
      X_k, & \text{if}\ 1 \leq k \leq n \\
      0, & \text{otherwise}
    \end{cases}.
  \end{equation*}
	As a consequence, we have $\sum_{j=1}^\infty \esssup \norm{d_j}^2= \sum_{j=1}^n \esssup\norm{X_j}^2 \leq n A^2$. Applying Pinelis' inequality (Lemma~\ref{r:pinelis_inequality}) with $b_*^2=nA^2$ yields
	\begin{align*}
		\Pr\brk*{ \norm*{\frac{1}{n}\sum_{i=1}^n X_i } > \epsilon  } 
		&= \Pr\brk*{ \norm*{ M_n } > n\epsilon  } 
		\leq \Pr\brk*{ \sup_{1\leq j\leq n}\norm*{ M_j } > n\epsilon} \leq 2\exp\prt*{-\frac{n\epsilon^2}{2A^2}}.
	\end{align*}
	We get the desired result by choosing $\epsilon=A\sqrt{2\log(2/δ)}n^{-1/2}$.
\end{tproofof*}

The next result is a Bernstein-type inequality for random vectors defined in a Hilbert space.
\begin{tlemma}{Bernstein inequality for Hilbert space-valued random vectors}{bernstein_vectors}
	Let $X_1, \dots, X_n$ be \tiid random variables in a Hilbert space $(\rkhs, \norm{·})$ such that
	\begin{itemize}
		\item $∀i∊[n], \E X_i=μ$,
		\item $∃ \sigma>0,∃ H>0, ∀ i ∊ [n],∀p≥2,\, \E \norm{X_i-μ}^p ≤ ½p!σ²H^{p-2}$.
	\end{itemize}
	Then, for any $δ∊]0,1[$,  we have with probability at least $1-δ$,
	\begin{align*}
		\norm*{\frac{1}{n}\sum_{i=1}^{n} X_i - μ\, }
		≤ \frac{2H\log(2/δ)}{n} + \sqrt{\frac{2 σ² \log(2/δ)}{n}}. 
	\end{align*}
\end{tlemma}
\begin{tproofof*}{r:bernstein_vectors}{}
    Fix a confidence level $δ \in (0, 1)$.
	Applying \pcite[Theorem 3.3.4]{yurinsky1995SumsGaussianVectors}
	on the \tiid centered random variables $ξᵢ=Xᵢ-μ$ with $B²=σ²n$, we get 
	\begin{align*}
		\Pr\brk*{\norm*{\frac{1}{n}\sum_{j=1}^{n} ξⱼ - μ} ≥ t}
		&≤ \Pr\brk*{\max_{1≤k≤n} k\, \norm*{\frac{1}{k}\sum_{j=1}^{k} ξⱼ - μ} ≥ \prt*{\frac{tn}{B}} B} \\
		&≤ 2\exp\prt*{-½\frac{(tn)²}{B²}\prt*{1+\frac{tHn}{B²}}^{-1}}.
	\end{align*}
	The RHS of the above is smaller than $δ$ if and only if
	\begin{align*}
		& t²n²  - t(2Hn\log(2/δ)) - 2B²\log(2/δ) ≥ 0.
	\end{align*}
	Denoting $Δ = 4H²n²\log(2/δ)² + 8 n²B²\log(2/δ) > 0$, this holds in particular if $t ≥ \frac{H\log(2/δ)}{n} + \frac{\sqrt{Δ}}{2n²}$, and thus a fortiori (using $\sqrt{Δ}≤\sqrt{4H²n²\log(2/δ)²}+\sqrt{8 n²B²\log(2/δ)}$) when
	\begin{align*}
		t &≥ \frac{2H\log(2/δ)}{n} +  \sqrt{\frac{2 σ² \log(2/δ)}{n}}.
	\end{align*}
\end{tproofof*}

The following lemma provides a Bernstein-type bound for the empirical mean of Hilbert space-valued centered random variables 'whitened" by regularized linear operator.

\begin{tlemma}{}{bernstein_with_covop}
	Let $X₁,…,X_n$ be \tiid random variables taking values in a separable Hilbert space $(\cH,\ip{·,·})$ with associated norm $\norm{·}$. We denote their mean by $μ_X \coloneqq \E X_1$ and their covariance by $\covop \coloneqq \E\brk*{ X_1 \kron X_1}$.

    Let $Q:\cH→\cH$ be a linear operator. For any $λ > 0$ and $δ∊]0,1[$, it holds with probability at least $1-δ$ that
	\[ \norm*{Q_λ^{-1/2}\left(\frac{1}{n} \sum_{i=1}^n Xᵢ- μ_X\right)}
	≤ \frac{4\esssup \norm*{Q_λ^{-1/2}X_1}\log(2/δ)}{n} + \sqrt{\frac{4 \Tr(Q_λ^{-1}\covop) \log(2/δ)}{n}}. \]
\end{tlemma}

\begin{tproofof*}{r:bernstein_with_covop}{}
	To prove the stated result we will apply \Cref{r:bernstein_vectors} on the random variables $(ζᵢ)_{1≤i≤n}$ defined by $ζᵢ=Q_λ^{-1/2}Xᵢ$. 
	Let $N_Q(λ)=\Tr(Q_λ^{-1}\covop)$ and $N_{Q,∞}(λ)\coloneqq\esssup \norm*{Q_λ^{-1/2}X_1}$.
	
	For any index $1 \leq i \leq n$, we have $\Eζ_1 = Q_λ^{-1/2}μ_X$, 
	\begin{align*}
	    \esssup \nrkhs{ζᵢ-\E[ζᵢ]}
		&≤ 2\esssup \nrkhs{ζᵢ} 
		= 2 N_∞(λ)^{1/2},
	\end{align*}	
	and,
	\begin{align*}
		\E\nrkhs{ζᵢ-\E[ζᵢ]}²
		&= \Tr(\E\ip{ζᵢ-\E[ζᵢ], ζᵢ-\E[ζᵢ]}) \\
		&= \Tr(\E\brk*{(ζᵢ-\E[ζᵢ])\kron(ζᵢ-\E[ζᵢ])}) \\
		&= \Tr(\E[ζᵢ\kron ζᵢ] - \Eζᵢ \kron \Eζᵢ) \\
		&≤ \Tr(\E[ζᵢ\kron ζᵢ]) \\
		&= \Tr(Q_λ^{-1/2}\covop Q_λ^{-1/2})\\
		&= N_Q(λ).
	\end{align*}
	Moreover, for any $p≥2$, 
	\begin{align*}
		\nrkhs{ζᵢ-\E[ζᵢ]}^p
		&≤ 
		(\E\nrkhs{ζᵢ-\E[ζᵢ]}²) (\esssup \nrkhs{ζᵢ-\E[ζᵢ]}^{p-2}) \\
		&≤ ½ (2N_Q(λ)) (2N_{Q,∞}(λ)^{1/2})^{p-2}\\
		&≤ ½ p! (2N_Q(λ)) (2N_{Q,∞}(λ)^{1/2})^{p-2}.
	\end{align*}
	The result follows from Lemma~\ref{r:bernstein_vectors}
	with constants $σ²=2N_Q(λ)$ and $H=2N_{Q,∞}(λ)^{1/2}$.
\end{tproofof*}

Lemma~\ref{r:bound_mean_embeddings_with_rgcovmhalf} is a specialization of \Cref{r:bernstein_with_covop} to bound the last term appearing in Lemma~\ref{r:error_decomposition} in our setting of Nyström uniform sampling.

\begin{tlemma}{}{bound_mean_embeddings_with_rgcovmhalf}
	Assume that the $m\geq1$ Nyström landmarks are sampled uniformly with replacement from the dataset $X_1, \dots, X_n$.  If $0<λ≤\noprkhs{\covop}$ and $λn≥12\supfmap²\log(4/δ)$, it holds with probability at least $1-δ$,
	\[ \norm*{C_λ^{-1/2}(\eme - \emem)} ≤ \frac{4\sqrt{\dsup}\log(4/δ)}{m} + \sqrt{\frac{12\deff \log(4/δ)}{m}}. \]
\end{tlemma}
\begin{tproofof*}{r:bound_mean_embeddings_with_rgcovmhalf}{}
	Fix the desired confidence level $δ \in (0, 1)$.
	Let us begin by conditioning w.r.t. to the dataset $X_1, \dots, X_n$. As the landmarks are assumed to be drawn i.i.d., we can apply \Cref{r:bernstein_with_covop} with $Q=\covop$ on the \tiid random variables $h_j \coloneqq\fmap(\tilde{X}_j), 1 \leq j \leq m$, satisfying $\E[h_1]=\eme$, 
	$\E[h_1\kron h_1] = \ecovop$ and $\esssup \nrkhs{\covopl^{-1/2}h_1}² ≤ \dsup$:
	it holds with probability at least $1-\nicefrac{δ}{2}$ (over the drawing of the landmarks) that
	\[ \norm*{Q_λ^{-1/2}(μ_X - \hat{μ}_X)}
	≤ \frac{4\sqrt{\dsup}\log(4/δ)}{m} + \sqrt{\frac{4\Tr(\covopl^{-1}\ecovop) \log(4/δ)}{m}}. \]
	
	Then, since we assumed $λ≤\noprkhs{\covop}$ and $λn≥12\supfmap²\log(4/δ)$, \Cref{r:bound_Nhalf} ensures that $\Tr(\covopl^{-1}\ecovop)≤3\deff$ with probability at least $1-\nicefrac{δ}{2}$ w.r.t. the dataset $X_1, \dots, X_n$.
	
    Finally, since the drawing of dataset and that of the indexes of the landmark are independent, the claimed bound holds with probability at least $(1-\nicefrac{δ}{2})(1-\nicefrac{δ}{2})\geq 1 - δ$.
\end{tproofof*}

The next lemma bounds the trace term involving the empirical covariance appearing in \Cref{r:bound_mean_embeddings_with_rgcovmhalf} by the effective dimension. 

\begin{tlemma}{}{bound_Nhalf}
	Let $δ>0$, $λ>0$ and $n \in \mathbb{N}$ be such that $λ≤\noprkhs{\covop}$ and $n\geq 12\dsup\log(2/δ)$. Then it holds with probability at least $1-δ$ that
	\begin{align*}
		\Tr(\covopl^{-1}\ecovop)
		&≤ 3\deff. 
	\end{align*}
\end{tlemma}
\begin{tproofof*}{r:bound_Nhalf}{}
Let us control the deviation of $\Tr(\covopl^{-1}\ecovop)$ from its expectation $\deff$. We have
	\begin{align*}
		\Tr(\covopl^{-1}\ecovop) - \deff = \Tr(\covopl^{-1}(\ecovop - \covop)) = \frac{1}{n} \sum_{i=1}^n \xi_i - \E[\xi_i],
	\end{align*}
	where we define $\xi_i \coloneqq \Tr(\covopl^{-1}\fmap(X_i)\kron\fmap(X_i)), i=1, \dots, n$. The random variables $\xi_i, 1\leq i \leq n,$ satisfy
	\begin{align*}
		|\xi_i-\E[\xi_i]| &= \absv*{\Tr(\covopl^{-1}(\fmap(X_i)\kron\fmap(X_i) - \covop))}  ≤ \nrkhs*{\covopl^{-1/2}\fmap(X_i)}² + \deff 
		≤ 2\dsup
	\end{align*}
	and
	\begin{align*}
	    \E[(\xi_i-\E[\xi_i])²] 
		&= \E[\xi_i²]-(\E \xi_i)²  
		≤ \esssup |\xi_i|\,  \E[\xi_i]
		≤ 2\dsup\deff.
	\end{align*}
	\Cref{r:bernstein_vectors} with $H=2\dsup$ and $σ²=2\dsup\deff$ ensures that with probability at least $1-δ$,
	\begin{align*}
		|\Tr(\covopl^{-1}\ecovop)-\deff| 
		≤ \frac{4\dsup\log(2/δ)}{n} + \sqrt{\frac{4\dsup\deff \log(2/δ)}{n}}. 
	\end{align*}
	Since $λ≤\noprkhs{\covop}$, we have $\deff=\Tr(\covop\covopl^{-1})≥\noprkhs*{\covop\covopl^{-1}}=\frac{\noprkhs{\covop}}{\noprkhs{\covop}+λ}≥1/2$. 
	Furthermore, using the assumption $n≥12\dsup\log(2/δ)$, it holds with probability at least $1-δ$, 
	\begin{align*}
		\Tr(\covopl^{-1}\ecovop)
		≤ \deff\prt*{1 + \frac{1}{3\deff} + \sqrt{\frac{1}{3\deff}}}
		≤ \deff\prt*{1 + \frac{2}{3} + \sqrt{\frac{2}{3}}} \leq 2.5 \mathcal{N}(λ).
	\end{align*}
\end{tproofof*}

\section{Experiments}
\label{s:experiments_appendix}






\subsection{Implementation Details}
\label{s:impl_details}

Experiments in \Cref{s:psobolev} have been run on a Intel(R) Core(TM) i7-7700HQ CPU @ 2.80GHz (4 cores, 8 threads) with 4 BLAS threads.
Experiments in \Cref{s:real_data} have been run on a AMD EPYC 7301 16-Core Processor @ 2.20GHz (32 cores, 64 threads) with 32 BLAS threads.
We did not use GPUs to make it easier to fairly compare the different methods and measure runtimes. Note however that some methods, such as the BLESS algorithm that we use to compute approximate leverage scores, have a GPU implementation and could be accelerated in this way.

The datasets can be freely downloaded from \url{https://www.openml.org/}, however to ensure reproducibility we provide the \texttt{CuratedDataset}\footnote{\url{https://gitlab.com/dzla/CuratedDatasets.jl}} Julia package which takes care of downloading, preprocessing and loading the data.
All datasets have been centered and reduced.

The method of \cite{belhadji2019KernelQuadratureDPPs} is reported in \Cref{s:psobolev} only in dimension $d=1$ because the code for the setting $d>1$ is not publicly available.

\subsection{Implementation of the Greedy Methods}
\label{s:greedy}

In \Cref{s:experiments} we considered three kernel-based greedy methods to compute quadratures rules. We provide here a few details on how such methods can be implemented. 
Note that we do not solve the (usually non-convex) optimization problem to select the new atom on $\dspace$, but rather do an approximate exhaustive search over the data samples.
For generality, we denote $f∊\rkhs$ the function to approximate, although in our context we always use these methods on $f=\eme$.
In the following, we denote $P_t$ the orthogonal projection on the space $\spa\cb{\fmap{\ldm{1}},…,\fmap{\ldm{t}}}$ spanned by the features of the so-far selected landmarks, $\mft[t]=[\fmap{\ldm{1}},…,\fmap{\ldm{t}}]$ the operator induced by their features and $\kmat[t]$ their kernel matrix.
The three considered methods are the following:
\begin{itemize}
	\item \textbf{Greedy minimization of the residual $P_t^\perp f$}, also known as the $f$-greedy method:
		\begin{align*}
			\ldm{t+1}
			&\de \argmin_{x∊X} |(P_t^\perp f)(x)|.
		\end{align*}
		Note that as we are optimizing over the dataset here (and not \eg over $\dspace$), this algorithm can be seen as \textbf{orthogonal matching pursuit} with the finite dictionary $\cb{\fmap{x₁},…,\fmap{x_n}}$, assuming the latter is normalized for the chosen kernel (which holds for instance for translation-invariant kernels).
	\item \textbf{Greedily maximization of $\det(\kmatm)$.}
		This method is also known as the \textbf{P-greedy method} in the kernel interpolation literature as it consists in maximizing the so-called power function:
		\begin{align*}
			\ldm{t+1}
			& \de \argmax_{x∊X}\,  \nrkhs*{P_t^\perp \fmap{x}}
		\end{align*}
		Note however that using the formula for the determinant of block matrices, 
		\begin{align*}
			\nrkhs*{P_t^\perp \fmap{x}}^2 
					&= \iprkhs{\fmap{x}, (I-\mft[t]\kmat[t]^{-1}\amft[t]) \fmap{x} } \\
					&= \kr(x, x) - \kr(x, \ldm_t)\kr(\ldm_t, \ldm_t)^{-1}\kr(\ldm_t, x) \numberthis\label{e:pfun_schur}\\
					&= \frac{\det\prt*{ \kmat[t\cup \cb{x}] }}{\det(\kmat[t])} 
					\quad\quad\text{where}\quad\quad
					\kmat[t\cup \cb{x}] \de \brk*{\begin{array}{cc}
						\kmat[t] & \mft[t]^*\fmap{x} \\
						\fmap{x}^*\mft[t] & \kr(x,x) 
					\end{array}}
		\end{align*}
		so that this indeed corresponds to greedily maximizing the determinant of selected points.
		This method has also been proposed in \pcite{demarchi2005NearoptimalDataindependentPoint} and used in multiple works such as \pcite{chen2018FastGreedyMAP}.
		It is the only of the 3 mentioned methods that does not depend on the function $f$ to approximate.
	\item \textbf{Greedy minimization of $\n{\Pm^\perp f}$:}
		\begin{align*}
			\ldm{t+1}
			&∊ \argmin_{x∊X}\nrkhs*{P_{t,x}^\perp f}
		\text{ where $P_{t,x}$ is the othogonal projection on }
			\spa(\fmap{\ldm{1}},…,\fmap{\ldm{t}},\fmap{x}).
		\end{align*}
		This method is also known as \textbf{$f/P$ greedy interpolation}, as the new landmark chosen at each iteration is the one minimizing the residual over power function ratio. 
		A rewriting of $P_{t,x}$ indeed yields the following relation:
		\begin{align*}
			\nrkhs*{P_{t,x}^\perp f}²
			&= \nrkhs*{P_{t}^\perp f}² - \prt*{ \frac{(P_t^\perp f)(x)}{\nrkhs{ P_t^\perp \fmap{x}}} }².
		\end{align*}
\end{itemize}

\paragraph{Algorithm} 
These three methods can be implemented as shown in \Cref{al:greedy_algo},
and we implemented this algorithm in Julia\footnote{\url{https://gitlab.com/achatali/greedykernelmethods.jl}}.

\newcommand\mycommfont[1]{{\smaller #1}}
\begin{algorithm}
	\SetCommentSty{mycommfont}
	\KwIn{Kernel $κ$, number of landmarks $l$, function evaluations $f_{|X}∊\bR^n$, data $X∊\bR^{d×n}$}
	\KwOut{Quadrature points $X[:,S]$}
	$C ← \zeros(l, n)$ 
		\tcp*{Size $l×n$}
	$\sqpfunc ← [κ(X[:,i], X[:,i])\text{ for i in 1:n}]$ 
		\tcp*{$(\nrkhs{P_t^⟂\fmap(X_i)}²)_{1≤i≤n}$, $O(nc_κ)$ time}
	$r ← f_{|X}$ 
		\tcp*{Residual, size $n$. $O(n²)$ time when $f=\eme$.}
	$c_f ← \zeros(l)$ 
		\tcp*{Coefficients of $f$ in $U$, size $l$}
	$S ← []$
		\tcp*{Support (set of indexes in $\cb{1,…,n}$)}
	$k ← 0$\;
	\While{$k<l$}{
		$\text{newatom\_criterion} ← $ \uIf{\text{P-greedy}}{
			$\sqpfunc$ \;
		}\uElseIf{\text{f-greedy}}{
			$\resc$ \;
		}\Else{ 
			$r.^2/\sqpfunc$ \;
		}
		$j ← \argmax_{i∊\cb{1,…,n}\backslash S} \text{newatom\_criterion}$ \;
		$S ← S ∪ \{j\}$ \; 
		$k ← k+1$\; 
		$K_{j} ← \textup{kernelmatrix}(κ, x_j, X)$   \tcp*{Size $1×n$, $O(nc_κ)$ time}
		$\textup{idxs} ← \sqpfunc .>$ 1e-10 
			\tcp*{For stability, update only points which are not already in the subspace}
		$C[k,\textup{idxs}] ← (K_{j}[\textup{idxs}] - \textup{vec}(C[:,j]'*C[:,\textup{idxs}])) / \textup{sqrt}(\sqpfunc[j])$ 
\nllabel{al:C_update} 
			\tcp*{$O(nl)$ time}
		$c_f[k] ← \res[j]/\textup{sqrt}(\sqpfunc[j])$
		\nllabel{al:cf_update} 
			\tcp*{Update coefficients of $f$}
		$\resc ← \resc\,  .- c_f[k]*C[k,:]$
			\tcp*{Update the residual}
		$\sqpfunc ← \sqpfunc - (C[k,:]ᵀ).^2 $ 
			\nllabel{al:pfun_update} 
			\tcp*{Update power function, $O(n)$ time}
	}
	\Return $X[:,S]$
	\caption{Greedy algorithms ($f$-greedy,$P$-greedy,$f/P$-greedy) for kernel interpolation\label{al:greedy_algo}%
	}
\end{algorithm}

\paragraph{Computational cost} 
The algorithm has a cost of $O(nm(m+c_κ))$ time complexity, where $c_κ$ denotes the kernel evaluation time and is typically of order $c_κ=O(d)$. Note that this cost does not include the computation of weights.
Although we write the three algorithms together for conciseness, note that the method consisting in greedily maximizing $\det(K_m)$ does not require to compute the residual (the method being then independent of the function to approximate). 
In particular in our setting $f=\eme$ and this would avoid the $O(n²)$ cost of initializing the residual.
The cost for computing the weights is $O(nm+m³)$ and is the same for all methods (we use the same expression as for all other quadratures methods in the paper). With a small modification, the algorithm above can maintain an estimation of the weights, however the overall complexity of the algorithm remains unchanged. 

\paragraph{Implementation} 
We define the following quantities for any $1≤t≤m$, which match the notations in \Cref{al:greedy_algo} when relevant:
\begin{itemize}
	\item $\tilde{Φ}_t \de[\fmap{\ldm{1}},…,\fmap{\ldm{t}}]:\bR^t → \rkhs$. 
	\item $U_t=[u₁,…,u_t]: \bR^t→\rkhs$ is the Gram-Schmidt basis obtained from $\tilde{Φ}_t$, \ie for any $t$ it holds
	\begin{align}
		u_{t+1} &\de \frac{ P_t^\perp\fmap{\ldm{t+1}} }{ \n{P_t^\perp\fmap{\ldm{t+1}}} }
		\label{e:def_ut}
	\end{align}
	\item $C∊\bR^{m×n}$ whose columns contain at step $t$ the coefficients in $U_t$ of
		the projected data features $(P_t \fmap{X_i})_{1≤i≤n}$, \ie the block of the first $t$ columns of $C$ is $C_{1:t,:}=U_t^*[\fmap{x₁},…,\fmap{x_n}]$.
	\item $S$ is a set containing the indexes of the so-far selected landmarks.
\end{itemize}
The algorithm then derives from the following observations.
\begin{itemize}
	\item Line~\ref{al:C_update} derives from \eqref{e:def_ut}, indeed for any $i∊\cb{1,…,n}$:
	\begin{align*}
		\iprkhs{u_{t+1}, \fmap{X_i}}
		&= \frac{ \iprkhs{(I-P_t)\fmap{\ldm{t+1}}, \fmap{X_i}} }{ \n{P_t^\perp\fmap{\ldm{t+1}}} } \\
		&= \frac{κ(\ldm{t+1},X_i) - \iprkhs{P_t\fmap(\ldm{t+1}), P_t\fmap(X_i)} }{ \nrkhs{ P_t^\perp \fmap{\ldm{t+1}}} }
	\end{align*}
	and using the fact that at any iteration the index $j$ is updated such that $\ldm{t+1}=x_j$.
	\item Line~\ref{al:cf_update} follows from
	\begin{align*}
		\ip{f,u_{t+1}} 
			&= \frac{ (P_t^\perp\fmap{\ldm{t+1}})^*f }{ \nrkhs{P_t^\perp\fmap{\ldm{t+1}}} }
			 = \frac{ (P_t^\perp f)(\ldm{t+1}) }{ \nrkhs{P_t^\perp\fmap{\ldm{t+1}}} }
	\end{align*}
	Not in particular that no evaluations of $f$ are required for this operation. 
	\item Eventually Line~\ref{al:pfun_update} corresponds to the joint update for all $i∊\cb{1,…, n}$ of the power function:
	\begin{align*}
	\n{P_{t+1}^\perp \fmap{X_i}}²
		&= \n{(P_{t}^\perp - u_{t+1}u_{t+1}^*) \fmap{X_i}}²\\
		&= \n{P_{t}^\perp\fmap{X_i} - u_{t+1}u_{t+1}^* \fmap{X_i}}²\\
		&= \n{P_{t}^\perp\fmap{X_i}}² + \n{u_{t+1}u_{t+1}^* \fmap{X_i}}² - 2\iprkhs{(I-P_{t})\fmap{X_i}, u_{t+1}u_{t+1}^* \fmap{X_i} } \\
		&= \n{P_{t}^\perp\fmap{X_i}}² - \iprkhs{u_{t+1}, \fmap{X_i}}²
	\end{align*}
	where we used the fact that $P_t u_{t+1}=0$ and $\n{u_{t+1}}=1$.
\end{itemize}

\subsection{Additional Experimental Results}
\label{s:additional_exp_results}

\begin{figure*}
	\centerline{\includegraphics[width=\linewidth,keepaspectratio]{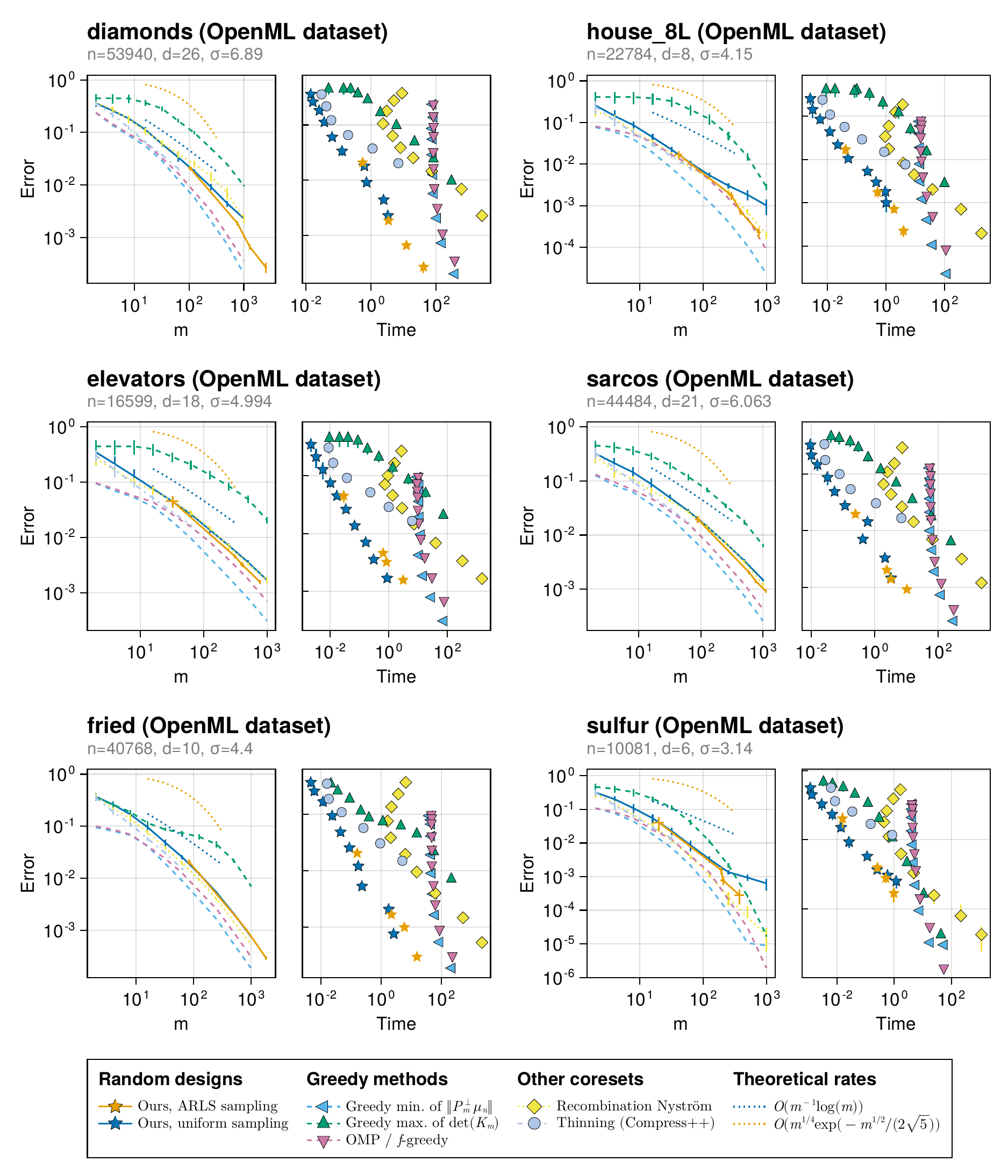}}
	\caption{\label{f:gkernel_appendix}
	Empirical results for OpenML datasets, Gaussian kernel.
	Each point is a median over 40 trials.
	}
\end{figure*}

\begin{figure*}
	\centerline{\includegraphics[width=\linewidth,keepaspectratio]{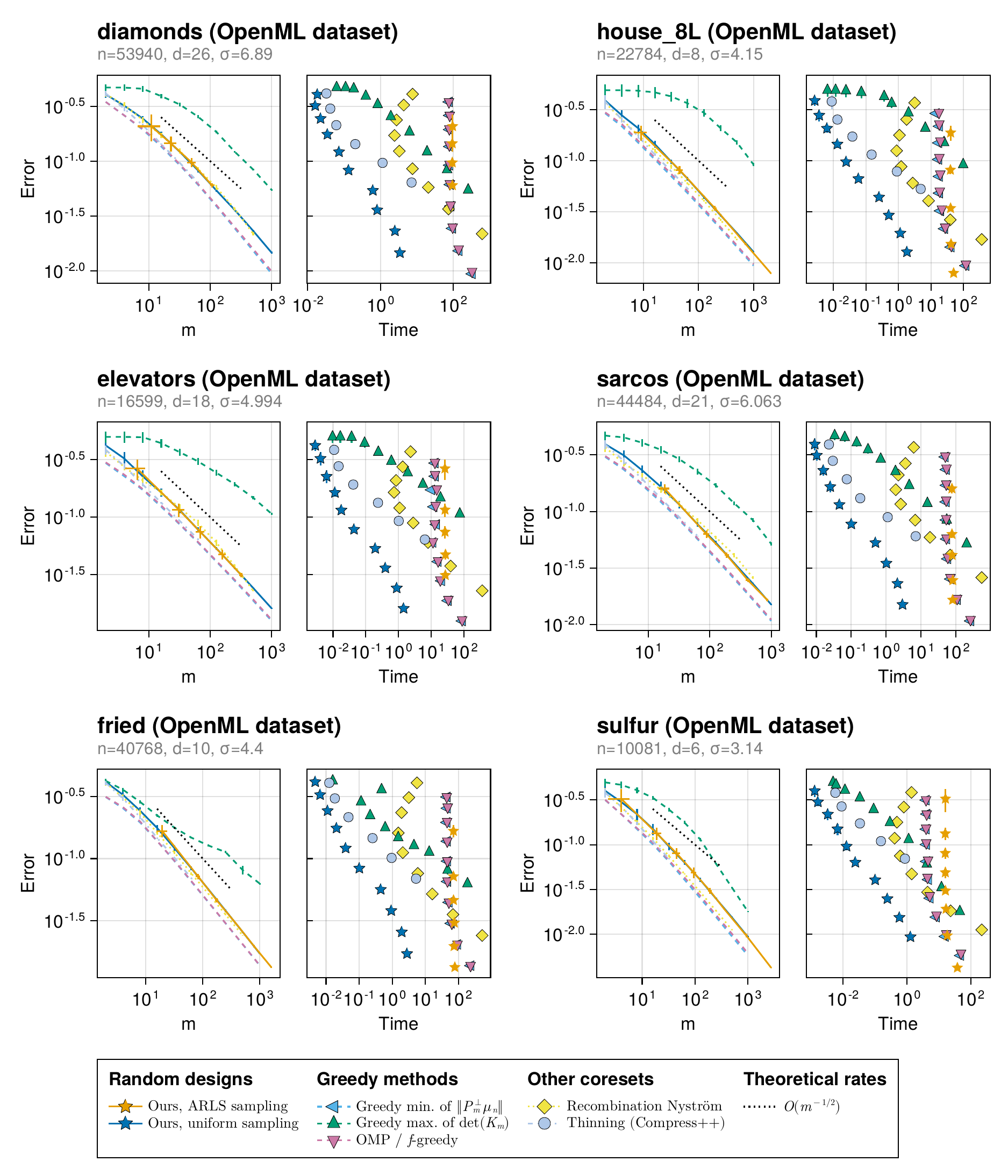}}
	\caption{\label{f:laplacian_appendix}
	Empirical results for OpenML datasets, Laplacian kernel.
	Each point is a median over 40 trials.
	}
\end{figure*}

We here provide empirical results for the setting of \Cref{s:real_data}, but on more datasets.
Results are reported in \Cref{f:gkernel_appendix} for the Gaussian kernel and \Cref{f:laplacian_appendix} for the Laplacian kernel.

\printbibliography 

@inproceedings{alaoui2015FastRandomizedKernel,
  title = {Fast Randomized Kernel Ridge Regression with Statistical Guarantees},
  booktitle = {Advances in {{Neural Information Processing Systems}}},
  author = {Alaoui, Ahmed and Mahoney, Michael W.},
  date = {2015},
  pages = {775--783},
  url = {https://papers.nips.cc/paper/2015/hash/f3f27a324736617f20abbf2ffd806f6d-Abstract.html}
}

@online{arbel2019MaximumMeanDiscrepancy,
  title = {Maximum {{Mean Discrepancy Gradient Flow}}},
  author = {Arbel, Michael and Korba, Anna and Salim, Adil and Gretton, Arthur},
  date = {2019-12-03},
  eprint = {1906.04370},
  eprinttype = {arXiv},
  eprintclass = {cs, stat},
  url = {http://arxiv.org/abs/1906.04370},
  urldate = {2022-06-23},
  langid = {english},
  pubstate = {prepublished}
}

@article{aronszajn1950TheoryReproducing,
  title = {Theory of Reproducing Kernels},
  author = {Aronszajn, Nachman},
  date = {1950},
  journaltitle = {Transactions of the American mathematical society},
  volume = {68},
  number = {3},
  pages = {337--404},
  publisher = {JSTOR}
}

@inproceedings{bach2012EquivalenceHerdingConditional,
  title = {On the Equivalence between Herding and Conditional Gradient Algorithms},
  booktitle = {Proceedings of the 29th {{International Coference}} on {{International Conference}} on {{Machine Learning}}},
  author = {Bach, Francis and Lacoste-Julien, Simon and Obozinski, Guillaume},
  date = {2012-06-26},
  series = {{{ICML}}'12},
  pages = {1355--1362},
  publisher = {Omnipress},
  location = {Madison, WI, USA},
  isbn = {978-1-4503-1285-1}
}

@article{bach2017EquivalenceKernelQuadrature,
  title = {On the Equivalence between Kernel Quadrature Rules and Random Feature Expansions},
  author = {Bach, Francis},
  date = {2017},
  journaltitle = {The Journal of Machine Learning Research},
  volume = {18},
  number = {1},
  pages = {714--751},
  publisher = {JMLR. org}
}

@inproceedings{belhadji2019KernelQuadratureDPPs,
  title = {Kernel Quadrature with {{DPPs}}},
  author = {Belhadji, Ayoub and Bardenet, Rémi and Chainais, Pierre},
  date = {2019-12-31},
  volume = {32},
  eprint = {1906.07832},
  eprinttype = {arXiv},
  eprintclass = {cs, stat},
  pages = {12927--12937},
  url = {http://arxiv.org/abs/1906.07832},
  urldate = {2021-03-16},
  eventtitle = {Advances in {{Neural Information Processing Systems}}},
  langid = {english}
}

@inproceedings{belhadji2020KernelInterpolationContinuous,
  title = {Kernel Interpolation with Continuous Volume Sampling},
  booktitle = {International {{Conference}} on {{Machine Learning}}},
  author = {Belhadji, Ayoub and Bardenet, Rémi and Chainais, Pierre},
  date = {2020-11-21},
  pages = {725--735},
  publisher = {PMLR},
  issn = {2640-3498},
  url = {http://proceedings.mlr.press/v119/belhadji20a.html},
  urldate = {2021-05-14},
  eventtitle = {International {{Conference}} on {{Machine Learning}}},
  langid = {english}
}

@inproceedings{belhadji2021AnalysisErmakovZolotukhinQuadrature,
  title = {An Analysis of {{Ermakov-Zolotukhin}} Quadrature Using Kernels},
  booktitle = {Advances in {{Neural Information Processing Systems}}},
  author = {Belhadji, Ayoub},
  date = {2021},
  volume = {34},
  pages = {27278--27289},
  publisher = {Curran Associates, Inc.},
  url = {https://proceedings.neurips.cc/paper/2021/hash/e531e258fe3098c3bdd707c30a687d73-Abstract.html},
  urldate = {2023-02-15}
}

@article{bos2010ComputingMultivariateFekete,
  title = {Computing {{Multivariate Fekete}} and {{Leja Points}} by {{Numerical Linear Algebra}}},
  author = {Bos, L. and De Marchi, S. and Sommariva, A. and Vianello, M.},
  date = {2010-01},
  journaltitle = {SIAM Journal on Numerical Analysis},
  shortjournal = {SIAM J. Numer. Anal.},
  volume = {48},
  number = {5},
  pages = {1984--1999},
  issn = {0036-1429, 1095-7170},
  doi = {10.1137/090779024},
  url = {http://epubs.siam.org/doi/10.1137/090779024},
  urldate = {2023-07-14},
  langid = {english}
}

@inproceedings{briol2015FrankWolfeBayesianQuadrature,
  title = {Frank-{{Wolfe Bayesian}} Quadrature: {{Probabilistic}} Integration with Theoretical Guarantees},
  shorttitle = {Frank-{{Wolfe Bayesian}} Quadrature},
  booktitle = {Advances in {{Neural Information Processing Systems}}},
  author = {Briol, François-Xavier and Oates, Chris and Girolami, Mark and Osborne, Michael A.},
  date = {2015},
  volume = {28}
}

@inproceedings{briol2017SamplingProblemKernel,
  title = {On the {{Sampling Problem}} for {{Kernel Quadrature}}},
  author = {Briol, Francois-Xavier and Oates, Chris J and Cockayne, Jon and Chen, Wilson Ye and Girolami, Mark},
  date = {2017},
  pages = {10},
  eventtitle = {International {{Conference}} on {{Machine Learning}}},
  langid = {english}
}

@article{briol2019ProbabilisticIntegrationRole,
  title = {Probabilistic Integration: {{A}} Role in Statistical Computation?},
  shorttitle = {Probabilistic Integration},
  author = {Briol, F. X. and Oates, C. J. and Girolami, M. and Osborne, M. A. and Sejdinovic, D.},
  date = {2019-02},
  journaltitle = {Statistical Science},
  volume = {34},
  number = {1},
  pages = {1--22},
  issn = {0883-4237},
  url = {http://dx.doi.org/10.1214/18-STS660},
  urldate = {2022-02-01},
  issue = {1},
  langid = {english}
}

@article{caponnetto2007OptimalRatesRegularized,
  title = {Optimal Rates for the Regularized Least-Squares Algorithm},
  author = {Caponnetto, Andrea and De Vito, Ernesto},
  date = {2007},
  journaltitle = {Foundations of Computational Mathematics},
  volume = {7},
  number = {3},
  pages = {331--368},
  publisher = {Springer}
}

@inproceedings{carratino2021ParKSoundEfficient,
  title = {{{ParK}}: {{Sound}} and {{Efficient Kernel Ridge Regression}} by {{Feature Space Partitions}}},
  shorttitle = {{{ParK}}},
  booktitle = {Advances in {{Neural Information Processing Systems}} 34 Pre-Proceedings},
  author = {Carratino, Luigi and Vigogna, Stefano and Calandriello, Daniele and Rosasco, Lorenzo},
  date = {2021-06-23},
  eprint = {2106.12231},
  eprinttype = {arXiv},
  url = {https://proceedings.neurips.cc/paper/2021/hash/32b9e74c8f60958158eba8d1fa372971-Abstract.html},
  urldate = {2021-07-07},
  eventtitle = {{{NeurIPS}} 2021},
  langid = {english}
}

@article{chatalic2021CompressiveLearningPrivacy,
  title = {Compressive Learning with Privacy Guarantees},
  author = {Chatalic, Antoine and Schellekens, Vincent and Houssiau, Florimond and De Montjoye, Yves-Alexandre and Jacques, Laurent and Gribonval, Rémi},
  date = {2021-05-15},
  journaltitle = {Information and Inference: A Journal of the IMA},
  shortjournal = {Information and Inference: A Journal of the IMA},
  issn = {2049-8772},
  doi = {10.1093/imaiai/iaab005},
  url = {https://hal.inria.fr/hal-02496896v2/document},
  urldate = {2021-05-18},
  issue = {iaab005}
}

@inproceedings{chatalic2022MeanNystromEmbeddings,
  title = {Mean {{Nyström Embeddings}} for {{Adaptive Compressive Learning}}},
  booktitle = {Proceedings of {{The}} 25th {{International Conference}} on {{Artificial Intelligence}} and {{Statistics}}},
  author = {Chatalic, Antoine and Carratino, Luigi and Vito, Ernesto De and Rosasco, Lorenzo},
  date = {2022-05-03},
  pages = {9869--9889},
  publisher = {PMLR},
  issn = {2640-3498},
  url = {https://proceedings.mlr.press/v151/chatalic22a.html},
  urldate = {2022-12-24},
  eventtitle = {International {{Conference}} on {{Artificial Intelligence}} and {{Statistics}}},
  langid = {english}
}

@inproceedings{chatalic2022NystromKernelMean,
  title = {Nyström {{Kernel Mean Embeddings}}},
  booktitle = {Proceedings of the 39th {{International Conference}} on {{Machine Learning}}},
  author = {Chatalic, Antoine and Schreuder, Nicolas and Rosasco, Lorenzo and Rudi, Alessandro},
  date = {2022-07-17},
  series = {Proceedings of {{Machine Learning Research}}},
  volume = {162},
  pages = {3006--3024},
  publisher = {PMLR},
  url = {https://proceedings.mlr.press/v162/chatalic22a.html}
}

@inproceedings{chen2010SupersamplesKernelHerding,
  title = {Super-Samples from Kernel Herding},
  booktitle = {Proceedings of the {{Twenty-Sixth Conference}} on {{Uncertainty}} in {{Artificial Intelligence}}},
  author = {Chen, Yutian and Welling, Max and Smola, Alex},
  date = {2010-07-08},
  series = {{{UAI}}'10},
  pages = {109--116},
  publisher = {AUAI Press},
  location = {Arlington, Virginia, USA},
  isbn = {978-0-9749039-6-5}
}

@article{chen2018FastGreedyMAP,
  title = {Fast {{Greedy MAP Inference}} for {{Determinantal Point Process}} to {{Improve Recommendation Diversity}}},
  author = {Chen, Laming and Zhang, Guoxin and Zhou, Eric},
  date = {2018},
  journaltitle = {Advances in Neural Information Processing Systems},
  volume = {31},
  pages = {5627--5638},
  url = {https://proceedings.neurips.cc/paper/2018/hash/dbbf603ff0e99629dda5d75b6f75f966-Abstract.html},
  urldate = {2021-04-21},
  langid = {english}
}

@article{cherief-abdellatif2022FiniteSampleProperties,
  title = {Finite Sample Properties of Parametric {{MMD}} Estimation: Robustness to Misspecification and Dependence},
  shorttitle = {Finite Sample Properties of Parametric {{MMD}} Estimation},
  author = {Chérief-Abdellatif, Badr-Eddine and Alquier, Pierre},
  date = {2022-02-01},
  journaltitle = {Bernoulli},
  shortjournal = {Bernoulli},
  volume = {28},
  number = {1},
  eprint = {1912.05737},
  eprinttype = {arXiv},
  eprintclass = {cs, math, stat},
  issn = {1350-7265},
  doi = {10.3150/21-BEJ1338},
  url = {http://arxiv.org/abs/1912.05737},
  urldate = {2023-07-24},
  langid = {english}
}

@book{cordes1987SpectralTheoryLinear,
  title = {Spectral Theory of Linear Differential Operators and Comparison Algebras},
  author = {Cordes, Heinz Otto},
  date = {1987},
  volume = {76},
  publisher = {Cambridge University Press},
  url = {https://books.google.com/books?hl=fr&lr=&id=6WlLjznD2zwC&oi=fnd&pg=PA1&dq=Spectral+Theory+of+Linear+Differential+Operators+and+Comparison+Algebra&ots=VPTwP_RKyw&sig=5WpPdzr4xNr1tQMWy0UsMM8_fqk},
  urldate = {2025-01-15}
}

@unpublished{dellavecchia2021RegularizedERMRandom,
  title = {Regularized {{ERM}} on Random Subspaces},
  author = {Della Vecchia, Andrea and Mourtada, Jaouad and De Vito, Ernesto and Rosasco, Lorenzo},
  date = {2021-02-25},
  eprint = {2006.10016},
  eprinttype = {arXiv},
  eprintclass = {cs, stat},
  url = {http://arxiv.org/abs/2006.10016},
  urldate = {2021-04-08},
  langid = {english}
}

@article{delyon2016IntegralApproximationKernel,
  title = {Integral Approximation by Kernel Smoothing},
  author = {Delyon, Bernard and Portier, François},
  date = {2016},
  url = {https://projecteuclid.org/journals/bernoulli/volume-22/issue-4/Integral-approximation-by-kernel-smoothing/10.3150/15-BEJ725.short},
  urldate = {2025-02-11}
}

@article{demarchi2005NearoptimalDataindependentPoint,
  title = {Near-Optimal Data-Independent Point Locations for Radial Basis Function Interpolation},
  author = {De Marchi, Stefano and Schaback, Robert and Wendland, Holger},
  date = {2005},
  journaltitle = {Advances in Computational Mathematics},
  volume = {23},
  number = {3},
  pages = {317--330},
  publisher = {Springer}
}

@article{denoyelle2019SlidingFrankWolfe,
  title = {The Sliding {{Frank}}–{{Wolfe}} Algorithm and Its Application to Super-Resolution Microscopy},
  author = {Denoyelle, Quentin and Duval, Vincent and Peyré, Gabriel and Soubies, Emmanuel},
  date = {2019-12},
  journaltitle = {Inverse Problems},
  shortjournal = {Inverse Problems},
  volume = {36},
  number = {1},
  pages = {014001},
  publisher = {IOP Publishing},
  issn = {0266-5611},
  doi = {10.1088/1361-6420/ab2a29},
  url = {https://dx.doi.org/10.1088/1361-6420/ab2a29},
  urldate = {2024-04-25},
  langid = {english}
}

@article{devore1996RemarksGreedyAlgorithms,
  title = {Some Remarks on Greedy Algorithms},
  author = {DeVore, R. A. and Temlyakov, V. N.},
  date = {1996-12},
  journaltitle = {Advances in Computational Mathematics},
  shortjournal = {Adv Comput Math},
  volume = {5},
  number = {1},
  pages = {173--187},
  issn = {1019-7168, 1572-9044},
  doi = {10.1007/BF02124742},
  url = {http://link.springer.com/10.1007/BF02124742},
  urldate = {2023-07-20},
  langid = {english}
}

@book{dick2010DigitalNetsSequences,
  title = {Digital Nets and Sequences: {{Discrepancy}} Theory and Quasi-Monte Carlo Integration},
  author = {Dick, Josef and Pillichshammer, Friedrich},
  date = {2010},
  publisher = {Cambridge University Press},
  url = {libgen.li/file.php?md5=80129262ed4655a31063167acc6a1b18},
  isbn = {978-0-511-90197-3 0-511-90197-6 978-0-521-19159-3 0-521-19159-9}
}

@book{dick2022LatticeRulesNumerical,
  title = {Lattice {{Rules}}: {{Numerical Integration}}, {{Approximation}}, and {{Discrepancy}}},
  shorttitle = {Lattice {{Rules}}},
  author = {Dick, Josef and Kritzer, Peter and Pillichshammer, Friedrich},
  date = {2022},
  series = {Springer {{Series}} in {{Computational Mathematics}}},
  volume = {58},
  publisher = {Springer International Publishing},
  location = {Cham},
  doi = {10.1007/978-3-031-09951-9},
  url = {https://link.springer.com/10.1007/978-3-031-09951-9},
  urldate = {2025-02-12},
  isbn = {978-3-031-09950-2 978-3-031-09951-9},
  langid = {english}
}

@book{diestel1977VectorMeasures,
  title = {Vector Measures},
  author = {Diestel, Joseph and Uhl, John Jerry},
  date = {1977},
  series = {Mathematical {{Surveys}} and {{Monographs}} 15},
  publisher = {American Mathematical Soc.},
  url = {http://gen.lib.rus.ec/book/index.php?md5=a2a6336235af17cd060efd7c9596a018},
  isbn = {0-8218-1515-6 978-0-8218-1515-1}
}

@unpublished{dwivedi2021KernelThinning,
  title = {Kernel {{Thinning}}},
  author = {Dwivedi, Raaz and Mackey, Lester},
  date = {2021-11-12},
  eprint = {2105.05842},
  eprinttype = {arXiv},
  eprintclass = {cs, math, stat},
  url = {http://arxiv.org/abs/2105.05842},
  urldate = {2022-03-23},
  langid = {english}
}

@online{dwivedi2022GeneralizedKernelThinning,
  title = {Generalized {{Kernel Thinning}}},
  author = {Dwivedi, Raaz and Mackey, Lester},
  date = {2022-07-19},
  eprint = {2110.01593},
  eprinttype = {arXiv},
  eprintclass = {cs, math, stat},
  url = {http://arxiv.org/abs/2110.01593},
  urldate = {2022-10-20},
  langid = {english},
  pubstate = {prepublished}
}

@book{engl2000RegularizationInverseProblems,
  title = {Regularization of {{Inverse Problems}}},
  author = {Engl, Heinz Werner and Hanke, Martin and Neubauer, A.},
  date = {2000},
  series = {Mathematics and {{Its Applications}}},
  publisher = {Springer Netherlands},
  url = {https://www.springer.com/gp/book/9780792341574},
  urldate = {2021-10-14},
  isbn = {978-0-7923-4157-4},
  langid = {english}
}

@article{fanuel2022NystromLandmarkSampling,
  title = {Nyström Landmark Sampling and Regularized {{Christoffel}} Functions},
  author = {Fanuel, Michaël and Schreurs, Joachim and Suykens, Johan A. K.},
  date = {2022-06-01},
  journaltitle = {Machine Learning},
  shortjournal = {Mach Learn},
  volume = {111},
  number = {6},
  pages = {2213--2254},
  issn = {1573-0565},
  doi = {10.1007/s10994-022-06165-0},
  url = {https://doi.org/10.1007/s10994-022-06165-0},
  urldate = {2023-09-05},
  langid = {english}
}

@article{fischer2020SobolevNormLearning,
  title = {Sobolev Norm Learning Rates for Regularized Least-Squares Algorithms},
  author = {Fischer, Simon and Steinwart, Ingo},
  date = {2020},
  journaltitle = {The Journal of Machine Learning Research},
  volume = {21},
  number = {1},
  pages = {8464--8501},
  publisher = {JMLRORG}
}

@article{garud2017DesignComputerExperiments,
  title = {Design of Computer Experiments: {{A}} Review},
  shorttitle = {Design of Computer Experiments},
  author = {Garud, Sushant S. and Karimi, Iftekhar A. and Kraft, Markus},
  date = {2017-11},
  journaltitle = {Computers \& Chemical Engineering},
  shortjournal = {Computers \& Chemical Engineering},
  volume = {106},
  pages = {71--95},
  issn = {00981354},
  doi = {10.1016/j.compchemeng.2017.05.010},
  url = {https://linkinghub.elsevier.com/retrieve/pii/S0098135417302090},
  urldate = {2023-07-05},
  langid = {english}
}

@online{harchaoui2008TestingHomogeneityKernel,
  title = {Testing for {{Homogeneity}} with {{Kernel Fisher Discriminant Analysis}}},
  author = {Harchaoui, Zaid and Bach, Francis and Moulines, Eric},
  date = {2008-04-07},
  eprint = {0804.1026},
  eprinttype = {arXiv},
  eprintclass = {stat},
  url = {http://arxiv.org/abs/0804.1026},
  urldate = {2024-10-09},
  langid = {english},
  pubstate = {prepublished}
}

@inproceedings{hayakawa2022PositivelyWeightedKernel,
  title = {Positively {{Weighted Kernel Quadrature}} via {{Subsampling}}},
  booktitle = {Advances in {{Neural Information Processing Systems}}},
  author = {Hayakawa, Satoshi and Oberhauser, Harald and Lyons, Terry},
  date = {2022-10-11},
  volume = {35},
  eprint = {2107.09597},
  eprinttype = {arXiv},
  eprintclass = {cs, math, stat},
  pages = {6886--6900},
  url = {http://arxiv.org/abs/2107.09597},
  urldate = {2023-02-09},
  langid = {english}
}

@inproceedings{hayakawa2023SamplingbasedNystromApproximation,
  title = {Sampling-Based {{Nyström Approximation}} and {{Kernel Quadrature}}},
  booktitle = {Proceedings of the 40th {{International Conference}} on {{Machine Learning}}},
  author = {Hayakawa, Satoshi and Oberhauser, Harald and Lyons, Terry},
  date = {2023-07-03},
  pages = {12678--12699},
  publisher = {PMLR},
  issn = {2640-3498},
  url = {https://proceedings.mlr.press/v202/hayakawa23a.html},
  urldate = {2023-12-11},
  eventtitle = {International {{Conference}} on {{Machine Learning}}},
  langid = {english}
}

@inproceedings{huszar2012OptimallyweightedHerdingBayesian,
  title = {Optimally-Weighted Herding Is {{Bayesian}} Quadrature},
  booktitle = {Proceedings of the {{Twenty-Eighth Conference}} on {{Uncertainty}} in {{Artificial Intelligence}}},
  author = {Huszár, Ferenc and Duvenaud, David},
  date = {2012-08-14},
  series = {{{UAI}}'12},
  pages = {377--386},
  publisher = {AUAI Press},
  location = {Arlington, Virginia, USA},
  isbn = {978-0-9749039-8-9}
}

@inproceedings{jaggi2013RevisitingFrankWolfeProjectionfree,
  title = {Revisiting {{Frank-Wolfe}}: {{Projection-free}} Sparse Convex Optimization},
  shorttitle = {Revisiting {{Frank-Wolfe}}},
  booktitle = {International {{Conference}} on {{Machine Learning}}},
  author = {Jaggi, Martin},
  date = {2013},
  pages = {427--435},
  publisher = {PMLR}
}

@article{kanagawa2019ConvergenceGuaranteesAdaptive,
  title = {Convergence Guarantees for Adaptive {{Bayesian}} Quadrature Methods},
  author = {Kanagawa, Motonobu and Hennig, Philipp},
  date = {2019},
  journaltitle = {Advances in neural information processing systems},
  volume = {32}
}

@article{kanagawa2020ConvergenceAnalysisDeterministic,
  title = {Convergence {{Analysis}} of {{Deterministic Kernel-Based Quadrature Rules}} in {{Misspecified Settings}}},
  author = {Kanagawa, Motonobu and Sriperumbudur, Bharath K. and Fukumizu, Kenji},
  date = {2020-02-01},
  journaltitle = {Foundations of Computational Mathematics},
  shortjournal = {Found Comput Math},
  volume = {20},
  number = {1},
  pages = {155--194},
  issn = {1615-3383},
  doi = {10.1007/s10208-018-09407-7},
  url = {https://doi.org/10.1007/s10208-018-09407-7},
  urldate = {2021-09-14},
  langid = {english}
}

@inproceedings{karnin2019DiscrepancyCoresetsSketches,
  title = {Discrepancy, {{Coresets}}, and {{Sketches}} in {{Machine Learning}}},
  booktitle = {Proceedings of the {{Thirty-Second Conference}} on {{Learning Theory}}},
  author = {Karnin, Zohar and Liberty, Edo},
  date = {2019-06-25},
  pages = {1975--1993},
  publisher = {PMLR},
  issn = {2640-3498},
  url = {https://proceedings.mlr.press/v99/karnin19a.html},
  urldate = {2022-07-01},
  eventtitle = {Conference on {{Learning Theory}}},
  langid = {english}
}

@article{karvonen2021KernelbasedInterpolationApproximate,
  title = {Kernel-Based Interpolation at Approximate {{Fekete}} Points},
  author = {Karvonen, Toni and Särkkä, Simo and Tanaka, Ken’ichiro},
  date = {2021-05-01},
  journaltitle = {Numerical Algorithms},
  shortjournal = {Numer Algor},
  volume = {87},
  number = {1},
  pages = {445--468},
  issn = {1572-9265},
  doi = {10.1007/s11075-020-00973-y},
  url = {https://doi.org/10.1007/s11075-020-00973-y},
  urldate = {2023-07-20},
  langid = {english}
}

@inproceedings{keriven2017CompressiveKmeans,
  title = {Compressive {{K-means}}},
  author = {Keriven, Nicolas and Tremblay, Nicolas and Traonmilin, Yann and Gribonval, Rémi},
  date = {2017-03-05},
  url = {https://hal.inria.fr/hal-01386077/document},
  urldate = {2017-04-06},
  eventtitle = {International {{Conference}} on {{Acoustics}}, {{Speech}} and {{Signal Processing}} ({{ICASSP}})},
  langid = {english}
}

@inproceedings{khanna2021GeometricRatesConvergence,
  title = {Geometric Rates of Convergence for Kernel-Based Sampling Algorithms},
  booktitle = {Proceedings of the {{Thirty-Seventh Conference}} on {{Uncertainty}} in {{Artificial Intelligence}}},
  author = {Khanna, Rajiv and Hodgkinson, Liam and Mahoney, Michael W.},
  date = {2021-12-01},
  pages = {2156--2164},
  publisher = {PMLR},
  issn = {2640-3498},
  url = {https://proceedings.mlr.press/v161/khanna21a.html},
  urldate = {2023-07-24},
  eventtitle = {Uncertainty in {{Artificial Intelligence}}},
  langid = {english}
}

@book{kress2014LinearIntegralEquations,
  title = {Linear {{Integral Equations}}},
  author = {Kress, Rainer},
  date = {2014},
  series = {Applied {{Mathematical Sciences}} ({{Switzerland}}) 82},
  edition = {3},
  publisher = {Springer-Verlag New York},
  url = {libgen.li/file.php?md5=a97b3aa662b057fd7b6b86d0407d7bf9},
  isbn = {1-4614-9592-X 978-1-4614-9592-5 978-1-4614-9593-2 1-4614-9593-8}
}

@book{kreyszig1989IntroductoryFunctionalAnalysis,
  title = {Introductory Functional Analysis with Applications},
  author = {Kreyszig, Erwin},
  date = {1989},
  series = {Wiley Classics Library},
  edition = {1},
  publisher = {Wiley},
  url = {libgen.li/file.php?md5=4bf619a633b163cb1b1ca2c8528c34cf},
  isbn = {0-471-50459-9 978-0-471-50459-7}
}

@article{kumar2012SamplingMethodsNystrom,
  title = {Sampling Methods for the {{Nyström}} Method},
  author = {Kumar, Sanjiv and Mohri, Mehryar and Talwalkar, Ameet},
  date = {2012},
  journaltitle = {The Journal of Machine Learning Research},
  volume = {13},
  number = {1},
  pages = {981--1006},
  publisher = {JMLR. org}
}

@inproceedings{lacoste-julien2015SequentialKernelHerding,
  title = {Sequential {{Kernel Herding}}: {{Frank-Wolfe Optimization}} for {{Particle Filtering}}},
  shorttitle = {Sequential {{Kernel Herding}}},
  booktitle = {Proceedings of the {{Eighteenth International Conference}} on {{Artificial Intelligence}} and {{Statistics}}},
  author = {Lacoste-Julien, Simon and Lindsten, Fredrik and Bach, Francis},
  date = {2015},
  pages = {544--552},
  publisher = {PMLR},
  issn = {1938-7228},
  url = {https://proceedings.mlr.press/v38/lacoste-julien15.html},
  urldate = {2022-10-03},
  eventtitle = {Artificial {{Intelligence}} and {{Statistics}}},
  langid = {english}
}

@book{laub2004MatrixAnalysisScientists,
  title = {Matrix {{Analysis}} for {{Scientists}} and {{Engineers}}},
  author = {Laub, Alan J.},
  date = {2004},
  publisher = {{SIAM: Society for Industrial and Applied Mathematics}},
  url = {libgen.li/file.php?md5=573356e1091d91db6701b31b3fa37ed1},
  isbn = {978-0-89871-576-7}
}

@inproceedings{locatello2017UnifiedOptimizationView,
  title = {A {{Unified Optimization View}} on {{Generalized Matching Pursuit}} and {{Frank-Wolfe}}},
  booktitle = {Proceedings of the 20th {{International Conference}} on {{Artificial Intelligence}} and {{Statistics}}},
  author = {Locatello, Francesco and Khanna, Rajiv and Tschannen, Michael and Jaggi, Martin},
  date = {2017-04-10},
  pages = {860--868},
  publisher = {PMLR},
  issn = {2640-3498},
  url = {https://proceedings.mlr.press/v54/locatello17a.html},
  urldate = {2023-07-24},
  eventtitle = {Artificial {{Intelligence}} and {{Statistics}}},
  langid = {english}
}

@article{mahoney2009CURMatrixDecompositions,
  title = {{{CUR}} Matrix Decompositions for Improved Data Analysis},
  author = {Mahoney, Michael W. and Drineas, Petros},
  date = {2009},
  journaltitle = {Proceedings of the National Academy of Sciences},
  volume = {106},
  number = {3},
  pages = {697--702},
  publisher = {National Acad Sciences}
}

@inproceedings{muandet2014KernelMeanEstimationa,
  title = {Kernel {{Mean Estimation}} and {{Stein Effect}}},
  booktitle = {Proceedings of the 31st {{International Conference}} on {{Machine Learning}}},
  author = {Muandet, Krikamol and Fukumizu, Kenji and Sriperumbudur, Bharath and Gretton, Arthur and Schoelkopf, Bernhard},
  date = {2014-01-27},
  pages = {10--18},
  publisher = {PMLR},
  issn = {1938-7228},
  url = {https://proceedings.mlr.press/v32/muandet14.html},
  urldate = {2022-03-09},
  eventtitle = {International {{Conference}} on {{Machine Learning}}},
  langid = {english}
}

@thesis{mueller2009KomplexitaetUndStabilitaet,
  type = {doctoralThesis},
  title = {Komplexität und Stabilität von kernbasierten Rekonstruktionsmethoden},
  author = {Müller, Stefan},
  date = {2009-03-19},
  doi = {10.53846/goediss-2481},
  url = {https://ediss.uni-goettingen.de/handle/11858/00-1735-0000-0006-B3BA-E},
  urldate = {2023-07-14},
  langid = {german}
}

@book{novak1988DeterministicStochasticError,
  title = {Deterministic and {{Stochastic Error Bounds In Numerical Analysis}}},
  author = {Novak, Erich},
  date = {1988},
  publisher = {Springer},
  langid = {english}
}

@article{novak2006FunctionSpacesLipschitz,
  title = {Function {{Spaces}} in {{Lipschitz Domains}} and {{Optimal Rates}} of {{Convergence}} for {{Sampling}}},
  author = {Novak, Erich and Triebel, Hans},
  date = {2006-02-01},
  journaltitle = {Constructive Approximation},
  shortjournal = {Constr Approx},
  volume = {23},
  number = {3},
  pages = {325--350},
  issn = {1432-0940},
  doi = {10.1007/s00365-005-0612-y},
  url = {https://doi.org/10.1007/s00365-005-0612-y},
  urldate = {2025-02-12},
  langid = {english}
}

@book{novak2010TractabilityMultivariateProblems,
  title = {Tractability of Multivariate Problems, {{Volume II}}: {{Standard}} Information for Functionals},
  author = {Novak, Erich and Wozniakowski, Henryk},
  date = {2010},
  series = {{{EMS}} Tracts in Mathematics},
  publisher = {European Mathematical Society},
  url = {libgen.li/file.php?md5=235180d9572408371ad75d45b44d28f9},
  isbn = {3-03719-084-1 978-3-03719-084-5}
}

@article{nystroem1930UeberPraktischeAufloesung,
  title = {Über Die Praktische Auflösung von Integralgleichungen mit Anwendungen auf Randwertaufgaben},
  author = {Nyström, E. J.},
  date = {1930},
  journaltitle = {Acta Mathematica},
  shortjournal = {Acta Math.},
  volume = {54},
  pages = {185--204},
  publisher = {Institut Mittag-Leffler},
  issn = {0001-5962, 1871-2509},
  doi = {10.1007/BF02547521},
  url = {https://projecteuclid.org/euclid.acta/1485887849},
  urldate = {2020-08-09},
  langid = {ngerman},
  mrnumber = {MR1555306},
  zmnumber = {56.0342.01}
}

@inproceedings{oates2017ProbabilisticModelsIntegration,
  title = {Probabilistic {{Models}} for {{Integration Error}} in the {{Assessment}} of {{Functional Cardiac Models}}},
  booktitle = {31st {{Conference}} on {{Neural Information Processing Systems}}},
  author = {Oates, Chris and Niederer, Steven and Lee, Angela and Briol, François-Xavier and Girolami, Mark},
  date = {2017},
  location = {Long Beach, CA, USA},
  langid = {english}
}

@inproceedings{paige2016SupersamplingReservoir,
  title = {Super-Sampling with a Reservoir},
  booktitle = {Proceedings of the {{Thirty-Second Conference}} on {{Uncertainty}} in {{Artificial Intelligence}}},
  author = {Paige, Brooks and Sejdinovic, Dino and Wood, Frank},
  date = {2016-06-25},
  series = {{{UAI}}'16},
  pages = {567--576},
  publisher = {AUAI Press},
  location = {Arlington, Virginia, USA},
  isbn = {978-0-9966431-1-5}
}

@inproceedings{pauwels2018RelatingLeverageScores,
  title = {Relating {{Leverage Scores}} and {{Density Using Regularized Christoffel Functions}}},
  booktitle = {Proceedings of the 32nd {{International Conference}} on {{Neural Information Processing Systems}}},
  author = {Pauwels, Edouard and Bach, Francis and Vert, Jean-Philippe},
  date = {2018},
  series = {{{NIPS}}'18},
  pages = {1670--1679},
  publisher = {Curran Associates Inc.},
  location = {Red Hook, NY, USA},
  venue = {Montréal, Canada}
}

@article{pinelis1994optimum,
  title = {Optimum Bounds for the Distributions of Martingales in {{Banach}} Spaces},
  author = {Pinelis, Iosif},
  date = {1994},
  journaltitle = {The Annals of Probability},
  pages = {1679--1706},
  publisher = {JSTOR}
}

@book{quarteroni2008NumericalApproximationPartial,
  title = {Numerical Approximation of Partial Differential Equations (Springer Series in Computational Mathematics)},
  author = {Quarteroni, Alfio and Valli, Alberto},
  date = {2008},
  series = {Springer Series in Computational Mathematics},
  edition = {1st ed. 1994. 2nd printing},
  publisher = {Springer},
  url = {libgen.li/file.php?md5=ac3515b91abdb143730b72d5599d2649},
  isbn = {978-3-540-85267-4 3-540-85267-0}
}

@mvbook{rasmussen2006GaussianProcessesMachine,
  title = {Gaussian Processes in Machine Learning},
  author = {Rasmussen, Carl Edward and Williams, Christopher K. I.},
  date = {2006},
  volume = {2},
  publisher = {MIT press Cambridge, MA},
  organization = {Springer},
  volumes = {3}
}

@book{reed1981FunctionalAnalysisVolume,
  title = {Functional {{Analysis}}, {{Volume}} 1},
  author = {Reed, Michael and Simon, Barry},
  date = {1981},
  series = {Methods of {{Modern Mathematical Physics}}},
  edition = {REV AND ENL},
  publisher = {Academic Press},
  url = {libgen.li/file.php?md5=cac630ef99201ba9335dd4a0c5e61b0f},
  isbn = {0-12-585050-6 978-0-12-585050-6}
}

@article{rivera2022QuadratureRulesSolving,
  title = {On Quadrature Rules for Solving {{Partial Differential Equations}} Using {{Neural Networks}}},
  author = {Rivera, Jon A. and Taylor, Jamie M. and Omella, Ángel J. and Pardo, David},
  date = {2022-04-01},
  journaltitle = {Computer Methods in Applied Mechanics and Engineering},
  shortjournal = {Computer Methods in Applied Mechanics and Engineering},
  volume = {393},
  pages = {114710},
  issn = {0045-7825},
  doi = {10.1016/j.cma.2022.114710},
  url = {https://www.sciencedirect.com/science/article/pii/S0045782522000810},
  urldate = {2023-09-08}
}

@inproceedings{rudi2015LessMoreNystrom,
  title = {Less Is More: {{Nyström}} Computational Regularization},
  booktitle = {Proceedings of the 28th International Conference on Neural Information Processing Systems - Volume 1},
  author = {Rudi, Alessandro and Camoriano, Raffaello and Rosasco, Lorenzo},
  date = {2015},
  series = {{{NIPS}}'15},
  pages = {1657--1665},
  publisher = {MIT Press},
  location = {Cambridge, MA, USA},
  url = {https://papers.nips.cc/paper/2015/hash/03e0704b5690a2dee1861dc3ad3316c9-Abstract.html},
  pagetotal = {9}
}

@inproceedings{rudi2018FastLeverageScore,
  title = {On Fast Leverage Score Sampling and Optimal Learning},
  booktitle = {Advances in {{Neural Information Processing Systems}}},
  author = {Rudi, Alessandro and Calandriello, Daniele and Carratino, Luigi and Rosasco, Lorenzo},
  date = {2018},
  pages = {5672--5682},
  url = {https://dl.acm.org/doi/10.5555/3327345.3327470}
}

@article{santin2022SamplingBasedApproximation,
  title = {Sampling Based Approximation of Linear Functionals in Reproducing Kernel {{Hilbert}} Spaces},
  author = {Santin, Gabriele and Karvonen, Toni and Haasdonk, Bernard},
  date = {2022-03-01},
  journaltitle = {BIT Numerical Mathematics},
  shortjournal = {Bit Numer Math},
  volume = {62},
  number = {1},
  pages = {279--310},
  issn = {1572-9125},
  doi = {10.1007/s10543-021-00870-3},
  url = {https://doi.org/10.1007/s10543-021-00870-3},
  urldate = {2023-07-12},
  langid = {english}
}

@article{schaback2018SuperconvergenceKernelbasedInterpolation,
  title = {Superconvergence of Kernel-Based Interpolation},
  author = {Schaback, Robert},
  date = {2018-11-01},
  journaltitle = {Journal of Approximation Theory},
  shortjournal = {Journal of Approximation Theory},
  volume = {235},
  pages = {1--19},
  issn = {0021-9045},
  doi = {10.1016/j.jat.2018.05.002},
  url = {https://www.sciencedirect.com/science/article/pii/S0021904518300650},
  urldate = {2025-02-07}
}

@inproceedings{shetty2022DistributionCompressionNearlinear,
  title = {Distribution {{Compression}} in {{Near-linear Time}}},
  author = {Shetty, Abhishek and Dwivedi, Raaz and Mackey, Lester},
  date = {2022-10-17},
  eprint = {2111.07941},
  eprinttype = {arXiv},
  eprintclass = {cs, math, stat},
  publisher = {arXiv},
  doi = {10.48550/arXiv.2111.07941},
  url = {http://arxiv.org/abs/2111.07941},
  urldate = {2023-02-15},
  eventtitle = {{{ICLR}} 2022}
}

@article{sriperumbudur2010,
  title = {Hilbert {{Space Embeddings}} and {{Metrics}} on {{Probability Measures}}},
  author = {Sriperumbudur, Bharath K. and Gretton, Arthur and Fukumizu, Kenji and Schölkopf, Bernhard and Lanckriet, Gert R. G.},
  date = {2010},
  journaltitle = {Journal of Machine Learning Research},
  volume = {11},
  pages = {1517--1561},
  issn = {1533-7928},
  url = {http://www.jmlr.org/papers/v11/sriperumbudur10a.html},
  urldate = {2017-09-27},
  issue = {Apr}
}

@article{sriperumbudur2012EmpiricalEstimationIntegral,
  title = {On the Empirical Estimation of Integral Probability Metrics},
  author = {Sriperumbudur, Bharath K. and Fukumizu, Kenji and Gretton, Arthur and Schölkopf, Bernhard and Lanckriet, Gert RG},
  date = {2012},
  journaltitle = {Electronic Journal of Statistics},
  volume = {6},
  pages = {1550--1599},
  publisher = {{Institute of Mathematical Statistics and Bernoulli Society}}
}

@article{tolstikhin2017minimax,
  title = {Minimax Estimation of Kernel Mean Embeddings},
  author = {Tolstikhin, Ilya and Sriperumbudur, Bharath K and Muandet, Krikamol},
  date = {2017},
  journaltitle = {The Journal of Machine Learning Research},
  volume = {18},
  number = {1},
  pages = {3002--3048},
  publisher = {JMLR. org}
}

@inproceedings{tsuji2022PairwiseConditionalGradients,
  title = {Pairwise {{Conditional Gradients}} without {{Swap Steps}} and {{Sparser Kernel Herding}}},
  booktitle = {Proceedings of the 39th {{International Conference}} on {{Machine Learning}}},
  author = {Tsuji, Kazuma K. and Tanaka, Ken’Ichiro and Pokutta, Sebastian},
  date = {2022-06-28},
  pages = {21864--21883},
  publisher = {PMLR},
  issn = {2640-3498},
  url = {https://proceedings.mlr.press/v162/tsuji22a.html},
  urldate = {2022-07-26},
  eventtitle = {International {{Conference}} on {{Machine Learning}}},
  langid = {english}
}

@book{wahba1990SplineModelsObservational,
  title = {Spline Models for Observational Data},
  author = {Wahba, Grace},
  date = {1990},
  series = {{{CBMS-NSF Regional Conference Series}} in {{Applied Mathematics}}},
  edition = {illustrated edition},
  publisher = {{SIAM: Society for Industrial and Applied Mathematics}},
  url = {libgen.li/file.php?md5=53c8198914fde9f3076619a98806ee1f},
  isbn = {978-0-89871-244-5 0-89871-244-0}
}

@book{wendland2004ScatteredDataApproximation,
  title = {Scattered {{Data Approximation}}},
  author = {Wendland, Holger},
  date = {2004},
  series = {Cambridge {{Monographs}} on {{Applied}} and {{Computational Mathematics}}},
  publisher = {Cambridge University Press},
  isbn = {0-521-84335-9 978-0-521-84335-5 978-0-511-26579-2}
}

@article{widom1963AsymptoticBehaviorEigenvalues,
  title = {Asymptotic Behavior of the Eigenvalues of Certain Integral Equations},
  author = {Widom, Harold},
  date = {1963},
  journaltitle = {Transactions of the American Mathematical Society},
  volume = {109},
  number = {2},
  pages = {278--295},
  publisher = {JSTOR}
}

@article{widom1964AsymptoticBehaviorEigenvalues,
  title = {Asymptotic Behavior of the Eigenvalues of Certain Integral Equations. {{II}}},
  author = {Widom, Harold},
  date = {1964},
  journaltitle = {Archive for Rational Mechanics and Analysis},
  volume = {17},
  number = {3},
  pages = {215--229},
  publisher = {Springer}
}

@inproceedings{williams2001UsingNystromMethod,
  title = {Using the {{Nyström}} Method to Speed up Kernel Machines},
  booktitle = {Advances in Neural Information Processing Systems},
  author = {Williams, Christopher and Seeger, Matthias},
  date = {2001},
  pages = {682--688},
  url = {https://dl.acm.org/doi/10.5555/3008751.3008847}
}

@book{yurinsky1995SumsGaussianVectors,
  title = {Sums and {{Gaussian Vectors}}},
  author = {Yurinsky, Vadim},
  date = {1995},
  series = {Lecture {{Notes}} in {{Mathematics}} 1617},
  edition = {1},
  publisher = {Springer-Verlag Berlin Heidelberg},
  isbn = {978-3-540-60311-5}
}

@article{DBLP:journals/ftml/MuandetFSS17,
  author    = {Krikamol Muandet and
               Kenji Fukumizu and
               Bharath K. Sriperumbudur and
               Bernhard Sch{\"{o}}lkopf},
  title     = {Kernel Mean Embedding of Distributions: {A} Review and Beyond},
  journal   = {Found. Trends Mach. Learn.},
  volume    = {10},
  number    = {1-2},
  pages     = {1--141},
  year      = {2017},
  url       = {https://doi.org/10.1561/2200000060},
  doi       = {10.1561/2200000060},
  bibsource = {dblp computer science bibliography, https://dblp.org}
}

@inproceedings{DBLP:conf/aistats/ParkJS16,
  author    = {Mijung Park and
               Wittawat Jitkrittum and
               Dino Sejdinovic},
  editor    = {Arthur Gretton and
               Christian C. Robert},
  title     = {{K2-ABC:} Approximate Bayesian Computation with Kernel Embeddings},
  booktitle = {Proceedings of the 19th International Conference on Artificial Intelligence
               and Statistics, {AISTATS} 2016, Cadiz, Spain, May 9-11, 2016},
  series    = {{JMLR} Workshop and Conference Proceedings},
  volume    = {51},
  pages     = {398--407},
  publisher = {JMLR.org},
  year      = {2016},
  url       = {http://proceedings.mlr.press/v51/park16.html},
  bibsource = {dblp computer science bibliography, https://dblp.org}
}

@article{hayati2020kernel,
  title={Kernel Mean Embedding of Probability Measures and its Applications to Functional Data Analysis},
  author={Hayati, Saeed and Fukumizu, Kenji and Parvardeh, Afshin},
  journal={arXiv preprint arXiv:2011.02315},
  year={2020}
}

@article{DBLP:journals/spm/SongFG13,
  author    = {Le Song and
               Kenji Fukumizu and
               Arthur Gretton},
  title     = {Kernel Embeddings of Conditional Distributions: {A} Unified Kernel
               Framework for Nonparametric Inference in Graphical Models},
  journal   = {{IEEE} Signal Process. Mag.},
  volume    = {30},
  number    = {4},
  pages     = {98--111},
  year      = {2013},
  url       = {https://doi.org/10.1109/MSP.2013.2252713},
  doi       = {10.1109/MSP.2013.2252713},
  bibsource = {dblp computer science bibliography, https://dblp.org}
}

@inproceedings{DBLP:conf/alt/SmolaGSS07,
  author    = {Alexander J. Smola and
               Arthur Gretton and
               Le Song and
               Bernhard Sch{\"{o}}lkopf},
  editor    = {Marcus Hutter and
               Rocco A. Servedio and
               Eiji Takimoto},
  title     = {A Hilbert Space Embedding for Distributions},
  booktitle = {Algorithmic Learning Theory, 18th International Conference, {ALT}
               2007, Sendai, Japan, October 1-4, 2007, Proceedings},
  series    = {Lecture Notes in Computer Science},
  volume    = {4754},
  pages     = {13--31},
  publisher = {Springer},
  year      = {2007},
  url       = {https://doi.org/10.1007/978-3-540-75225-7\_5},
  doi       = {10.1007/978-3-540-75225-7\_5},
  bibsource = {dblp computer science bibliography, https://dblp.org}
}

@inproceedings{DBLP:conf/aaai/KimP18,
  author    = {Kee{-}Eung Kim and
               Hyun Soo Park},
  editor    = {Sheila A. McIlraith and
               Kilian Q. Weinberger},
  title     = {Imitation Learning via Kernel Mean Embedding},
  booktitle = {Proceedings of the Thirty-Second {AAAI} Conference on Artificial Intelligence,
               (AAAI-18), the 30th innovative Applications of Artificial Intelligence
               (IAAI-18), and the 8th {AAAI} Symposium on Educational Advances in
               Artificial Intelligence (EAAI-18), New Orleans, Louisiana, USA, February
               2-7, 2018},
  pages     = {3415--3422},
  publisher = {{AAAI} Press},
  year      = {2018},
  url       = {https://www.aaai.org/ocs/index.php/AAAI/AAAI18/paper/view/16807},
  bibsource = {dblp computer science bibliography, https://dblp.org}
}

@article{DBLP:journals/corr/ZouLPS14,
  author    = {Shaofeng Zou and
               Yingbin Liang and
               H. Vincent Poor and
               Xinghua Shi},
  title     = {Nonparametric Detection of Anomalous Data via Kernel Mean Embedding},
  journal   = {CoRR},
  volume    = {abs/1405.2294},
  year      = {2014},
  url       = {http://arxiv.org/abs/1405.2294},
  eprinttype = {arXiv},
  eprint    = {1405.2294},
  bibsource = {dblp computer science bibliography, https://dblp.org}
}

@inproceedings{DBLP:conf/icml/ZhangSMW13,
  author    = {Kun Zhang and
               Bernhard Sch{\"{o}}lkopf and
               Krikamol Muandet and
               Zhikun Wang},
  title     = {Domain Adaptation under Target and Conditional Shift},
  booktitle = {Proceedings of the 30th International Conference on Machine Learning,
               {ICML} 2013, Atlanta, GA, USA, 16-21 June 2013},
  series    = {{JMLR} Workshop and Conference Proceedings},
  volume    = {28},
  pages     = {819--827},
  publisher = {JMLR.org},
  year      = {2013},
  url       = {http://proceedings.mlr.press/v28/zhang13d.html},
  bibsource = {dblp computer science bibliography, https://dblp.org}
}

@inproceedings{DBLP:conf/nips/MuandetFDS12,
  author    = {Krikamol Muandet and
               Kenji Fukumizu and
               Francesco Dinuzzo and
               Bernhard Sch{\"{o}}lkopf},
  editor    = {Peter L. Bartlett and
               Fernando C. N. Pereira and
               Christopher J. C. Burges and
               L{\'{e}}on Bottou and
               Kilian Q. Weinberger},
  title     = {Learning from Distributions via Support Measure Machines},
  booktitle = {Advances in Neural Information Processing Systems 25: 26th Annual
               Conference on Neural Information Processing Systems 2012. Proceedings
               of a meeting held December 3-6, 2012, Lake Tahoe, Nevada, United States},
  pages     = {10--18},
  year      = {2012},
  url       = {https://proceedings.neurips.cc/paper/2012/hash/9bf31c7ff062936a96d3c8bd1f8f2ff3-Abstract.html},
  bibsource = {dblp computer science bibliography, https://dblp.org}
}

@InProceedings{pmlr-v80-balog18a,
  title = 	 {Differentially Private Database Release via Kernel Mean Embeddings},
  author =       {Balog, Matej and Tolstikhin, Ilya and Sch{\"o}lkopf, Bernhard},
  booktitle = 	 {Proceedings of the 35th International Conference on Machine Learning},
  pages = 	 {414--422},
  year = 	 {2018},
  editor = 	 {Dy, Jennifer and Krause, Andreas},
  volume = 	 {80},
  series = 	 {Proceedings of Machine Learning Research},
  month = 	 {10--15 Jul},
  publisher =    {PMLR},
  url = 	 {https://proceedings.mlr.press/v80/balog18a.html}
}

@article{weisstein2002lambert,
  title={Lambert W-function},
  author={Weisstein, Eric W},
  journal={https://mathworld. wolfram. com/},
  year={2002},
  publisher={Wolfram Research, Inc.}
}

@article{DBLP:journals/jmlr/GrettonBRSS12,
  author    = {Arthur Gretton and
               Karsten M. Borgwardt and
               Malte J. Rasch and
               Bernhard Sch{\"{o}}lkopf and
               Alexander J. Smola},
  title     = {A Kernel Two-Sample Test},
  journal   = {J. Mach. Learn. Res.},
  volume    = {13},
  pages     = {723--773},
  year      = {2012},
  url       = {http://dl.acm.org/citation.cfm?id=2188410},
  bibsource = {dblp computer science bibliography, https://dblp.org}
}

@inproceedings{DBLP:conf/nips/LiCCYP17,
  author    = {Chun{-}Liang Li and
               Wei{-}Cheng Chang and
               Yu Cheng and
               Yiming Yang and
               Barnab{\'{a}}s P{\'{o}}czos},
  editor    = {Isabelle Guyon and
               Ulrike von Luxburg and
               Samy Bengio and
               Hanna M. Wallach and
               Rob Fergus and
               S. V. N. Vishwanathan and
               Roman Garnett},
  title     = {{MMD} {GAN:} Towards Deeper Understanding of Moment Matching Network},
  booktitle = {Advances in Neural Information Processing Systems 30: Annual Conference
               on Neural Information Processing Systems 2017, December 4-9, 2017,
               Long Beach, CA, {USA}},
  pages     = {2203--2213},
  year      = {2017},
  url       = {https://proceedings.neurips.cc/paper/2017/hash/dfd7468ac613286cdbb40872c8ef3b06-Abstract.html},
  bibsource = {dblp computer science bibliography, https://dblp.org}
}

@inproceedings{DBLP:conf/ismb/BorgwardtGRKSS06,
  author    = {Karsten M. Borgwardt and
               Arthur Gretton and
               Malte J. Rasch and
               Hans{-}Peter Kriegel and
               Bernhard Sch{\"{o}}lkopf and
               Alexander J. Smola},
  title     = {Integrating structured biological data by Kernel Maximum Mean Discrepancy},
  booktitle = {Proceedings 14th International Conference on Intelligent Systems for
               Molecular Biology 2006, Fortaleza, Brazil, August 6-10, 2006},
  pages     = {49--57},
  year      = {2006},
  url       = {https://doi.org/10.1093/bioinformatics/btl242},
  doi       = {10.1093/bioinformatics/btl242},
  bibsource = {dblp computer science bibliography, https://dblp.org}
}

@inproceedings{DBLP:conf/iclr/SutherlandTSDRS17,
  author    = {Danica J. Sutherland and
               Hsiao{-}Yu Tung and
               Heiko Strathmann and
               Soumyajit De and
               Aaditya Ramdas and
               Alexander J. Smola and
               Arthur Gretton},
  title     = {Generative Models and Model Criticism via Optimized Maximum Mean Discrepancy},
  booktitle = {5th International Conference on Learning Representations, {ICLR} 2017,
               Toulon, France, April 24-26, 2017, Conference Track Proceedings},
  publisher = {OpenReview.net},
  year      = {2017},
  url       = {https://openreview.net/forum?id=HJWHIKqgl},
  bibsource = {dblp computer science bibliography, https://dblp.org}
}

@article{muller1997integral,
  title={Integral probability metrics and their generating classes of functions},
  author={M{\"u}ller, Alfred},
  journal={Advances in Applied Probability},
  volume={29},
  number={2},
  pages={429--443},
  year={1997},
  publisher={Cambridge University Press}
}

@article{sriperumbudur2009integral,
  title={On integral probability metrics,$\backslash$phi-divergences and binary classification},
  author={Sriperumbudur, Bharath K. and Fukumizu, Kenji and Gretton, Arthur and Sch{\"o}lkopf, Bernhard and Lanckriet, Gert RG},
  journal={arXiv preprint arXiv:0901.2698},
  year={2009}
}

@book{gelman1995bayesian,
	title={Bayesian data analysis},
	author={Gelman, Andrew and Carlin, John B and Stern, Hal S and Rubin, Donald B},
	year={1995},
	publisher={Chapman and Hall/CRC}
}

@book{davis2007methods,
	title={Methods of numerical integration},
	author={Davis, Philip J and Rabinowitz, Philip},
	year={2007},
	publisher={Courier Corporation}
}

@book{newman1999monte,
  title={Monte Carlo methods in statistical physics},
  author={Newman, Mark EJ and Barkema, Gerard T},
  year={1999},
  publisher={Clarendon Press}
}

\end{document}